%% file: aaai2026_arxiv_v2.tex
\documentclass[letterpaper]{article} % DO NOT CHANGE THIS

\input{aaai_header} % Added this line, 07/15/25

\usepackage{aaai2026}  % DO NOT CHANGE THIS
\usepackage{times}  % DO NOT CHANGE THIS
\usepackage{helvet}  % DO NOT CHANGE THIS
\usepackage{courier}  % DO NOT CHANGE THIS
\usepackage[hyphens]{url}  % DO NOT CHANGE THIS
\usepackage{graphicx} % DO NOT CHANGE THIS
\usepackage{natbib}  % DO NOT CHANGE THIS AND DO NOT ADD ANY OPTIONS TO IT
\usepackage{caption} % DO NOT CHANGE THIS AND DO NOT ADD ANY OPTIONS TO IT
\usepackage{algorithm}
\usepackage{bm}
\DeclareMathSizes{10}{10}{6.5}{6.5}
\usepackage{lipsum}
\usepackage{cuted}
\usepackage{capt-of}
\newcommand{\appcref}[1]{%
  \begingroup
    \crefname{section}{Appendix}{Appendices}%
    \cref{#1}%
  \endgroup
}
\title{\ours: Bridging Kinematic Plans and Physical Control at Test Time}
\author{
    Dohun Lim\textsuperscript{\rm 1},
    Minji Kim\textsuperscript{\rm 1},
    Jaewoon Lim\textsuperscript{\rm 1},
    Sungchan Kim\textsuperscript{\rm 1}\footnote{Corresponding author}
}
\affiliations{
    \textsuperscript{\rm 1}Department of Computer Science and Artificial Intelligence, Jeonbuk National University, Korea\\
    \{dohun.lim, kmkmj927, ljy3513, s.k\}@jbnu.ac.kr
}

\begin{document}

\maketitle

\input{figs/teaser}
\input{00_abstract}
\input{01_intro}
\input{02_related_work}
%
\input{03_method_v2}
\input{04_experiments_v2}
%
\input{10_conclusion}
%
\input{14_acknowledgements}
\bibliography{aaai2026}
\clearpage
\begin{strip}
    \centering
    {\LARGE \textbf{\ours: Bridging Kinematic Plans and Physical Control at Test Time\\(\supp)}}\par
    \vspace{6em}
\end{strip}
\bigskip
\input{12_appendix_arxiv}

\end{document}

%% file: aaai_header.tex
\input{aaai_macro}  % Add packages to _macros.tex

\usepackage{xr-hyper}

\makeatletter
\newcommand*{\addFileDependency}[1]{
  \typeout{(#1)}
  \@addtofilelist{#1}
  \IfFileExists{#1}{}{\typeout{No file #1.}}
}

\makeatother

\definecolor{cvprblue}{rgb}{0.21,0.49,0.74}
\usepackage[capitalize]{cleveref}
\crefname{section}{Sec.}{Secs.}
\crefname{table}{Table}{Tables}
\crefname{figure}{Fig.}{Figs.}

%% file: aaai_macro.tex
\usepackage{amsmath}	
\usepackage{amssymb}	
\usepackage{booktabs}
\usepackage{times}
\usepackage{microtype}
\usepackage{epsfig}
\usepackage{caption}
\usepackage{placeins}
\usepackage{color, colortbl}
\usepackage{enumitem}
\usepackage{tabularx}
\usepackage{xstring}
\usepackage{multirow}
\usepackage{xspace}
\usepackage{subcaption}
\usepackage{xcolor}
\usepackage[hang,flushmargin]{footmisc}

% ...

\newcommand{\supp}{Appendix\xspace}
\newcommand{\RV}[1]{{%
    \textbf{%
        \ifstrequal{#1}{1}{\textcolor{red}{R#1}}{%
        \ifstrequal{#1}{2}{\textcolor{blue}{R#1}}{%
        \ifstrequal{#1}{3}{\textcolor{magenta}{R#1}}{%
        \ifstrequal{#1}{4}{\textcolor{teal}{R#1}}{%
                           \textcolor{cyan}{R#1}%
        }}}}%
    }%
}}

\usepackage{multirow}
\usepackage{arydshln}
\usepackage{makecell}
\usepackage{enumitem}
\usepackage{mathtools}
\usepackage{algpseudocode}
\usepackage{kotex}
\usepackage{comment}

\usepackage{tikz}
\usepackage{adjustbox} % optional, for better alignment
\newcommand{\ie}{\textit{i}.\textit{e}.}

\newcommand{\parahead}[1]{\textbf{#1}}

\newcommand{\ours}{BRIC}

\DeclareMathOperator{\loss}{\mathcal{L}}

\usepackage{blindtext}
\usepackage[most]{tcolorbox} 
\definecolor{block-gray}{gray}{0.95}

\newtcolorbox{zitat}[2][]{%
    colback=block-gray,
    boxrule=0pt,
    boxsep=0pt,
    top=2pt, bottom=2pt,
    before skip=2pt, after skip=2pt,
    breakable,
    enhanced jigsaw,
    borderline west={4pt}{0pt}{gray},
    title={#2\par},
    colbacktitle={block-gray},
    coltitle={black},
    fonttitle={\large\bfseries},
    attach title to upper={},
    #1,
}

%% file: figs/teaser.tex
\begin{figure*}[t]
    \centering
    \includegraphics[width=0.9\linewidth]{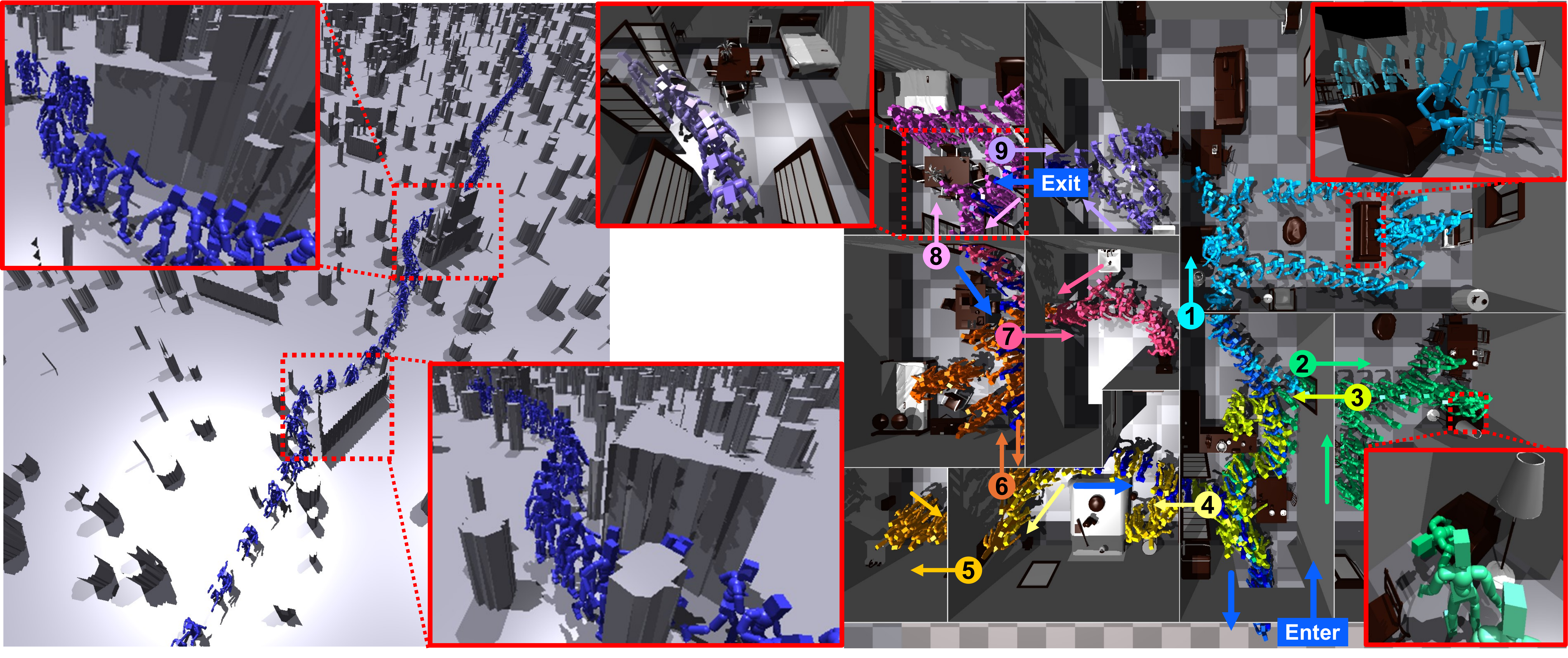}
    \caption{
        \emph{\ours{}} introduces a test-time adaptation (TTA) framework for \emph{physically plausible long-term motion} generation, effectively and efficiently bridging the execution gap between autoregressive diffusion-based kinematic planners and physics-based controllers.
        It enables robust execution of highly extended and challenging tasks, such as (left) reliable \emph{obstacle avoidance over 100$\,\mathrm{m}$} and (right) navigation through an indoor environment spanning about \emph{12 minutes for human-scene interaction}, with the visit order of rooms annotated in the figure.
        Animated videos of these tasks, along with additional visualizations, are available on the project page at \url{https://bric2026.github.io/}.
    }
    \label{fig_teaser}
\end{figure*}

%% file: 00_abstract.tex
\begin{abstract}

We propose \ours{}, a novel test-time adaptation (TTA) framework that enables long-term human motion generation by resolving execution discrepancies between diffusion-based kinematic motion planners and reinforcement learning-based physics controllers.
While diffusion models can generate diverse and expressive motions conditioned on text and scene context, they often produce physically implausible outputs, leading to execution drift during simulation.
To address this, \ours{} dynamically adapts the physics controller to noisy motion plans at test time, while preserving pre-trained skills via a loss function that mitigates catastrophic forgetting.
% In addition, \ours{} introduces a lightweight test-time guidance mechanism that steers the diffusion model in the signal space without requiring parameter updates.
In addition, \ours{} introduces a lightweight test-time guidance mechanism that steers the diffusion model in the signal space without updating its parameters.
By combining both adaptation strategies, \ours{} ensures consistent and physically plausible long-term executions across diverse environments in an effective and efficient manner.
We validate the effectiveness of \ours{} on a variety of long-term tasks, including motion composition, obstacle avoidance, and human-scene interaction, achieving state-of-the-art performance across all tasks.
%Code will be released.

\end{abstract}

%% file: 01_intro.tex
\section{Introduction}
\label{sec:intro}

%
% 1. Why is the problem important?
% 2. Why is the problem difficult?
% 3. What is your key idea? You can point to the teaser figure.
% 4. Give the important technical details.
% 5. How do you evaluate your method? What are the most impressive results?
% 6. A summary of your contribution (possibly as a bullet list).
% 7. Maybe a short overview of the paper.
%

Despite recent advances in computer vision, graphics, and robotics that have enabled substantial progress in generating realistic human motions from text and scene inputs~\cite{xu2023interdiff,yang2024f,diller2024cg,song2024hoianimator,xiao2024unified,pan2024synthesizing,tessler2024maskedmimic,wu2024human,chen2025sitcomcrafterplotdrivenhumanmotion,xu2025intermimic,xu2025parc}, the generation of \emph{physically plausible long-term motion} remains an open challenge.
Diffusion-based text-to-motion models~\cite{tevet2023human, zhang2024motiondiffuse} have recently gained significant attention by leveraging large-scale datasets such as HumanML3D~\cite{guo2022generating}.
% Diffusion-based text-to-motion models~\cite{tevet2023human, zhang2024motiondiffuse} have gained significant attention by leveraging large-scale datasets such as HumanML3D~\cite{guo2022generating}, requiring the composition of diverse motion primitives into coherent long-term sequences~\cite{barquero2024seamless} to achieve smooth transitions~\cite{barquero2024seamless,zhao2025dart}, reliable obstacle avoidance~\cite{karunratanakul2023guided}, and fine-grained human-scene interaction (HSI)~\cite{xiao2024unified,chen2025sitcomcrafterplotdrivenhumanmotion,wang2025sims}.
These models require the composition of diverse motion primitives into coherent long-term sequences~\cite{barquero2024seamless} to achieve smooth transitions~\cite{barquero2024seamless,zhao2025dart}, reliable obstacle avoidance~\cite{karunratanakul2023guided}, and fine-grained human-scene interaction (HSI)~\cite{xiao2024unified,chen2025sitcomcrafterplotdrivenhumanmotion}.

%
%텍스트로부터 사람과 유사한 모션을 생성하는 것은 컴퓨터 비전과 그래픽스, 로보틱스의 중대한 과제이다. 사람의 삶에 대한 모션은 시간 관점에서 매우 길기 때문에 결과적으로 long-term 모션을 생성하는 과제로 이어지며, 이 과제를 달성하기 위해 많은 연구들이 서로 다른 동작간의 부드러운 전이와 장면과의 상호작용을 만족하는 모션을 생성하려고 노력하였다. 하지만 대부분 연구들은 물리 시뮬레이션 기반이 아닌 kinematics 기반 방법에 초점이 맞추어져 floating과 같은 아티팩트로부터 자유롭지 않다는 문제가 있다.

Despite their expressiveness, diffusion-based kinematic models often fail to enforce physical constraints, leading to unrealistic artifacts such as foot skating and floating.
% To address these shortcomings, recent approaches have adopted a two-stage pipeline that improves physical realism by aligning motion sequences executed by a reinforcement learning (RL)-based controller with noisy plans generated by a diffusion model typically by adapting the motion planner to the RL policy~\cite{ren2023insactor, yuan2023physdiff, serifi2024robot}, and vice versa~\cite{tevet2025closd, xu2025parc}.
To address these shortcomings, recent approaches adopt a two-stage pipeline that improves physical realism by aligning motion sequences executed by a reinforcement learning (RL)-based controller with noisy plans generated by a diffusion model, typically by adapting the planner to the policy~\cite{serifi2024robot} or vice versa~\cite{ren2023insactor,tevet2025closd,xu2025parc}.
% This alignment is typically achieved by adapting the motion planner to the RL policy~\cite{serifi2024robot}, and vice versa~\cite{ren2023insactor, tevet2025closd, xu2025parc}.
However, this two-stage structure introduces several limitations.

% \red{[TODO: Long-term 모션 생성 언급이 필요함]}
%우리는 diffusion-based text-to-motion 모델을 모션 플래너로 이용하고 물리 시뮬레이션 환경에서의 강화학습 정책기가 생성된 모션 계획을 실행하는 구조를 선택해서, 물리적인 constraint를 만족하는 동시에 long-term 모션 생성을 달성하고자 한다. 이러한 구조는 \cite{xu2025parc, tevet2025closd, serifi2024robot, yuan2023physdiff, ren2023insactor} 연구들과 유사하지만, 이전 연구들은 long-term 모션 생성에 적합하지 않다는 한계점이 존재한다. 
First, the inherent noise in the generated plans makes physics-based controllers prone to error accumulation over time, leading to drift that impedes the execution of long-term motion sequences~\cite{tevet2025closd, xu2025parc}. 
% \cite{serifi2024robot}은 생성된 모션 계획에 대한 강화학습의 가치함수 값을 최대화시키도록 모션 플래너를 튜닝하지만, 확산 모델의 자체적인 불확실성으로 인해 튜닝된 모션 플래너로부터 생성되는 모션일지라도 노이즈로부터 자유롭지 않다.
%Although~\cite{serifi2024robot} proposes to tune the motion planner to maximize the RL agent’s value function, the generated motions still remain noisy due to the intrinsic stochasticity of diffusion models.
% Second, most methods rely on offline training with hand-crafted surrogate rewards, limiting their generalization to novel goals or environments~\cite{serifi2024robot}.
% 둘째, 두 단계를 통해 모션을 생성하는 구조에서 물리 제어기를 튜닝할 때 발생되는 치명적 망각 현상이다. \cite{xu2025parc, tevet2025closd}는 생성된 노이지 모션 계획을 모방하도록 물리 제어기를 튜닝을 하지만, 이러한 접근은 특정 스타일과 task를 위해 계획된 모션에 과적합되어 일반화 성능 해칠 염려가 있다. 
Second, tuning the policy to track noisy motion plans, as done in recent works~\cite{xu2025parc, tevet2025closd}, can induce catastrophic forgetting, resulting in overfitting to specific styles or tasks and hindering generalization.
% \red{일반적으로, 추가적인 supervision 없이 text-to-motion 모션 플래너는 장면과의 상호작용이 포함된 모션 생성하지 못한다. \cite{janner2022diffuser, karunratanakul2023guided} 연구들은 샘플링 도중에 obstacle avoidance나 joint 위치와 같은 목적 함수를 최소화하도록 샘플을 가이드하였지만, 이는 오히려 물리적으로 비현실적인 노이즈를 심화시킬 수 있어서 노이즈에 강인한 정책기를 훈련하기 위한 일반화된 방법이 대두된다.}
% 또한 우리는 모션 플래너의 추가적인 학습 없이 장면과의 상호작용이 있는 모션을 생성하기 위해, \cite{huang2024constrained, karunratanakul2023guided, karunratanakul2024optimizing, janner2022diffuser} 연구들과 유사하게 샘플링 도중 obstacle avoidance와 같은 목적 함수를 최소화하도록 샘플을 유도하고자 한다. 하지만 유도된 모션 계획에 물리적 제약이 고려되지 않았기 때문에 이러한 방식으로 생성된 모션에 물리적으로 비현실적인 노이즈가 오히려 심화될 수 있다. 따라서 노이즈에 강인하며 일반화된 정책기를 훈련하기 위한 방법이 필요시된다.
Third, diffusion-based models typically cannot perform scene-aware planning or obstacle avoidance without additional supervision~\cite{yi2024generating,caesar2020nuscenes}.
To avoid retraining these motion planners, test-time guidance has emerged as an alternative means of controlling diffusion-based outputs to satisfy such user-defined constraints~\cite{huang2024constrained,rempe2023trace,janner2022diffuser,karunratanakul2024optimizing,karunratanakul2023guided}.
However, since this method does not enforce physical constraints, the guided motions may still exhibit self-collisions or physically implausible contacts.
%Therefore, there is a critical need for methods that can train policies which are robust to noisy plans while maintaining generalization across diverse tasks.

% 이러한 문제를 해결하기 위한 한가지 접근은 test-time adaptation (TTA)이다~\cite{wang2021tent,wang2022continual}.
% TTA는 사전학습된 정책이 실행환경의 분포에 적응할 수 있도록 테스트 시점에 정책 파라메터를 동적으로 업데이트하는 방식으로 입력 분포의 변화나 노이즈에 대응할 수 있는 강인성을 제공한다. 특히 본 연구에서는 물리 시뮬레이션에서 확산 기반 모션 플랜의 노이즈에 적응하기 위한 TTA를 물리 제어기에 적용하는 구조를 채택한다.
% 한편 확산 모델을 재학습하지 않고도 목적 함수에 따라 모션을 조정하려는 test-time guidance 기술들도 주목받고 있다~\cite{huang2024constrained,karunratanakul2024optimizing,rempe2023trace,janner2022diffuser,karunratanakul2023guided}.
% 이들은 목표 도달, 장애물 정보들의 목적 함수를 정의하여 샘플링 중 이를 반영하므로써, 추가 학습 없이도 유도된 모션을 생성할 수 있게 한다.
% 그러나 이러한 방식은 대개 확산 모델의 gradient 계산을 요구하기 때문에, TTA와 함께 사용시 계산 비용이 증가하고~\cite{huang2024constrained,karunratanakul2023guided}, guidance된 모션이 실제 제어기 분포와 정렬되지 않아 실행 실패를 야기할 수 있다.
% 따라서 TTA와의 결합을 고려할때, 보다 효율적이고 robust한 guidance 설계가 필요하다.

To address these challenges, we propose \emph{\ours{}}, a test-time adaptation (TTA) framework~\cite{wang2021tent,wang2022continual} that tightly couples diffusion-based kinematic planners with physics-based controllers.
\ours{} features two key components: (1) a TTA-based motion policy that robustly tracks noisy motion plans, and (2) a lightweight signal-space test-time guidance strategy that aligns motion plans with the controller's executed motion distribution without requiring costly backpropagation through the diffusion model.

\ours{} adapts a physics-based controller (the source domain) to the noisy motion plans generated by a diffusion model (the target domain) by updating the policy parameters online in response to distributional shifts or input noise.
Crucially, our approach preserves previously acquired skills via a catastrophic forgetting-aware loss~\cite{ebrahimi2020adversarial,parisi2019continual}.
% \red{Meanwhile, test-time guidance has emerged as an alternative mean of controlling diffusion-based outputs to meet user-defined objectives, such as goal-reaching or obstacles avoidance~\cite{huang2024constrained,rempe2023trace,janner2022diffuser,karunratanakul2024optimizing,karunratanakul2023guided}.}
% These methods optimizs objective functions during sampling without retraining the model.
% However, most require backpropagation through the diffusion model, leading to high computational costs and reduces compatibility with TTA.
While conventional test-time guidance methods~\cite{huang2024constrained,karunratanakul2023guided,janner2022diffuser} generate motions that satisfy user-defined objectives, they often rely on repeated backpropagation through the diffusion model, resulting in significant computational overhead.
%This incurs substantial computational overhead and reduces compatibility with TTA, which assumes fixed upstream generators.
% \red{Moreover, guided motions may diverge from the physics controller's training distribution, often causing execution failures in simulation.}
% This motivates the need for efficient and robust test-time guidance strategies that complement TTA.
To overcome this limitation, we introduce an efficient test-time guidance strategy that rapidly generates motion plans within the proposed TTA framework without backpropagation, enabling fast and scalable inference.

% 본 논문에서는 이러한 문제들을 해결하기 위해, 확산 기반 운동학 플래너와 물리 제어기를 효율적으로 정렬시키는 새로운 test-tiem adaptation 프레임워크인 \ours{}를 제안한다.
% \ours{}는 다음과 같은 두 가지 핵심 구성요소를 바탕으로 설계된다.
% 첫째, test-time에 모션 정책기를 적응시켜, 노이즈가 포함된모션 플랜을 강인하게 추종할 수 있도록 하며, catastrophic forgetting~\cite{ebrahimi2020adversarial,parisi2019continual}을 방지하는 손실 설계를 통해 기존 스킬의 손상을 최소화한다.
% 둘째, 확산 모델의 파라메터를 고정한 채, 목적 함수를 직접 신호 공간에 반영하는 lightweight한 guidance 방식을 적용하여, 모션 플랜이 제어기의 분포에 더욱 잘 정렬되도록 한다.

By integrating these strategies, \ours{} produces motions that are both consistent with the intended plans and physically executable.
We evaluate \ours{} on long-term tasks including text-driven motion composition, obstacle avoidance, and HSI, demonstrating state-of-the-art performance across all benchmarks.
Our main contributions are as follows:
\begin{itemize}
    \item We introduce \ours{}, a novel TTA framework that adapts a physics-based controller to noisy motion plans generated by a diffusion-based kinematic planner.
    
    \item We propose a new loss function that mitigates catastrophic forgetting, allowing the motion policy to adapt while preserving essential skills.
    
    \item We develop a computationally efficient test-time guidance method that avoids backpropagation through the diffusion model and generalizes to unseen environments.
    
    \item We demonstrate that \ours{} achieves state-of-the-art results on challenging long-term tasks, including motion composition, obstacle avoidance, and HSI.
\end{itemize}

%% file: 02_related_work.tex
\section{Related Work}
\label{sec:related_work}

% \parahead{Test-Time Adaptation in Generative and Control Models.}
% Test-time adaptation(TTA)은 모델이 배포된 이후 테스트 환경 변화에 적응하도록 파라메터를 동적으로 보정하는 방법이다.
% 대표적으로 TENT~\cite{wang2021tent}는 예측 엔트로피 최소화를 통해 배치 정규화 계층의 파라미터를 업데이트하고, CoTTA~\cite{wang2022continual}는 exponential moving average~(EMA) 기반 예측과 파라미터의 확률적 초기화를 통해 catastrophic forgetting을 방지하였다. 
% 최근에는 생성 모델에도 TTA가 적용되며, Diffusion-TTA~\cite{prabhudesai2023diffusion}는 생성 모델의 출력을 활용해 분류기를 테스트 입력에 적응시켜 out-of-distribution(OOD) 상황에서의 정확도를 개선하였다~\cite{tsai2024gda,prabhudesai2023diffusion,yu2023distribution}.
% 우리의 \ours{}는 이러한 TTA 개념을 강화학습 기반 제어기에 적용하여, 확산 모델이 생성한 noisy한 모션 플랜 분포에 적응할 수 있도록 한다. 
% 특히 CoTTA~\cite{wang2022continual}가 분류기에 초점을 둔 반면, \ours{}는 자기회귀적인 정책기와 연속적 상태 변환에 대응해야하는 물리 제어기에 TTA를 적용한다는 점에서 차별성이 있다.
% \red{[todo] reference 형식 정리}
\parahead{Test-Time Adaptation in Generative and Control Models.}
Test-time adaptation (TTA) refers to techniques that dynamically update model parameters during deployment to accommodate distribution shifts.
TENT~\cite{wang2021tent} addresses this by minimizing prediction entropy to update batch normalization parameters, while CoTTA~\cite{wang2022continual} mitigates catastrophic forgetting using exponential moving average predictions and stochastic resets.
More recently, TTA has been extended to generative models.
For instance, Diffusion-TTA~\cite{prabhudesai2023diffusion} adapts a classifier to test-time inputs based on the outputs from a diffusion model, improving robustness under out-of-distribution conditions~\cite{tsai2024gda, prabhudesai2023diffusion, yu2023distribution}.
\ours{} extends this paradigm to RL by adapting a physics-based controller to the noisy motion plan distribution generated by a diffusion model.
Unlike CoTTA, which focuses on classification, \ours{} addresses the more complex problem of adapting an autoregressive control policy for physics-based state transitions, introducing unique challenges specific to control.

% \parahead{Test-Time Guidance for Diffusion-based Motion Generation.} 
% 확산 모델~\cite{ho2016generative}은 이미지~\cite{yin2024one}와 모션~\cite{tevet2023human,zhang2024motiondiffuse} 생성에서 뛰어난 성능을 보이고 있으며, 생성 과정에서 classifier-free guidance~\cite{ho2021classifierfree}를 통해 조건 정보를 반영한다.
% 최근 Reinforcement Learning(RL)의 value function~\cite{rempe2023trace}, way point~\cite{janner2022diffuser}, keyframe point~\cite{karunratanakul2023guided,karunratanakul2024optimizing}, 장애물 회피~\cite{rempe2023trace,karunratanakul2024optimizing}, 장면 정보~\cite{zhao2025dart,yi2024generating} 등을 목적 함수로 활용하여, test-time에 생성된 모션을 제약하는 guidance 방법들이 주목받고 있다.

\parahead{Test-Time Guidance for Diffusion-Based Motion Generation.}
Diffusion models~\cite{ho2020denoising} have shown strong performance in both image~\cite{yin2024one} and motion~\cite{tevet2023human, zhang2024motiondiffuse} generation, often employing classifier-free guidance~\cite{ho2021classifierfree} to incorporate conditioning signals at inference time.
Recent work has explored this paradigm by guiding diffusion-based motion generation through the optimization of objective functions during sampling.
Such objectives include RL value functions~\cite{rempe2023trace}, spatial waypoints~\cite{janner2022diffuser}, keyframes~\cite{karunratanakul2023guided, karunratanakul2024optimizing}, obstacle avoidance~\cite{rempe2023trace, karunratanakul2024optimizing}, and scene-level constraints~\cite{zhao2025dart, yi2024generating}.
%
% 이들 방식은 대개 latent space에서 시작하여 신호 공간(signal space)으로 변환한 후, 목적 함수에 따라 샘플을 최적화한다.
% 그러나 이러한 방식은 확산 모델의 graident를 요구하기 때문에, TTA와 함께 사용할 경우 계산 비용이 높고, 물리 시뮬레이션 환경에서의 실행과 정렬되지 않는 문제가 발생할 수 있다.\red{(ref필요?)}
% \ours{}는 이러한 단점을 보완하기 위해 신호 공간에서 직접 목적 함수를 반영하는 효율적인 guidance 방식을 제안한다.
%
These methods typically operate in the latent space, converting samples to the signal space and optimizing them based on task-specific objectives.
However, they require gradient computation through the diffusion model, which incurs substantial computational cost and limits their applicability in TTA settings.
Moreover, the resulting guided motions are often physically implausible when deployed in physics simulations, leading to execution failures.
To overcome these limitations, \ours{} introduces an efficient test-time guidance method that directly optimizes task objectives in the signal space, without relying on backpropagation through the diffusion model.

% \parahead{Bridging Diffusion Models and Physics-based Controllers.}
% 운동학 기반의 확산 모델~\cite{tevet2023human,zhang2024motiondiffuse}과 물리 기반 제어기~\cite{luo2023perpetual}를 결합하여, 물리적으로 타당한 동작을 생성하려는 시도들이 활발히 이루어지고 있다~\cite{yuan2023physdiff,ren2023insactor,wu2025uniphys,serifi2024robot,tevet2025closd}.
% PhysDiff~\cite{yuan2023physdiff}는 생성된 모션을 시뮬레이션에 투영하여 artifact를 줄였고, InsActor~\cite{ren2023insactor}는 미분 가능한 시뮬레이터를 통해 강화학습 제어기를 훈련하였다.
% CLoSD~\cite{tevet2025closd}는 확산 플래너와 RL 제어기를 폐쇄 루프로 연결하였고, RobotMDM~\cite{serifi2024robot}는 RL 제어기의 value network를 이용해 오프라인으로 모션 플래너를 정렬하였다. 
% UniPhys~\cite{wu2025uniphys}는 제어기와 확산 모델을 단일 구조로 통합하였지만, 조합 확장성과 제어기 모듈성에 제약이 존재한다.
% \ours{}는 이러한 기존 방식들과 달리, 사전학습된 확산 플래너와 RL 제어기 간의 간극을 테스트 시점에서 직접 메우는 구조를 채택한다.
% 이를 통해 다양한 조합과 환경 조건에서도 유연한 적용이 가능하며, 실시간 적응성과 모듈화된 제어 아키텍쳐를 동시에 실현할 수 있다.

\parahead{Bridging Diffusion Models and Physics-Based Controllers.}
An increasing body of research aims to integrate kinematic diffusion models~\cite{tevet2023human, zhang2024motiondiffuse} with physics-based controllers~\cite{luo2023perpetual} to generate physically plausible motion~\cite{yuan2023physdiff, ren2023insactor, wu2025uniphys, serifi2024robot, tevet2025closd}.
PhysDiff~\cite{yuan2023physdiff} reduces unrealistic artifacts by projecting generated motions into simulation, while InsActor~\cite{ren2023insactor} trains a RL controller using a differentiable simulator.
CLoSD~\cite{tevet2025closd} establishes a closed loop between a diffusion-based planner and an RL controller in an autoregressive manner, whereas RobotMDM~\cite{serifi2024robot} aligns the diffusion planner offline using the controller's value function.
UniPhys~\cite{wu2025uniphys} unifies the planner and controller into a single architecture, but at the cost of modularity and composability.
%\red{PARC~\cite{xu2025parc}는 운동학 기반 모션 플래너와 물리 기반 모션 추적기가 서로의 출력을 자가지도학습 신호로 활용하도록 번갈아서 훈련시켰으나, 플래너와 추적기가 동일한 데이터 분포를 따르고 text-to-motion을 지원하지 않음.}
PARC~\cite{xu2025parc} alternates training between the motion planner and the RL policy by using each other's outputs as self-supervised signals.
However, PARC trains both components under a shared data distribution, whereas our method explicitly handles distribution shifts at deployment time.
%and the system does not support text-to-motion generation.
%In contrast, \ours{} adapts the RL policy to the distribution of motion plan different from  bridges the gap between a pretrained diffusion-based planner and an RL controller at test time.
%This design allows modular intergration of heteroneneous components, real-time, and generalization across diverse environments.

%% file: 03_method_v2.tex
\input{figs/overall_procedure}

\section{Method}
\label{sec:method}

% \ours{}는 다음과 같은 세 가지 핵심 구성 요소로 이루어진다: (1) 자기회귀 구조의 확산 기반 motion planner, (2) RL 기반 물리 제어 정책, (3) test-time에서 두 모듈의 정렬을 위한 양방향 적응 기법.
% 본 절에서는 먼저 두 구성 요소인 확산 기반 motion planner와 강화학습 기반 motion controller의 동작 방식을 정리한 후, \ours{}의 전체 구조를 설명한다.

% \ours{} consists of three key components: (1) an autoregressive diffusion-based motion planner, (2) a reinforcement learning based physics controller, and (3) an adaptation mechanism that aligns the two modules at test time.
%\ours{} comprises three key components: (1) an autoregressive diffusion-based motion planner, (2) a RL-based physics controller, and (3) an adaptation mechanism that aligns the distribution of the physics controller with that of the motion planner at test time.
%We begin by providing an overview of the diffusion-based motion planner and the RL-based motion controller, followed by the proposed adaptive framework.

%-------------------------------------------------------------------------
\subsection{Preliminaries} \label{subsec_preliminaries}

\parahead{Diffusion Based Motion Planning.}
We adopt the autoregressive diffusion model DiP from CLoSD~\cite{tevet2025closd} as our motion planner, denoted by $G$, which generates a motion sequence $\boldsymbol{x}_0^{1:H} \in \mathbb{R}^{H \times d}$, where $H$ is the sequence length and $d$ is the dimensionality of each motion frame.
%is based on a diffusion model~\cite{ho2020denoising} and generates motion sequences in an autoregressive manner, A sample $\boldsymbol{x}_0^{1:H} \in \mathbb{R}^{H \times d}$, represented in relative coordinates as in HumanML3D~\cite{guo2022generating},
%Let $H$ denote the length of the generated motion sequence and $d$ the dimensionality of the motion data.
%A sample $\boldsymbol{x}_0^{1:H} \in \mathbb{R}^{H \times d}$, represented in relative coordinates as in HumanML3D~\cite{guo2022generating}
The forward process of $G$ gradually adds Gaussian noise to transform $\boldsymbol{x}_0^{1:H}$ into $\boldsymbol{x}_T^{1:H} \sim \mathcal{N}(\mathbf{0}, \mathbf{I})$, where $\mathbf{I} \in \mathbb{R}^{Hd \times Hd}$ is the identity matrix~\cite{ho2020denoising}.
The model learns the reverse process by minimizing the reconstruction loss as:
%We adopt the autoregressive diffusion model proposed in CLoSD~\cite{tevet2025closd}, which is trained to minimize the following reconstruction loss:
%
\begin{equation} \label{eq_diff_objective}
    \mathcal{L} = \mathbb{E} \left[ \|\boldsymbol{x}_0^{1:H} - G(\boldsymbol{x}_t^{1:H}, \boldsymbol{x}_0^{\text{past}},t,c)\|_2^2 \right],
\end{equation}
where $c$ is a conditioning input that includes a text description and, optionally, the target joint positions. 
% where the condition $c$ includes text and target joint position, 
$\boldsymbol{x}_0^{\text{past}}$ represents the preceding motion, and $t \sim \mathcal{U} [1,T]$ denotes the diffusion step.
% During inference at each step $t$, classifier-free guidance, denoted as $g(\boldsymbol{x}_t^{1:H})$, is applied to interpolate between the unconditional and conditional predictions~\cite{ho2021classifierfree}, progressively denoising $\boldsymbol{x}_t^{1:H}$ toward $\boldsymbol{x}_0^{1:H}$.
%We define $g(\boldsymbol{x}_t^{1:H})$ as the resulting output after applying CFG.
During inference, we apply classifier free guidance~\cite{ho2021classifierfree}, denoted by $g$, to blend conditional and unconditional predictions:
\begin{align*}
%\begin{split}
    % g(\boldsymbol{x}_t^{1:H}) \coloneq &G(\boldsymbol{x}_t^{1:H},t,\boldsymbol{x}_0^{\text{past}},\emptyset) \\
    % &+ s \left\{ G(\boldsymbol{x}_t^{1:H},t,\boldsymbol{x}_0^{\text{past}},c) - G(\boldsymbol{x}_t^{1:H},t,\boldsymbol{x}_0^{\text{past}},\emptyset) \right\}.
    g(\boldsymbol{x}_t^{1:H})
        = (1-s) G(\boldsymbol{x}_t^{1:H},t,\boldsymbol{x}_0^{\text{past}},\emptyset)
        + s G(\boldsymbol{x}_t^{1:H},t,\boldsymbol{x}_0^{\text{past}},c),
%\end{split}
\end{align*}
where $s$ is a guidance scale.
The denoising process progressively refines $\boldsymbol{x}_t^{1:H}$ toward $\boldsymbol{x}_0^{1:H}$.
%%%%%%%%%
% Because the generated sequence $\boldsymbol{x}_0^{1:H}$ is represented in relative coordinates as in HumanML3D~\cite{guo2022generating}, it is then converted into absolute coordinates as $\boldsymbol{x}_{\text{plan}}^{1:H}=\Psi(\boldsymbol{x}_0^{1:H})$ via $\Psi$, which is used for physics simulation.
%%%%%%%%%%%%%%%%

\parahead{Physics Based RL Motion Tracking.}
To track the generated motion plan in simulation, we employ Perpetual Humanoid Control (PHC)~\cite{luo2023perpetual}, a general-purpose RL controller for tracking reference motions.
The controller $\pi$ receives as input a state $\boldsymbol{s}^h = ( \boldsymbol{s}_{\text{p}}^h, \boldsymbol{s}_{\text{g}}^h)$ at time step $h$, where $\boldsymbol{s}_{\text{p}}^h$ is the current proprioceptive state, and $\boldsymbol{s}_{\text{g}}^h$ is the goal state.
Each motion state includes position $\mathcal{P}$, velocity $\dot{\mathcal{P}}$, rotation $\mathcal{R}$, and angular velocity $\dot{\mathcal{R}}$.
% The policy $\pi (\boldsymbol{a}^h | \boldsymbol{s}^h)=\mathcal{N}(\mu(\boldsymbol{s}^h),\sigma)$ outputs an  action $\boldsymbol{a}^h$ interpreted as a proportional-derivative (PD) control target for controlling an SMPL-based humanoid~\cite{loper2015smpl}.
The policy $\pi (\boldsymbol{a}^h | \boldsymbol{s}^h)=\mathcal{N}(\mu(\boldsymbol{s}^h),\sigma)$ is modeled as a Gaussian distribution with a fixed diagonal covariance~\cite{luo2023perpetual}, where the mean $\mu(\boldsymbol{s}^h)$ is predicted by a neural network. The sampled action $\boldsymbol{a}^h$ is interpreted as a proportional-derivative control target for the SMPL-based humanoid~\cite{loper2015smpl}.

As the motion plan provides only positional information, the goal state is formulated using keypoint imitation:
$\boldsymbol{s}_{\text{g}}^h = \boldsymbol{s}_\text{ref}^{h+1} \ominus \boldsymbol{s}_\text{p}^h\coloneq (\mathcal{P}^{h+1}_{\text{ref}}-\mathcal{P}^{h}, \dot{\mathcal{P}}^{h+1}_{\text{ref}}-\dot{\mathcal{P}}^{h}, \mathcal{P}^{h+1}_{\text{ref}})$.
% \red{The subscript ``ref'' denotes reference data.}
% The policy is optimized with PPO~\cite{schulman2017proximal} to maximize a composite reward
% $r^h = r_{\text{imitation}}^h + r_{\text{amp}}^h + r_{\text{energy}}^h$,
% where $r_{\text{imitation}}^h$ encourages accurate motion tracking, $r_{\text{energy}}^h$ discourages excessive efforts~\cite{peng2018deepmimic}, and $r_{\text{amp}}^h$ promotes stylistic realism via an adversarial motion discrimination~\cite{peng2021amp}.
The policy is optimized using PPO~\cite{schulman2017proximal} to maximize a composite reward that includes imitation accuracy, energy efficiency~\cite{peng2018deepmimic}, and stylistic realism via adversarial motion discrimination~\cite{peng2021amp}.

%-------------------------------------------------------------------------
\subsection{Method Overview} \label{subsec_method_overview}

% \cref{fig_overall_procedure}는 test-time에서 확산기반 motion planner와 모션 제어 policy를 정렬하기 위해 \ours{}의 작동 과정을 보여줌.
% \ours{}는 모션 컨트롤러 정책기가 사전학습된 분포(즉 모션 스타일~\red{[AMASS]~\cite{mahmood2019amass}})를 source domain으로, 확산 모델이 출력한 모션 plan의 분포를 target domain으로 설정함.
% 물리 시뮬레이션 측에서 모션 정책기의 파라메터를 업데이트하여 모션 컨트롤러가 확산 모델 출력 분포에 적응하도록 함~(\cref{subsec_policy_tta}).
% TTA와 더불어서 환경의 변화에 실시간적으로 대응하기 위해, 장애물 회피~\cite{rempe2023trace}와 같은 제약조건에 알맞게 모션 플랜을 효과적으로 가이드시키는 test-time guidance을 소개하고자 함~(\cref{subsec_constrained_diffusion}).
%
% \cref{fig_overall_procedure} illustrates the overall procedure of \ours{}, which aligns the diffusion based motion planner and the motion control policy at test time.
We define a long-term task as a sequence of subtasks, with each subtask serving as a minimal behavioral motion unit.
\cref{fig_overall_procedure} illustrates the overall procedure of \ours{}, which consists of three main components: (1) an autoregressive diffusion based motion planner, (2) a RL-based physics controller, and (3) a test-time adaptation mechanism that aligns controller's behavior with the distribution of the motion planner.
%, which aligns representation of the motion control policy to the diffusion based motion planner's at test time.
The adaptation framework in \ours{} considers the output distribution of the pretrained RL policy, trained on a large scale human motion dataset~\cite{mahmood2019amass}, the source domain, and the distribution produced by the diffusion planner as the target domain.
To bridge gap between these domains, \ours{} updates the policy parameters during inference to better track motion sampled from the planner~(\cref{subsec_policy_tta}).
In addition to TTA, we introduce a test time guidance strategy that efficiently directs motion plan generation under task specific constraints, enabling rapid and robust adaptation to dynamic environments~(\cref{subsec_constrained_diffusion}).

%-------------------------------------------------------------------------
\subsection{Test-Time Adaptation of Physics-based Policy to Kinematics Motion}
\label{subsec_policy_tta}

% The propose TTA framework aims at (1) ensuring that the policy accurately tracks the motion plan $\boldsymbol{x}_{\text{plan}}^{1:H}$ generated by the diffusion model and (2) making the policy robust to the inherent noise in the motion plan, thereby preventing tracking failures.
% To achieve this, \ours{} introduces a loss to mitigate error accumulation in the autoregressive policy and to preserve the skills of the original policy by preventing catastrophic forgetting (CF).
The proposed TTA framework aims to ensure that the policy accurately tracks the generated motion plans.
To this end, \ours{} addresses two key challenges: (1) preserving the skills of the original policy by mitigating catastrophic forgetting, and (2) reducing error accumulation in the autoregressive policy caused by noise in the motion plan, which may cause physically implausible behavior and tracking failures.

\parahead{Mitigating Catastrophic Forgetting.}
At test time, a policy that overfits to the motion plan may suffer from catastrophic forgetting, where it loses previously acquired skills from the source domain~\cite{ebrahimi2020adversarial,parisi2019continual}.
This problem is particularly critical in our setting, where both the planner and the controller operate in an autoregressive manner.
% The PHC policy used in \ours{} is trained with PPO~\cite{schulman2017proximal} and consists of three modules: an actor $\pi_{\text{PHC}}$, a value network $V$, and a discriminator $D$, is trained to distinguish between real and fake motions~\cite{peng2021amp}, for style reward.
% \red{The discriminator $D:\tau\in\mathbb{R}^{{d_{\text{AMP}}} \times H_{\text{AMP}}} \rightarrow [0,1]$ takes as input a motion sequence $\tau$ over $H_{\text{AMP}}$ frames and outputs a reward term $r_{\text{AMP}}^h$ that measures style consistency with reference motions.}
Our policy is trained using PPO~\cite{schulman2017proximal} and comprises three components: an actor $\pi$, a value function $V$, and a discriminator $D$ used to compute style rewards by distinguishing reference motions from simulated ones~\cite{peng2021amp}.
% During TTA, the parameters of the three networks, $\theta_{V}$, $\theta_{\pi}$, and $\theta_{D}$, are updated to $\theta_V^{'}$, $\theta_\pi^{'}$, and $\theta_D^{'}$, respectively.
% 우리는 세 개의 네트워크들을 PPO를 통해 훈련시키는 방식으로 TTA를 진행하고자 한다.
% \red{이때 CF를 방지하기 위해, \cite{wang2022continual}에 영감을 받아 우리는 이미 학습된 세 개의 네트워크들을 복사한 $\theta_{\pi}'$와 $\theta_{V}'$, $\theta_{D}'$를 exponential moving average 방식으로 천천히 업데이트 시켰다. $\theta_{\pi}'$와 $\theta_{V}'$, $\theta_{D}'$가 지닌 이전에 학습했던 유용한 능력들을 TTA가 적용 중인 $\theta_{\pi}$와 $\theta_{V}$, $\theta_{D}$에 반영하기 위해, 아래와 같은 CF loss를 최소화시켰다.}

To mitigate catastrophic forgetting during TTA, we introduce a loss that encourages consistency between the current (online) networks and their exponentially averaged counterparts.
%preserves the previously acquired capabilities of $\pi$, $V$, and $D$.
Following the approach in~\cite{wang2022continual}, we maintain the online networks, $\theta_{\pi}$, $\theta_V$, and $\theta_D$, which are updated through PPO during adaptation.
In parallel, we maintain target networks $\theta_{\pi}'$, $\theta_V'$, and $\theta_D'$ updated via exponential moving average:
%of the corresponding online networks $\theta_{\pi}$, $\theta_V$, and $\theta_D$.
%For example, the actor is updated as
$\theta_\pi^{'} \leftarrow \alpha\theta_\pi^{'} + (1-\alpha)\theta_\pi$ with a smoothing factor $\alpha \in [0,1]$.
These target networks serve as memory buffers that retain prior knowledge for use in TTA.
We enforce consistency between the online and target networks using the following loss:
% To preserve existing capabilities, we minimize the following CF loss:
%
% \begin{equation} \label{eq_phc_catastrophic_forget_objective}
% \begin{split}
%     \loss_{\text{CF}} =&
%         ( V(\boldsymbol{s}^h;\theta_{V}) - V(\boldsymbol{s}^h;\theta_{V}^{'}) )^2 + 
%         \| \mu(\boldsymbol{s}^h;\theta_{\pi}) - \mu(\boldsymbol{s}^h;\theta_{\pi}^{'}) \|_2^2 \\
%         +& ( D(\boldsymbol{\tau}; \theta_D) - D(\boldsymbol{\tau};\theta_D^{'}) )^2,
% \end{split}
% \end{equation}
\begin{align} \label{eq_phc_catastrophic_forget_objective}
  \mathcal{L}_{\text{CF}}
    & =\; \bigl(V(\boldsymbol{s}^h;\theta_V) - V(\boldsymbol{s}^h;\theta_V')\bigr)^{2}
        + \bigl\lVert \mu(\boldsymbol{s}^h;\theta_\pi) - \mu(\boldsymbol{s}^h;\theta_\pi') \bigr\rVert_{2}^{2} \notag \\
        & + \bigl(D(\boldsymbol{\tau};\theta_D) - D(\boldsymbol{\tau};\theta_D')\bigr)^{2}
\end{align}
%
% where $\boldsymbol{\tau}$ is a ten-frame motion segment obtained from simulation~\cite{luo2023perpetual,peng2021amp}.
where $\boldsymbol{\tau}$ is a ten-frame motion segment, extracted during simulation~\cite{luo2023perpetual,peng2021amp}.
% The updates are applied using an exponential moving average scheme, where for $\alpha \in [0,1]$, the actor is updated as $\theta_\pi^{'} \leftarrow \alpha\theta_\pi^{'} + (1-\alpha)\theta_\pi$~\cite{wang2022continual}.
% This allows a balance between adaptation and skill retention.

\parahead{Improving Robustness to Noisy Motion Plans.}
Motion plans generated by diffusion models are not constrained by physical laws and may exhibit various forms of noise, including floating, penetration, or abrupt pose transitions~\cite{yuan2023physdiff,ren2023insactor}.
To enhance robustness against such noise, \ours{} introduces a noise-aware loss that measures the KL divergence between the policy distributions conditioned on the executed trajectory and the noisy motion plan.
The generated motion plans $\boldsymbol{x}_0^{1:H}$ provide only relative positional information~\cite{guo2022generating}, whereas the executed trajectory includes both global positions and rotations.
To enable comparison, we define a transformation function $\Psi$ that converts the relative coordinates into global one, yielding $\boldsymbol{x}_{\text{plan}}^{1:H} = \Psi(\boldsymbol{x}_0^{1:H})$.
We then apply a position-to-rotation network (P2R)~\cite{li2021hybrik} to estimate rotations from the position-only inputs.
This process produces global position $\mathcal{P}^h$, rotation $\mathcal{R}^h$, linear velocity $\dot{\mathcal{P}}^h$, and angular velocity $\dot{\mathcal{R}}^h$ for each time step $h$ of the motion plans.
%, from which we construct the policy input state $\boldsymbol{s}_{\text{plan}}^h$.
%In particular, at time step $h$, the global position and rotation are given by $\mathcal{P}_{\text{plan}}^h = \boldsymbol{x}_{\text{plan}}^h$ and $\mathcal{R}_{\text{plan}}^h=P2R(\mathcal{P}_{\text{plan}}^h)$, respectively.
%
% Using these quantities, we construct the policy input state as 
% $\boldsymbol{s}_{\text{plan}}^h = (\boldsymbol{s}_\text{plan-p}^h, \boldsymbol{s}_\text{plan-g}^h)$, where the current motion state is
% $\boldsymbol{s}_{\text{plan-p}}^h \coloneq (\mathcal{P}_{\text{plan}}^h,\mathcal{R}_{\text{plan}}^h,\dot{\mathcal{P}}_{\text{plan}}^h,\dot{\mathcal{R}}_{\text{plan}}^h)$
% and the goal state is defined as
% $\boldsymbol{s}_\text{plan-g}^h \coloneq (\mathcal{P}_{\text{plan}}^{h+1} - \mathcal{P}_{\text{plan}}^h, \dot{\mathcal{P}}_{\text{plan}}^{h+1} - \dot{\mathcal{P}}_{\text{plan}}^h, \mathcal{P}_{\text{plan}}^{h+1})$.
% We define the noise in the motion plan as the difference between the planned and executed states: $\boldsymbol{s}_{\text{plan}}^h - \boldsymbol{s}^h$.
Using these quantities, we construct the planned policy input state as 
$\boldsymbol{s}_{\text{plan}}^h = (\boldsymbol{s}_\text{plan-p}^h, \boldsymbol{s}_\text{plan-g}^h)$, where the current motion state is
$\boldsymbol{s}_{\text{plan-p}}^h \coloneq (\mathcal{P}_{\text{plan}}^h,\mathcal{R}_{\text{plan}}^h,\dot{\mathcal{P}}_{\text{plan}}^h,\dot{\mathcal{R}}_{\text{plan}}^h)$
and the goal state is defined as
$\boldsymbol{s}_\text{plan-g}^h \coloneq (\mathcal{P}_{\text{plan}}^{h+1} - \mathcal{P}_{\text{plan}}^h, \dot{\mathcal{P}}_{\text{plan}}^{h+1} - \dot{\mathcal{P}}_{\text{plan}}^h, \mathcal{P}_{\text{plan}}^{h+1})$.
We define the noise in the motion plan as the difference between the planned and executed states: $\boldsymbol{s}_{\text{plan}}^h - \boldsymbol{s}^h$.
The noise-aware loss encourages the policy to remain robust under such perturbation, and is defined as the KL divergence:
\begin{equation} \label{eq_policy_robustness}
    \loss_{\text{Robust}} = D_{\text{KL}}(
        \pi (\boldsymbol{a} \mid \boldsymbol{s}^h) ||
        \pi(\boldsymbol{a} \mid \boldsymbol{s}^h + \beta (\boldsymbol{s}_{\text{plan}}^h - \boldsymbol{s}^h) ),
\end{equation}
where $\beta \in [0, 1]$ is a scaling factor that is gradually increased during training to ensure stable learning.
%without overwhelming the policy with large perturbations at the beginning.

%\parahead{Final TTA Learning Objective.}
The final loss used to update the motion control policy combines the standard PPO loss with the catastrophic forgetting and robustness objectives as:
\begin{equation} \label{eq_phc_adaptation_objective}
    \loss_{\text{PPO-TTA}} = \loss_{\text{PPO}} + \lambda_{\text{CF}} \loss_{\text{CF}} + \lambda_{\text{Robust}} \loss_{\text{Robust}},
\end{equation}
where $\lambda_{\text{CF}}$ and $\lambda_{\text{Robust}}$ are weighting coefficients.
Please refer to \appcref{appendix_subsec:test_time_adaptation} for additional details of the proposed adaptation framework.

\subsection{Efficient Test-time Guidance}
\label{subsec_constrained_diffusion}

%Existing studies on test time guidance~\cite{karunratanakul2023guided,karunratanakul2024optimizing,rempe2023trace,huang2024constrained} aim to minimize a predefined objective function $\loss_{\text{guide}}: \boldsymbol{x}_{\text{plan}}^{1:H} \rightarrow \mathbb{R}_{\geq 0}$ that encodes task constraints such as the value function in reinforcement learning~\cite{rempe2023trace}, target keypoint positions~\cite{huang2024constrained}, or obstacle avoidance conditions~\cite{rempe2023trace,huang2024constrained}.
Existing approaches to test-time guidance aim to optimize a predefined objective function, denoted as $\loss_{\text{guide}}( \boldsymbol{x}_{\text{plan}}^{1:H} )$, which encodes task-specific constraints~\cite{karunratanakul2023guided,karunratanakul2024optimizing,rempe2023trace,huang2024constrained}.
%These methods typically operate on the noisy latent sample $\boldsymbol{x}_t^{1:H}$ from the diffusion model and incorporate the objective function during sampling.
%Specifically, they reconstruct the motion in signal space as $\boldsymbol{x}_0^{1:H}=g(\boldsymbol{x}_t^{1:H})$, compute the objective, and backpropagate the gradient to the latent space to perform the following update:
Given a noisy latent sample $\boldsymbol{x}_t^{1:H}$, these methods reconstruct the motion in the signal space as $\boldsymbol{x}_0^{1:H} = g(\boldsymbol{x}_t^{1:H})$, evaluate the objective, and backpropagate the gradient to the latent space as:
%to perform the following update:
%
\begin{equation} \label{eq_x_t_update}
% \boldsymbol{x}_t^{1:H} \leftarrow \boldsymbol{x}_t^{1:H} - \eta \nabla{\boldsymbol{x}_t^{1:H}} \loss_{\text{guide}}( \Psi(g(\boldsymbol{x}_{t}^{1:H}))) / | \nabla{\boldsymbol{x}_t^{1:H}} \loss_{\text{guide}}(\Psi(g(\boldsymbol{x}_{t}^{1:H}))) |2,
\boldsymbol{x}_t^{1:H} \leftarrow \boldsymbol{x}_t^{1:H} - \eta \nabla_{\boldsymbol{x}_t^{1:H}} \loss_{\text{guide}}( \Psi(g(\boldsymbol{x}_{t}^{1:H}))),
\end{equation}
where $\eta$ is a step size.
% \red{The updated $\boldsymbol{x}_t^{1:H}$ is then denoised to $\boldsymbol{x}_{t-1}^{1:H}$ via the posterior distribution $q(\boldsymbol{x}_{t-1}^{1:H} | \boldsymbol{x}_{t}^{1:H},g(\boldsymbol{x}_{t}^{1:H}))$.
% This process is repeated iteratively until $t=1$, gradually incorporating the guidance signal into the sampled motion.}
The updated latent sample $\boldsymbol{x}_t^{1:H}$ is then denoised to $\boldsymbol{x}_{t-1}^{1:H}$ via the posterior distribution $q(\boldsymbol{x}_{t-1}^{1:H} | \boldsymbol{x}_{t}^{1:H},g(\boldsymbol{x}_{t}^{1:H}))$.
This process is repeated iteratively until $t = 1$, gradually injecting the guidance signal into the denoising trajectory.
% \red{문장 맞는지?}

%\parahead{Proposed Method for Efficient Sampling.}
% However, this approach requires costly gradients computation in the latent space, which becomes especially burdensome when the diffusion model $G$ employs on a Transformer architecture~\cite{vaswani2017attention}.
However, this approach requires costly gradients computation in the latent space, which becomes especially burdensome when the diffusion model $G$ employs a Transformer architecture~\cite{vaswani2017attention}.
%As a result, test time sampling becomes significantly slower and consumes excessive GPU resources.
To address this inefficiency, we propose a lightweight alternative.
Instead of differentiating through the diffusion model, we directly minimize the objective in the signal space by updating the generated motion:
%$\boldsymbol{x}_0^{1:H}$ rather than the latent sample as:
%
\begin{equation} \label{eq_x_0_update}
% \boldsymbol{x}0^{1:H} \leftarrow \boldsymbol{x}0^{1:H} -
% \eta \nabla{\boldsymbol{x}{0}^{1:H}}
% \loss{\text{guide}}(\Psi(\boldsymbol{x}0^{1:H})) /
% | \nabla{{\boldsymbol{x}}0^{1:H}} \loss{\text{guide}}(\Psi(\boldsymbol{x}_0^{1:H})) |_2.
\boldsymbol{x}_0^{1:H} \leftarrow \boldsymbol{x}_0^{1:H} -
\eta \nabla_{\boldsymbol{x}_{0}^{1:H}}
\loss_{\text{guide}}(\Psi(\boldsymbol{x}_0^{1:H})).
\end{equation}
We then resample the corresponding latent variable using the forward process: $\boldsymbol{x}_t^{1:H} \sim q(\boldsymbol{x}_t^{1:H} | \boldsymbol{x}_0^{1:H})$.
Since this method avoids computing gradients with respect to the parameters of the diffusion model $G$, it offers substantial gains in both memory and computational efficiency.
Our guidance method integrates collision loss~\cite{rempe2023trace}, motion smoothness~\cite{schulman2014motion}, and heading consistency~\cite{guo2022generating} to improve both efficiency and efficacy.

%\parahead{Synergistic effects.}
% Beyond its efficiency, the proposed method supports a wide range of objective functions such as collision avoidance, distance constraints, and keyframe alignment.
% It provides an efficient test time guidance mechanism that enables control over the diffusion model’s output without requiring additional training.
% % Moreover, when combined with TTA, the guided plans can be generated more quickly, which in turn enhances the effectiveness of policy adaptation, as demonstrated in our experiments.
% % \red{TTA+guidance의 synergy 설명으로 변경}
% % \blue{기존의 test-time guidance 방식은 목적 함수를 샘플에 반영하기 위해 확산 모델 $G$에 대한 역전파가 필요하여 높은 지연 시간이 발생한다. 반면, 제안한 방법인 \cref{eq_x_0_update}는 확산 모델을 미분하지 않고도 guidance를 적용할 수 있어 latency를 효과적으로 줄일 수 있다. 이로 인해 \cref{eq_x_0_update}로 생성된 모션 플랜을 target data로 활용할 경우, TTA 과정에서의 적응 시간이 크게 단축된다. 해당 효과에 대한 정량적 결과는 \cref{xx}에 제시되어 있다.}
% \blue{As mentioned, conventional test-time guidance methods~\cite{huang2024constrained,karunratanakul2024optimizing,rempe2023trace} require backpropagation through the diffusion model $G$, resulting in significant latency.
% In contrast, our approach (\cref{eq_x_0_update}) applies guidance without differentiating through $G$, enabling a more efficient sampling process.
% As a result, when the motion plan guided by \cref{eq_x_0_update} is used as target data, the overall adaptation time during TTA is substantially reduced.
% Quantitative results demonstrating this effect are provided in \red{xx}.}
%
Beyond its standalone advantage, this guidance strategy exhibits strong synergy with the proposed TTA framework.
Our signal space optimization in~\cref{eq_x_0_update} eliminates the need for backpropagation through $G$, enabling fast sampling.
This efficiency extends to the entire TTA pipeline: when motion plans generated via this lightweight guidance are used as targets for adaptation, the policy update process becomes significantly faster.
In effect, fast guidance enables fast adaptation, forming a virtuous cycle.
This synergy is a key advantage that improves the practicality of long-term motion generation.
%It enables the rapid synthesis of physically plausible motion outputs, making the overall systems more suitable for real-world deployment.
The full autoregressive procedure between test-time guidance (TTG) and TTA is provided in \appcref{alg:adaptation_policy} in Appendix.

%% file: figs/overall_procedure.tex
\begin{figure*}[t]
    \centering
    \includegraphics[width=1\textwidth]{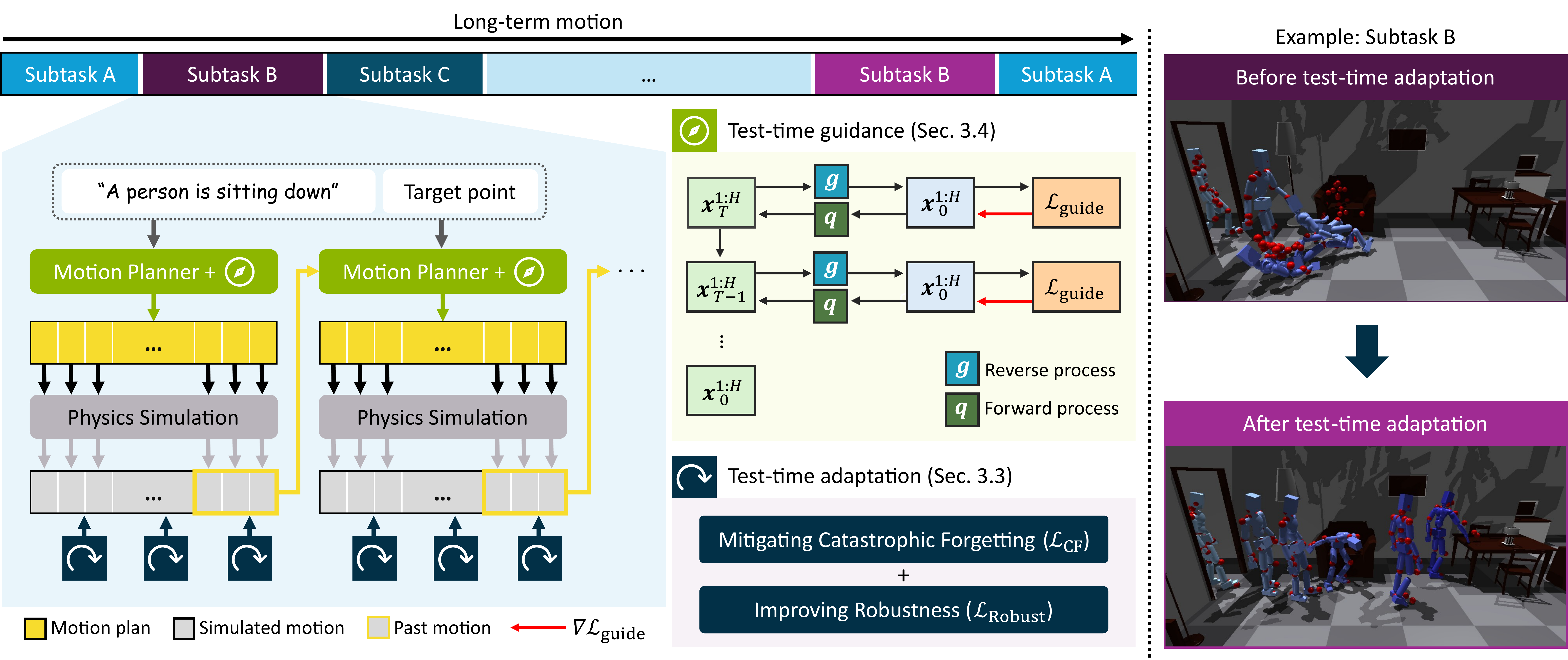}
    \caption{
        % \textbf{The overall procedure.}
        \emph{The overall procedure.}
        % (Left) Test-time adaptation and test-time guidance are applied to bridge the distribution gap between the motion planner and the physics based controller in an autoregressive manner. 
        % (Center) Test-time adaptation (\cref{subsec_policy_tta}) updates the policy to mitigate catastrophic forgetting ($\mathcal{L}_{\text{CF}}$) and improve robustness to noise ($\mathcal{L}_{\text{Robust}}$), while test-time guidance (\cref{subsec_constrained_diffusion}) iteratively refines sampled motion plans by optimizing a task-specific objective.
        % (Right) Before adaptation, the agent struggles with physical plausibility and fails to execute the motion plan reliably.
        % After adaptation, the agent demonstrates robust and physically consistent motions that successfully accomplish the assigned tasks.
        (Left) Test-time adaptation and guidance bridge the distribution gap between the motion planner and physics based controller in an autoregressive manner.
        (Center) Adaptation mitigates catastrophic forgetting ($\mathcal{L}_{\text{CF}}$) and improves robustness ($\mathcal{L}_{\text{Robust}}$), while guidance refines motion plans by optimizing task objectives.
        (Right) Before adaptation, the agent fails to track motion plans indicated by red dots.
        After adaptation, it performs robust and successful motions.
    }
    \label{fig_overall_procedure}
\end{figure*}

%% file: 04_experiments_v2.tex
\section{Experiments}
\label{sec_exp}

\subsection{Setup}
\label{subsec_exp_setup}

\parahead{Tasks.}
We evaluate \ours{} on four long-term tasks: text-to-motion (T2M), goal-reaching, obstacle avoidance, and indoor human-scene interaction (HSI).
For T2M, we follow the FlowMDM protocol~\cite{barquero2024seamless}, generating extended motion sequences by composing subtasks corresponding to from 32 text instructions in the HumanML3D test set~\cite{guo2022generating}. 
% \red{subtasks corresponding to from 32 text instructions} in the HumanML3D test set~\cite{guo2022generating}.

For goal-reaching, we adopt the setup from CLoSD~\cite{tevet2025closd}, where the distal target is defined by scaling an initial distance, uniformly sampled from 1 to 3\,$\mathrm{m}$, by a random factor in $[1, 100]$, yielding target distances up to 300\,$\mathrm{m}$.
For obstacle avoidance, we use terrains with obstacles as defined in~\cite{rempe2023trace,rudin2022learning}, applying the same distance setup as in the goal-reaching task. 
Both tasks are evaluated across four locomotion styles: \emph{walk}, \emph{walk fast}, \emph{jog}, and \emph{jog slow}.

For the HSI task, we use the largest indoor environment in the ProcTHOR dataset~\cite{procthor}, which contains nine interconnected rooms.
The agent is required to navigate all rooms and complete four object-interaction subtasks: \emph{REACH}, \emph{SIT}, \emph{GETUP}, and \emph{TOUCH}.
% We use ChatGPT-4~\cite{achiam2023gpt} to generate HSI scene plan from structured combination of text instructions and scene configurations (see \cref{appendix_sec:scene_plan} for further details).
We define an HSI scene plan as a sequence of such subtasks. It is generated using ChatGPT-4~\cite{achiam2023gpt} from structured combination of text instructions and scene configurations (see \appcref{appendix_sec:scene_plan} for further details).

\parahead{Evaluation Metrics.}
We evaluate motion expressiveness using R-Precision (R-Prec), Multimodal Distance (MM-Dist), Diversity~(Div) and Fréchet Inception Distance (FID)~\cite{guo2022generating}, and assess transition smoothness using Peak Jerk (PJ) and Area Under the Jerk Curve (AUJ)~\cite{barquero2024seamless}.
Physical plausibility is measured by counting occurrences of artifacts such as \emph{penetration}~(Pen), \emph{floating}~(Float), and \emph{foot skating}~(Skate)~\cite{yuan2023physdiff}.
% \red{
% Unlike kinematic models, physics-based simulation may fail to complete motion execution.
% To capture this, we introduce \emph{execution rate}, defined as the ratio of successfully simulated frames to the total number of frames generated by the motion planner.
% %Motion execution fidelity is quantified by the \emph{execution rate}, defined as the ratio of successfully executed motion frames to the total number of frames. 
% We further weight PJ and AUJ by this execution rate, which is especially important for evaluating long-term tasks (see \cref{appendix_subsec:details_evaluation}).
% Finally, we report \emph{success rate} of a task, defined as the fraction of successful trials out of multiple runs.
% % For T2M, results are averaged over ten sampled motion plans sampled per a text prompt.
% For T2M, results are averaged over ten motion sequences generated per text prompt.
% For all other task, results are averaged over 1000 simulation trials.
% }
%
Unlike kinematic models, physics-based simulation may fail to complete motion execution.
To capture this, we introduce \emph{execution rate}, defined as the ratio of successfully simulated frames to the total number of frames generated by the motion planner.
%Motion execution fidelity is quantified by the \emph{execution rate}, defined as the ratio of successfully executed motion frames to the total number of frames. 
We further weight PJ and AUJ by this execution rate, which is especially important for evaluating long-term tasks (see \appcref{appendix_subsec:details_evaluation}).
Finally, we report \emph{success rate} of a task, defined as the fraction of successful trials out of multiple runs.
% For T2M, results are averaged over ten sampled motion plans sampled per a text prompt.
For T2M, results are averaged over ten motion sequences generated per text prompt.
For all other task, results are averaged over 1000 simulation trials.

\textbf{Implementation Details.}
All experiments are conducted using IsaacGym~\cite{makoviychuk2021isaac}.
Following~\cite{tevet2025closd}, we perform TTA every 32 frames.
The condition input $c$ is specified as a text instruction or a combination of a text and a target point.
Each motion plan has a length of $H = 60$, with the motion context $\boldsymbol{x}_0^{\text{past}}$ is set to 20 preceding frames.
% \red{Throughout the experiments, we compare against a baseline model, denoted as \emph {Baseline}, which naively combines PHC~\cite{luo2023perpetual} and DiP~\cite{tevet2025closd} without using test-time adaptation or guidance.}
% Throughout the experiments, we compare against a baseline model, denoted as \emph {Baseline}, which naively combines the physics-based controller PHC~\cite{luo2023perpetual} and the motion planner DiP~\cite{tevet2025closd} without using test-time adaptation or guidance.
% Throughout the experiments, we compare against a baseline model, denoted as \emph {Baseline}, which naively combines PHC~\cite{luo2023perpetual} and DiP~\cite{tevet2025closd} without using test-time adaptation or guidance.
%
% Throughout the experiments, we compare against a baseline model, denoted as \emph {Baseline}, which naively combines PHC and DiP without using test-time adaptation or guidance.
%See \cref{appendix_subsec:experimental_setup} for details.
%
% We apply the proposed test-time guidance from~\cref{eq_x_0_update} only to the obstacle avoidance and HSI tasks, using a collision detection loss~\cite{rempe2023trace}.
We apply the proposed test-time guidance from~\cref{eq_x_0_update} only to the obstacle avoidance and HSI tasks.
For obstacle avoidance, only collision detection loss is used~\cite{rempe2023trace}.
For HSI, we generate target point trajectories for \emph{REACH} actions using the A* algorithm~\cite{hart1968formal}, where collision detection serves as the heuristic function.
This trajectory is fed into all evaluated methods.
% See \cref{appendix_subsec:experimental_setup,appendix_subsec:hsi_description} for details.
See \appcref{appendix_subsec:experimental_setup,appendix_subsec:hsi_description} for details.
\input{tables/longterm_text2motion/main_longterm_text2motion_table_v3}
% \input{figs/qualitative_results}

% \blue{
% \parahead{Comparison of Methods.}
% Throughout the experiments, we compare against a baseline model, denoted as \emph {Baseline}, which naively combines PHC and DiP without using TTA.
% %\ours{} and CLoSD are built upon \emph {Baseline}.
% CLoSD is evaluated for all tasks.
% MaskedMimic~\cite{tessler2024maskedmimic} is compared only for goal-reaching, as it provides no quantitative results for T2M and states poor performance on long-horizon tasks. UniPhys~\cite{wu2025uniphys} is compared only on goal-reaching using reported values.
% UniHSI~\cite{xiao2024unified} is compared only on HSI following the evaluation protocol of CLoSD.
% }
\parahead{Comparison of Methods.}
Throughout the experiments, we compare against a baseline model, denoted as \emph {Baseline}, which naively combines PHC and DiP without using TTA.
%\ours{} and CLoSD are built upon \emph {Baseline}.
CLoSD is evaluated for all tasks.
MaskedMimic~\cite{tessler2024maskedmimic} is compared only for goal-reaching, as it provides no quantitative results for T2M and states poor performance on long-horizon tasks. UniPhys~\cite{wu2025uniphys} is compared only on goal-reaching using reported values.
UniHSI~\cite{xiao2024unified} is compared only on HSI following the evaluation protocol of CLoSD.

\subsection{Text-to-Motion}
\label{subsec:expeiment_text2motion}

We compare \ours{} on the T2M task with state-of-the-art physics-based and kinematic models.
Specifically, we evaluate against the physics-based model CLoSD~\cite{tevet2025closd}, and the kinematic model FlowMDM~\cite{barquero2024seamless}.
As an additional kinematic reference, we also evaluate our motion planner alone, referred to as \emph{DiP}~\cite{tevet2025closd}.
All diffusion models in this evaluation are conditioned solely on text inputs.

\textbf{Results.}
\cref{tab_long_term_t2m} shows that \ours{} (last row) consistently outperforms CLoSD across all metrics,  demonstrating the effectiveness of the proposed TTA in aligning the RL policy with the target motion plan distribution.
Notably, \ours{} achieves nearly twice the performance of CLoSD in two typically conflicting execution rate and FID~\cite{guo2022generating}, highlighting its superior robustness and motion quality.
As expected, kinematic models perform best in diversity and expressiveness metrics.
Nonetheless, \ours{} achieves comparable or better scores in PJ and AUJ, which are critical for long-term motions quality~\cite{barquero2024seamless}.
Moreover, when accounting for failure via execution-weighting metrics, \ours{} maintains a significantly larger advantages over CLoSD in both PJ and AUJ.
% \red{Baseline and CLoSD show lower skate errors only because they fail early, setting later frames to zero.
% In contrast, \ours{} continues generating longer motions, leading to higher raw skate.
% %For a fair comparison, we report a failure-penalized version~(values in parentheses in \cref{tab_long_term_t2m}).
% Under failure-penalized metrics~(values in parentheses in \cref{tab_long_term_t2m}), \ours{} maintains a significantly larger advantage over CLoSD in both PJ and AUJ.}
These results indicate that \ours{} is well-suited for generating coherent and expressive long-term motion with smooth transitions across diverse motion segments.

\subsection{Goal-Reaching and Obstacle Avoidance}
\label{subsec_experiment_longterm_reach_task}

We evaluate \ours{} against CLoSD, MaskedMimic~\cite{tessler2024maskedmimic}, and UniPhys~\cite{wu2025uniphys} using success rate as the primary metric.
For goal-reaching, we follow the standard UniPhys setup, evaluating performance over two target distance ranges: 1 to 2\,$\mathrm{m}$ and 3 to 6\,$\mathrm{m}$.
We also assess a long-term setting as described in~\cref{subsec_exp_setup}.
For both \ours{} and CLoSD, the condition input $c$ includes both text and a target position.
% \red{
% For obstacle avoidance, we employ TTG for \ours{} and CLoSD.
% %since the motion planner is trained without supervision for obstacle avoidance.
% }
%
% \input{tables/longterm_obstacle_avoidance_task_table}
\input{figs/longterm_tasks}
\input{tables/longterm_obstacle_avoidance_task_table_camera_ready_v2}
\input{figs/longterm_hsi_procthor_task/longterm_hsi_procthor_task}
\input{figs/qualitative_results_v2}

\textbf{Results.}
As shown in~\cref{fig_results_goal_reaching_obstacle_avoidance}(a), \ours{} consistently achieves the highest success rates, averaged over \emph{walk} and \emph{jog} locomotion styles in the standard setting.
Under the long-term setting, we compare \ours{} with CLoSD, identified as the strongest existing model, across six distance scales and four locomotion styles: \emph{walk}, \emph{walk fast}, \emph{jog slow}, and \emph{jog}. 
\ours{} demonstrates strong robustness, maintaining a success rate above 0.9 even under the most challenging  conditions.
In contrast, CLoSD suffers significant performance degradation as distance increases, reaching near-zero success under the largest scale (`scale 100').

For obstacle avoidance, \cref{table_obstacle_avoidance_task_camera_ready} shows that both test-time guidance variants of  \ours{} perform comparably to one another, and both outperform CLoSD.
% Implementation details are provided in~\cref{appendix_subsec:goal_reach_description}.
Implementation details are provided in~\appcref{appendix_subsec:goal_reach_description}.

\subsection{Human-Scene Interaction} \label{subsec_exp_HSI}

We evaluate \ours{} on a long-term HSI task, comparing it with CLoSD and UniHSI~\cite{xiao2024unified}.
% The task is organized according to a predefined scene plan consisting of four subtasks, executed within nine-room indoor environment from the ProcTHOR dataset~\cite{procthor}.
The task follows a predefined scene plan composed of four subtasks, executed within a nine-room indoor environment from the ProcTHOR dataset~\cite{procthor}.
For the \emph{REACH} subtask, the motion plan is conditioned on either \emph{walk} or \emph{walk fast}, sampled uniformly.
The scene plan begins with the humanoid agent entering the environment from the outside (\emph{Enter}), visiting each of the nine rooms sequentially from \emph{Room1} to \emph{Room9}, and exiting the scene by returning to the starting point (\emph{Exit}).
In each room, the agent is instructed to navigate toward a target objects while avoiding furniture, and to interact with targets such as sofas or tables with varying geometries.
% Please refer to \cref{appendix_subsec:hsi_description} for details of the HSI task and scene plan construction.
% Please refer to \cref{appendix_subsec:hsi_description} for details of the HSI task.
Please refer to \appcref{appendix_subsec:hsi_description} for details of the HSI task.
% Please refer to \cref{appendix_subsec:hsi_description} and \cref{appendix_sec:scene_plan} for details of the HSI task and scene plan construction, respectively.
%
% \input{figs/longterm_hsi_procthor_task/longterm_hsi_procthor_task}

\textbf{Results.}
\cref{fig_results_HSI} shows that success rate decreases as the agent progresses through rooms for all methods.
Rooms such as \emph{Room1}, \emph{Room2}, and \emph{Room9} are particularly challenging due to complex object layouts, requiring precise navigation and robust generalization in object interaction.
\ours{} maintains a success rate of 0.4, outperforming both CLoSD and UniHSI.
While CLoSD initially performs better than UniHSI, both degrade to near-zero success, and finetuning in UniHSI does not consistently yield improvements.

%\blue{\cite{tevet2025closd} 방법으로 baseline을 test-time guidance과 결합해서 파인튜닝 시킨 결과~(``CLoSD$^{\#}$+TTG''), CLoSD보다 눈에 띄는 성능 향상이 있지만 \ours{}보다 최종 성공률이 좋지 않은 것을 관찰하였다. 이 관찰은 test-time guidance를 통해 모션 계획에 물체와 상호작용 능력을 증진시키도록 생성해서 물리 정책기의 상호작용 능력을 향상시킬 수 있지만, guidance가 적용된 모션 계획에 여전히 물리 시뮬레이션이 실현하기 어려운 노이즈가 포함도었음을 가리킨다. 반면 \ours{}는 강인한 정책기 덕분에 test-time guidance에 효과적으로 대응한다.}
Adding test-time guidance with finetuned CLoSD (``CLoSD$^{\#}$+TTG'') further improves its performance, but \ours{} still achieves the highest final success rate.
This indicates that while test-time guidance is effective, motion plans may remain noisy, highlighting the need for adaptive motion control as in \ours{}.

\subsection{Ablation Study}
\label{subsec_ablation}

The bottom section of \cref{tab_long_term_t2m} presents an ablation study on the effect of the two regularization terms, $\loss_{\text{CF}}$ and $\loss_{\text{Robust}}$, used in policy adaptation.
Each term individually improves motion expressiveness and physical plausibility, while their combination yields the best overall performance.
%``\blue{\ours{}~$-\mathcal{L}_{\text{Robust}} - \mathcal{L}_{\text{CF}}$''은 T2M에 대해서 파인튜닝된  CLoSD$^{\#}$으로, \ours{}와 비교해서 실행 비율과 표현력 메트릭들의 차이가 눈에 띈다. 이는 CLoSD 방법보다 \ours{}가 DiP의 target domain에 더 잘 적응했음을 보여준다.}

\cref{table_obstacle_avoidance_task_camera_ready} shows the effect of the proposed test-time guidance.
Even without adaptation, the ``Baseline+TTG'' variant outperforms CLoSD, demonstrating the benefit of lightweight signal-space optimization. 
%\blue{여기에 TTA와 test-time guidanec가 같이 결합하면 강인하고 일반화된 정책기 덕분에 대폭 성공률이 증가한다.}
Combines TTA to ``Baseline+TTG'' (\ours{}) yields a further increase in success rate, enabled by the adapted policy.
Furthermore, our method in~\cref{eq_x_0_update} achieves $2.0 \times$ faster execution and $2.7 \times$ lower  memory usage compared to the latent-space gradient method in~\cref{eq_x_t_update}, enabling significantly more efficient adaptation.

\subsection{Qualitative Evaluation}
\label{subsec_exp_qualitative_results}

\cref{fig_teaser} visualizes full motion trajectories generated by \ours{} for long-term obstacle avoidance and HSI tasks.
\cref{fig_qualitative_results} presents a qualitative comparison between \ours{} and CLoSD on these tasks.
In obstacle avoidance, CLoSD frequently fails to track the motion plan accurately, leading to collisions or freezing, as shown in \cref{fig_qualitative_results}(a).
\cref{fig_qualitative_results}(b) highlights a \emph{SIT} action performed on an armchair in \emph{Room2} during the HSI task.
%\blue{이 armchair는 양 팔걸이간의 폭이 매우 좁은 물체로, \emph{SIT}을 수행할 때 휴머노이드의 팔이 자주 팔걸이에 걸려서 실패한다.}
CLoSD fails to generalize to the chair's unseen shape, with the agent's arms becoming caught on the armrests.
In contrast, \ours{} performs interaction successfully, demonstrating the benefit of integrating test-time adaptation and guidance for generalizing to novel object geometries.
% See \cref{appendix_sec:visualization} for additional visualization results.
See \appcref{appendix_sec:visualization} for additional visualization results.
%
% \subsection{Limitations}
% \label{subsec_limitation}
%
% \ours{} performs frame-wise local optimization during test-time guidance, which may limit its ability to plan over long-horizon requiring global path awareness.
% If the distribution of subtasks in a long sequence is imbalanced, the adaptation procedure may cause the policy to overfit to frequent subtasks, reducing generalization to rare ones.
% In addition, simulation failures early in the sequence, especially on difficult subtasks, can compound and degrade performance in later stages due to the autoregressive nature of execution.

%% file: tables/longterm_text2motion/main_longterm_text2motion_table_v3.tex
\begin{table*}[t]
    \centering
    \scriptsize
    \begin{tabular}{l c cccc cccc ccc} \toprule
        & \multirow{2}{*}{Succ. / Exec.} & \multicolumn{4}{c}{Subsequences} & \multicolumn{4}{c}{Transition} & \multicolumn{3}{c}{Physics-based metrics} \\
         
        \cmidrule(lr){3-6} \cmidrule(lr){7-10} \cmidrule(lr){11-13} 
        
        & & R-prec~$\uparrow$ & FID~$\downarrow$ & Div~$\rightarrow$ & MM-Dist~$\downarrow$ & FID~$\downarrow$ & Div~$\rightarrow$ & PJ~$\rightarrow$ & AUJ~$\downarrow$ & Float~$\downarrow$ & Skate~$\downarrow$ & Pen~$\downarrow$
        \\ \hline
        
        % GT & - & $22.9$ & $206\cdot 10^{-3}$ & $0.000$ & 0.796 & 0.00 & 9.34 & 2.97 & 0.00 & 9.54 & 0.04 & 0.07 
        GT & - & 0.796 & 0.00 & 9.34 & 2.97 & 0.00 & 9.54 & 0.04 & 0.07 & $22.9$ & $206\cdot 10^{-3}$ & $0.000$ 
        \\ \hline \hline
        
        % FlowMDM~\cite{barquero2024seamless} & - & $21.40$ & $6.64$ & $0.04$ & 0.685 & 0.29 & 9.58 & 3.61 & 1.38 & 8.79 & 0.06 & 0.51 
        FlowMDM & - & 0.685 & 0.29 & 9.58 & 3.61 & 1.38 & 8.79 & 0.06 & 0.51 & $21.40$ & $6.64$ & $0.04$ 
        \\ \hline 
        
        % \red{DART}~\cite{zhao2025dart} & - & $21.40$ & $6.64$ & $0.04$ & 0.685 & 0.29 & 9.58 & 3.61 & 1.38 & 8.79 & 0.06 & 0.51 
        % \\
        
        % DiP~\cite{tevet2025closd} & - & $11.02$ & $7.24$ & $0.07$ & 0.456 & 2.30 & 7.78 & 5.07 & 3.27 & 7.81 & 0.70 & 1.80 
        DiP & - & 0.456 & 2.30 & 7.78 & 5.07 & 3.27 & 7.81 & 0.70 & 1.80 & $11.02$ & $7.24$ & $0.07$ 
        \\ \hline \hline

        % \multirow{2}{*}{Baseline} & \multirow{2}{*}{$0.113$ / $0.367$} & \multirow{2}{*}{$0.234$} & \multirow{2}{*}{$21.63$} & \multirow{2}{*}{$5.02$} & \multirow{2}{*}{$7.40$} & \multirow{2}{*}{$25.23$} & \multirow{2}{*}{$5.08$} & $0.09$ & $1.07$ & $22.75$ & $2.02$ & $0.03$
        % \\
        % & & & & & & & & ($0.25$) & ($5.85$) & ($610.23$) & ($49.60$) & ($0.11$)
        % \\ \hline
        Baseline & $0.113 / 0.367$ & $0.234$ & $21.63$ & $5.02$ & $7.40$ & $25.23$ & $5.08$ & $0.09$~($0.25$) & $1.07$~($5.85$) & $22.75$~($610.23$) & $2.02$~($49.60$) & $0.03$~($0.11$)
        \\
        
        % \multirow{2}{*}{CLoSD} & \multirow{2}{*}{$0.141$ / $0.401$} & \multirow{2}{*}{$0.243$} & \multirow{2}{*}{$19.48$} & \multirow{2}{*}{$5.45$} & \multirow{2}{*}{$7.29$} & \multirow{2}{*}{$22.97$} & \multirow{2}{*}{$5.37$} & $0.09$ & $1.26$ & $22.49$ & $2.81$ & $0.04$
        % \\
        % & & & & & & & & ($0.25$) & ($6.21$) & ($591.15$) & ($47.49$) & ($0.13$)
        % \\ 
        CLoSD & $0.141 / 0.401$ & $0.243$ & $19.48$ & $5.45$ & $7.29$ & $22.97$ & $5.37$ & $0.09$~($0.25$) & $1.26$~($6.21$) & $22.49$~($591.15$) & $2.81$~($47.49$) & $0.04$~($0.13$) 
        \\
        
        \hline \hline

        % \multirow{2}{*}{\ours{}~$-\mathcal{L}_{\text{CF}} - \mathcal{L}_{\text{Robust}}$} & \multirow{2}{*}{$0.322$ / $0.575$} & \multirow{2}{*}{$0.277$} & \multirow{2}{*}{$13.03$} & \multirow{2}{*}{$6.50$} & \multirow{2}{*}{$6.87$} & \multirow{2}{*}{$15.53$} & \multirow{2}{*}{$6.43$} & $0.07$ & $0.42$ & $22.92$ & $3.13$ & $0.03$
        % \\
        % & & & & & & & & ($0.12$) & ($2.08$) & ($383.13$) & ($36.51$) & ($0.09$)
        % \\ \hline
        \ours{}~$-\mathcal{L}_{\text{CF}} - \mathcal{L}_{\text{Robust}}$ & $0.322 / 0.575$ & $0.277$ & $13.03$ & $6.50$ & $6.87$ & $15.53$ & $6.43$ & $0.07$~($0.12$) & $0.42$~($2.08$) & $22.92$~($383.13$) & $3.13$~($36.51$) & $0.03$~($0.09$)
        \\

        % \multirow{2}{*}{\ours{}~$-\mathcal{L}_{\text{CF}}$} & \multirow{2}{*}{$0.428$ / $0.626$} & \multirow{2}{*}{$0.292$} & \multirow{2}{*}{$12.06$} & \multirow{2}{*}{$6.60$} & \multirow{2}{*}{$6.77$} & \multirow{2}{*}{$14.16$} & \multirow{2}{*}{$6.62$} & $0.06$ & $0.30$ & $20.22$ & $4.24$ & $0.02$
        % \\
        % & & & & & & & & ($0.11$) & ($1.61$) & ($301.69$) & ($34.92$) & ($0.09$)
        % \\ \hline
        \ours{}~$-\mathcal{L}_{\text{CF}}$ & $0.428 / 0.626$ & $0.292$ & $12.06$ & $6.60$ & $6.77$ & $14.16$ & $6.62$ & $0.06$~($0.11$) & $0.30$~($1.61$) & $20.22$~($301.69$) & $4.24$~($34.92$) & $0.02$~($0.09$) 
        \\

        % \multirow{2}{*}{\ours{}~$-\mathcal{L}_{\text{Robust}}$} & \multirow{2}{*}{$0.441$ / $0.693$} & \multirow{2}{*}{$0.324$} & \multirow{2}{*}{$8.54$} & \multirow{2}{*}{$7.11$} & \multirow{2}{*}{$6.46$} & \multirow{2}{*}{$9.55$} & \multirow{2}{*}{$7.36$} & $0.07$ & $0.36$ & $19.73$ & $2.52$ & $0.04$
        % \\
        % & & & & & & & & ($0.10$) & ($1.35$) & ($276.44$) & ($32.51$) & ($0.08$)
        % \\ \hline
        \ours{}~$-\mathcal{L}_{\text{Robust}}$ & $0.441 / 0.693$ & $0.324$ & $8.54$ & $7.11$ & $6.46$ & $9.55$ & $7.36$ & $0.07$~($0.10$) & $0.36$~($1.35$) & $19.73$~($276.44$) & $2.52$~($32.51$) & $0.04$~($0.08$)
        \\ 
        
        % \multirow{2}{*}{\ours{}~(Proposed)} & \multirow{2}{*}{$0.494$ / $0.703$} & \multirow{2}{*}{$0.326$} & \multirow{2}{*}{$7.96$} & \multirow{2}{*}{$7.16$} & \multirow{2}{*}{$6.45$} & \multirow{2}{*}{$8.90$} & \multirow{2}{*}{$7.35$} & $0.07$ & $0.45$ & $18.13$ & $3.42$ & $0.04$
        % \\
        % & & & & & & & & ($0.11$) & ($1.52$) & ($243.61$) & ($32.98$) & ($0.17$)
        % \\
        \ours{}~(Proposed) & $0.494 / 0.703$ & $0.326$ & $7.96$ & $7.16$ & $6.45$ & $8.90$ & $7.35$ & $0.07$~($0.11$) & $0.45$~($1.52$) & $18.13$~($243.61$) & $3.42$~($32.98$) & $0.04$~($0.17$)
        \\
        \bottomrule \hline
    \end{tabular}
    % }
    \caption{
        %\textbf{Comparisons on T2M using HumanML3D~\cite{guo2022generating}.} 
        % \textbf{Comparisons on T2M.}
        \emph{Comparisons on T2M.}
        ``$-\loss_{\text{CF}}$'' and ``$-\loss_{\text{Robust}}$'' denote variants of \ours{} with the catastrophic forgetting loss in~\cref{eq_phc_catastrophic_forget_objective} and robustness loss in~\cref{eq_policy_robustness} removed, respectively.
        % \red{``$-\loss_{\text{CF}}$ $-\loss_{\text{Robust}}$'' removes both terms and is equivalent to CLoSD finetuned for T2M.}
        ``$-\loss_{\text{CF}}$ $-\loss_{\text{Robust}}$'' removes both terms and is equivalent to CLoSD finetuned for T2M.
        % Parenthesized numbers indicate values of the physics and transition related metrics weighted by their respective execution rates.
        % \red{``Succ.'' and ``Exec.'' denote success and execution rates, respectively.}
        ``Succ.'' and ``Exec.'' denote success and execution rates, respectively.
        Parenthesized numbers show physics and transition metric values weighted by execution rates.
    }
    \label{tab_long_term_t2m}
\end{table*}

%% file: figs/longterm_tasks.tex
\begin{figure}[t]
    \centering
    \includegraphics[width=1.0\linewidth]{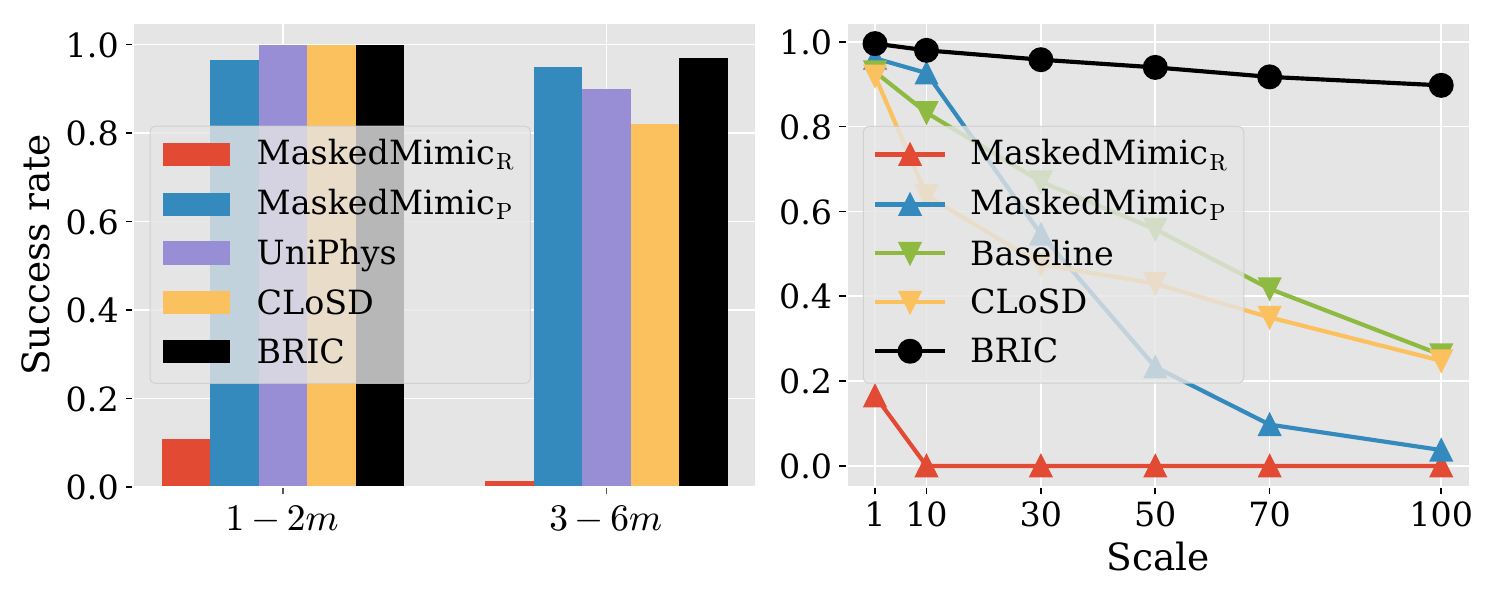}
    \caption{
        % \textbf{Comparisons on goal-reaching.}
        \emph{Comparisons on goal-reaching.}
        (Left) Standard setting and (Right) long-term setting, each evaluated over two target distance ranges.
        MaskedMimic$_{\{\mathrm{R}, \mathrm{P}\}}$ report results for the ``reach'' and ``path-following'' modes, respectively, as described in~\cite{tessler2024maskedmimic}.
    }
    \label{fig_results_goal_reaching_obstacle_avoidance}
\end{figure}

%% file: tables/longterm_obstacle_avoidance_task_table_camera_ready_v2.tex
\begin{table}[t]
    \centering
    \scriptsize
    % \resizebox{1.0\linewidth}{!}{%
    \begin{tabular}{l ccc cccc} \toprule
        \multirow{3}{*}{Method} & \multicolumn{3}{c}{Average Success Rate} & \multicolumn{4}{c}{Computing Cost} \\

        \cmidrule(lr){2-4} \cmidrule(lr){5-8} 

        & \multicolumn{3}{c}{Scale} & \multicolumn{2}{c}{TTA} & \multicolumn{2}{c}{Eval} \\
        \cmidrule(lr){2-4} \cmidrule(lr){5-6} \cmidrule(lr){7-8}
        & 1 & 10 & 30 & Time & Mem & Speed & Mem 
        \\ \hline
        % & \multicolumn{3}{c}{Scale} & \multicolumn{2}{c}{TTA} & \multicolumn{2}{c}{Eval} \\
        % \cmidrule(lr){2-4} \cmidrule(lr){5-6} \cmidrule(lr){7-8}
        % & \multirow{2}{*}{1} & \multirow{2}{*}{10} & \multirow{2}{*}{30} & Time & Mem & Speed & Mem 
        % \\ 
        % & & & & (Sec) & (MiB) & (FPS) & (MiB) \\ \hline
        
        %CLoSD~\cite{tevet2025closd} & $0.80$ & $0.55$ & $0.40$ & - & - & - & -
        CLoSD & $0.80$ & $0.55$ & $0.40$ & - & - & - & -
        \\
        
        %CLoSD$^{*}$~\cite{tevet2025closd} & $0.81$ & $0.60$ & $0.43$ & - & -  & - & -
        Baseline + TTG & $0.81$ & $0.60$ & $0.43$ & - & -  & - & -
        \\ \hline
    
        % BRiC$^{\dagger}$ & $\bold{0.96}$ & $\bold{0.83}$ & $0.71$ & $19.67$ & $43574$ & $8.50$ & $34436$
        %BRiC$^{\dagger}$ & $\bold{0.96}$ & $\bold{0.83}$ & $0.71$ & $30.24$ & $43574$ & $13.58$ & $34436$
        %\\
        \ours{} + LAT$\_$TTG & $\bold{0.96}$ & $\bold{0.83}$ & $0.71$ & $30.24$ & $43574$ & $13.58$ & $34436$
        \\
        
        % BRiC & $0.92$ & $\bold{0.83}$ & $\bold{0.76}$ & $\bold{8.84}$ & $\bold{35092}$ & $\bold{4.35}$ & $\bold{12780}$
        \ours{} & $0.92$ & $\bold{0.83}$ & $\bold{0.76}$ & $\bold{16.21}$ & $\bold{35092}$ & $\bold{27.03}$ & $\bold{12780}$
        \\ \bottomrule \hline
    \end{tabular}
    % }
    % \caption{
    %     % \textbf{Comparisons on obstacle avoidance.} 
    %     \emph{Comparisons on obstacle avoidance.}
    %     ``Baseline + TTG'' applies the proposed test-time guidance to the baseline.
    %     ``\ours{} + LAT$\_$TTG'' uses conventional latent space-based test-time guidance~\cite{huang2024constrained}.
    %     For each adaptation, elapsed ``Time'' in \red{seconds}, ``Speed'' in frames per second (FPS), and the GPU memory usage (``Mem'') \red{in MiB} are reported.
    %     Best results are shown in bold.
    % }
    \caption{
        \emph{Comparisons on obstacle avoidance.}
        ``Baseline + TTG'' applies the proposed test-time guidance to the baseline.
        ``\ours{} + LAT$\_$TTG'' uses conventional latent space-based test-time guidance~\cite{huang2024constrained}.
        For each adaptation, elapsed ``Time'' in seconds, ``Speed'' in frames per second (FPS), and the GPU memory usage (``Mem'') in MiB are reported.
        Best results are shown in bold.
    }
    \label{table_obstacle_avoidance_task_camera_ready}
\end{table}

%% file: figs/longterm_hsi_procthor_task/longterm_hsi_procthor_task.tex
\begin{figure}[t]
    \centering
    \includegraphics[width=1.0\linewidth]{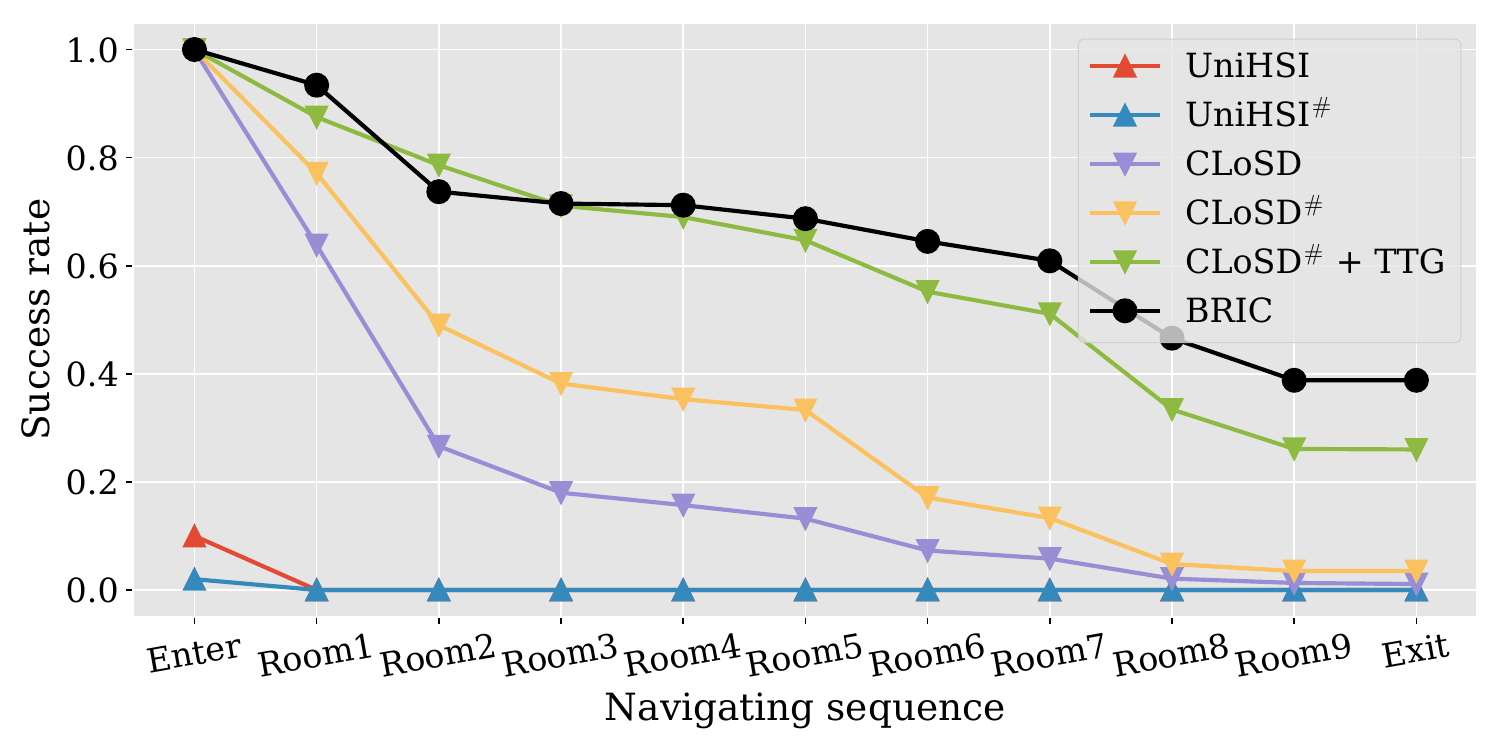}
    % \includegraphics[width=0.7\linewidth]{example-image}
    % \caption{
    %     \textbf{Comparisons on the HSI task.} 
    %     The horizontal axis indicates the sequence of rooms visited by the agent according to the scene plan. 
    %     Models marked with ``$^{\#}$'' are finetuned for the HSI task~\cite{xiao2024unified,tevet2025closd}.
    % }
    \caption{
        % \textbf{Comparisons on the HSI task.}
        \emph{Comparisons on the HSI task}.
        The horizontal axis indicates the sequence of rooms visited by the agent according to the scene plan. 
        Models marked with ``${\#}$'' are finetuned for the HSI task~\cite{xiao2024unified,tevet2025closd}.
    }
    \label{fig_results_HSI}
\end{figure}

%% file: figs/qualitative_results_v2.tex
\begin{figure}[t]
\captionsetup[subfigure]{justification=centering}
    \centering
    \includegraphics[width=0.8\linewidth]{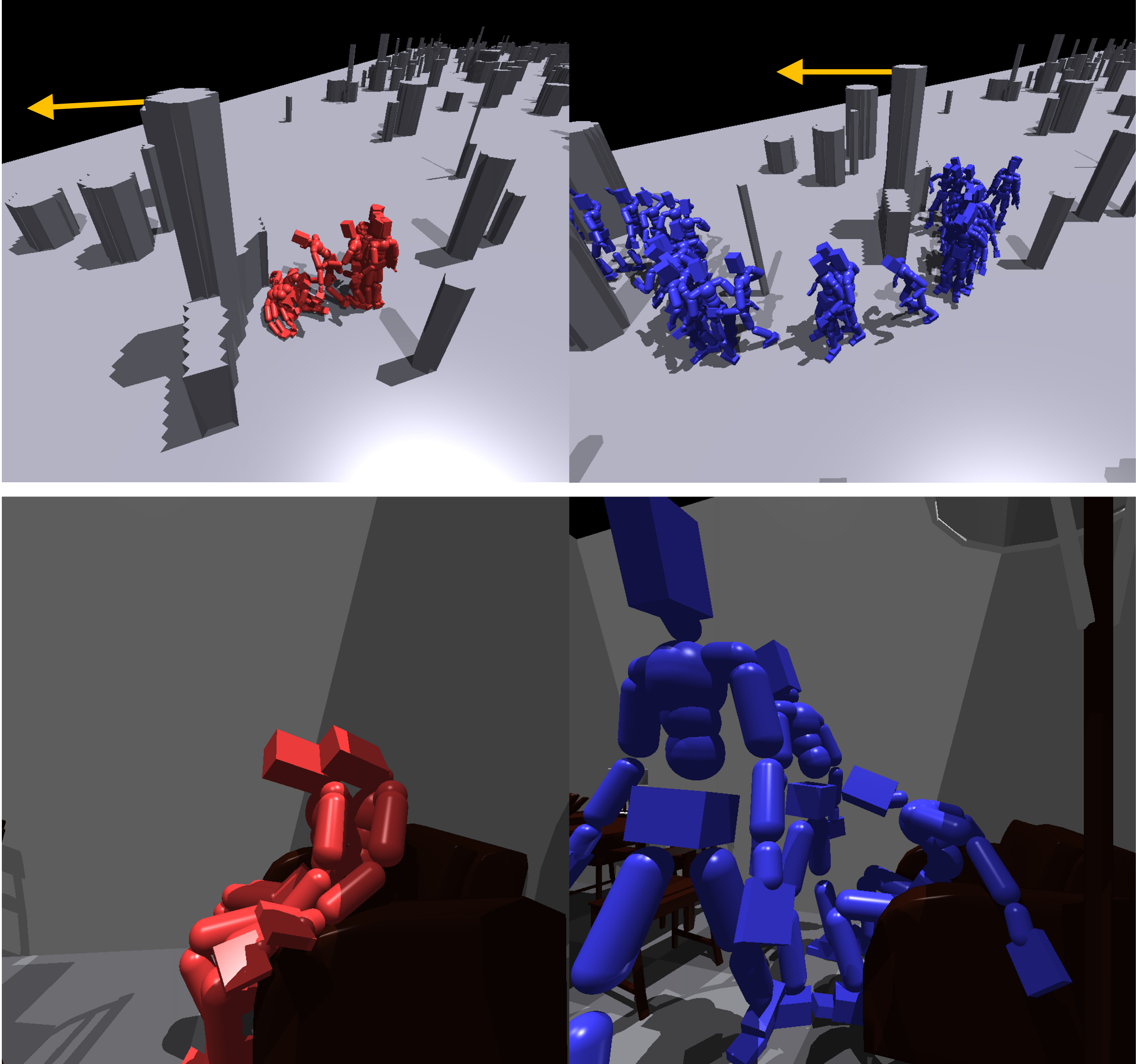}
    % \begin{subfigure}[t]{0.8\linewidth}
    %     \centering
    %     % \includegraphics[width=1\textwidth]{example-image}
    %     \includegraphics[width=1\textwidth]{figs/qualitative_results/qual_res_oa_task_v2.pdf}
    %     %\caption{Long-term obstacle avoidance task}
    %     \caption{}
    % \end{subfigure}\vfil%
    % \begin{subfigure}[t]{0.8\linewidth}
    %     \centering
    %     \includegraphics[width=1\textwidth]{figs/qualitative_results/qual_res_hsi_task_v2.pdf}
    %     %\caption{Long-term human scene interaction}
    %     \caption{}
    % \end{subfigure}\vfil
    % \caption{
    %     % \textbf{Qualitative comparison between CLoSD (red) and our method (blue)}.
    %     \emph{Qualitative comparison between CLoSD (red) and \red{\ours} (blue)}.
    %     (Top) In the obstacle avoidance task, CLoSD fails to navigate around obstacles, while our method succeeds under the same conditions. Motion direction is indicated by yellow arrows.
    %     (Bottom) In the HSI task, CLoSD becomes stuck and fails to sit, whereas \red{\ours} successfully completes \emph{SIT}, \emph{GETUP}, and \emph{REACH} actions (from right to left).
    % }
    \caption{
        \emph{Qualitative comparison between CLoSD (red) and \ours (blue)}.
        (Top) In the obstacle avoidance task, CLoSD fails to navigate around obstacles, while our method succeeds under the same conditions. Motion direction is indicated by yellow arrows.
        (Bottom) In the HSI task, CLoSD becomes stuck and fails to sit, whereas \ours successfully completes \emph{SIT}, \emph{GETUP}, and \emph{REACH} actions (from right to left).
    }
    \label{fig_qualitative_results}
\end{figure}

%% file: 10_conclusion.tex
\section{Conclusion}
\label{sec:conclusion}

We introduced \ours{}, a novel test time adaptation framework for long term motion generation.
\ours{} enables a physics-based policy to robustly execute noisy motion plans from a diffusion-based kinematic planner via online adaptation.
By incorporating a loss that mitigates catastrophic forgetting and a KL divergence auxiliary loss for noise robustness, the policy rapidly adapts while retraining prior skills.
Furthermore, we proposed a lightweight test-time guidance method that optimizes task objectives directly in the signal space without backpropagating through the diffusion model, enabling efficient inference control.
Extensive experiments shows that \ours{} consistently outperforms prior methods in success rates and expressiveness, while preserving physical plausibility.
%As a next step, we plan to expand this unidirectional adaptation framework into a bidirectional setting where both the planner and the controller can be jointly adapted.

%% file: 14_acknowledgements.tex
\section*{Acknowledgements}

This work was supported by the National Research Foundation of Korea (NRF) through the
Korean government (MSIT) under grants (RS-2022-NR069649 and RS-2025-02263810).

%% file: 12_appendix_arxiv.tex
\appendix

\counterwithin{figure}{section}
\counterwithin{table}{section}
\counterwithin{equation}{section}

\section{Implementation Details}
\label{appendix_sec:implementation}

\subsection{Experimental Setup}
\label{appendix_subsec:experimental_setup}

% 이 장에서는 본문에서 제안한 TTA와 test-time guidance에 대해서 보다 자세한 설명을 제공하고자 한다. 먼저, \cref{appendix_subsec:test_time_adaptation}와 \cref{appendix_subsec:test_time_guidance}은 TTA의 과정과 test-time guidance에 대한 자세한 설명을 그리고 \cref{appendix_subsec:details_evaluation}에서는 failure case를 고려한 평가에 대해서 설명을 제공한다.
%This section provides a detailed description of the proposed test-time adaptation (TTA) and test-time guidance.
%\cref{appendix_subsec:test_time_adaptation} and \cref{appendix_subsec:test_time_guidance} describe the procedures for TTA and test-time guidance, respectively, while \cref{appendix_subsec:details_evaluation} explains how evaluation metrics incorporate failure cases.

A physics simulator runs at 60\,Hz, supporting 1024 parallel environments.
We use a physics-based policy that runs at 30\,Hz~\cite{luo2023perpetual}.
% We employ two types of motion planners: $G_{\text{text}}$, trained with text only, and $G_{\text{text+target}}$, trained with both text and target location.
Two types of motion planners are employed: $G_{\text{text}}$, trained with text only, and $G_{\text{text+target}}$, trained with both text and target location.
Texts and target positions, as the input condition $c$, are embedded using DistilBERT~\cite{sanh2019distilbert} and a multilayer perceptron (MLP), respectively.
% These embeddings are fed into the Transformer decoder of the motion planner to enable cross-attention with $\boldsymbol{x}_{t}^{1:H}$. 
These embeddings are passed into the Transformer decoder of the motion planner to facilitate cross-attention with $\boldsymbol{x}_{t}^{1:H}$.
We use a guidance scale $s$ of 5 and 7.5 for classifier-free guidance in $G_{\text{text}}$ and $G_{\text{text+target}}$, respectively~\cite{tevet2025closd}. 

The three networks used during TTA, \emph{Policy}, \emph{Value}, and \emph{Discriminator}, are initialized with pre-trained parameters.
% The loss coefficients in~\cref{eq_phc_adaptation_objective} are set to $\lambda_{\text{robust}} \in \{10^{-4}, 10^{-5}\}$ and $\lambda_{\text{CF}} = 1$.
All hyperparameters used for \ours{} are summarized in \cref{appendix_tab:bric_hparams}.
% \todo{표1이 되도록}
% Policy updates for TTA are performed over 4000 epochs for HSI and 2000 epochs for the other tasks.
% The entire procedure of TTA takes approximately 6 hours for T2M and 4 days for HSI on a Nvidia RTX 6000 Ada GPU.
The entire procedure of TTA takes approximately 6 hours for T2M, 4 hours for goal-reaching, 8 hours for obstacle avoidance, and 4 days for HSI, respectively, on an Nvidia RTX 6000 Ada GPU.
%For each task, we evaluate \ours{} after independently adapting \red{XXX epochs}.(맞는지?)
%%%%%%%%%%% -------------------------------------------------

\subsection{Execution Rate-Aware Metric}
\label{appendix_subsec:details_evaluation}

% \red{Notation 수정해야함.}
This section explains how we compute failure-aware evaluation metrics for text-to-motion generation. 
% Specifically, the metrics that incorporate execution failures include PJ and AUJ, which measure smoothness across motion transitions, as well as Pen, Skate, and Float, which assess physical plausibility.
Failure-aware evaluation is conducted on the PJ, AUJ, Pen, Skate, and Float metrics.
Except for PJ, all these metrics are designed such that lower values indicate better performance.
For PJ, a value closer to 0.04 is desirable.
Let $\boldsymbol{x}_{\text{exec}}^{1:H}$ be the executed trajectory obtained via physics simulation, given a $\boldsymbol{x}_{\text{plan}}^{1:H}$.
% We define the execution rate $\text{ExecutionRate}(\boldsymbol{x}_{\text{exec}}^{1:H},\boldsymbol{x}_{\text{plan}}^{1:H})$ as the ratio between the lengths of $\boldsymbol{x}_{\text{plan}}^{1:H}$ and $\boldsymbol{x}_{\text{exec}}^{1:H}$.
We define the execution rate $\text{ExecutionRate}(\boldsymbol{x}_{\text{exec}}^{1:H},\boldsymbol{x}_{\text{plan}}^{1:H})$ as the ratio of the lengths of $\boldsymbol{x}_{\text{plan}}^{1:H}$ and $\boldsymbol{x}_{\text{exec}}^{1:H}$.
To account for failure cases, we apply a weighted average using the inverse of the execution rate.
Let $\text{Pen}(\boldsymbol{x}_{\text{exec}}^{1:H}) = \max\left(0, \left( \min_{k \in \{1, \dots, K\}} \boldsymbol{x}_{\text{exec},k}^{1:H}\right) - 0.005 \right)$ be the penetration value of $\boldsymbol{x}_{\text{exec}}^{1:H}$, where $K$ denotes the number of joints in the humanoid. We compute the weighted average as
\begin{equation*}
% \mathbb{E}_{\mathcal{M}_{\text{HML3D}}} \left[ 1 / \text{ExecutionRate}_i \cdot \text{Pen}_{i} \right].
\mathbb{E}_{\boldsymbol{x}_{\text{exec}}^{1:H},\boldsymbol{x}_{\text{plan}}^{1:H}} \left[ \frac{1}{\text{ExecutionRate}(\boldsymbol{x}_{\text{exec}}^{1:H},\boldsymbol{x}_{\text{plan}}^{1:H})} \cdot \text{Pen}(\boldsymbol{x}_{\text{exec}}^{1:H}) \right].
\end{equation*}
%
% We measure the other metrics using the failure-aware protocol, in the same manner as above.
The other metrics are evaluated using the same failure-aware protocol.
\input{tables/bric_task_specific_hyperparameters}

\section{Method Details}
\label{appendix_sec:method_implementation}

\subsection{Test-Time Adaptation}
\label{appendix_subsec:test_time_adaptation}

% \textbf{Adaptation Process.}
Given a textual instruction and a target position as condition inputs, the motion planner $G$ generates a motion sequence $\boldsymbol{x}_{\text{plan}}^{1:H}$ using classifier-free guidance.
The adaptation procedure is detailed in \cref{alg:adaptation_policy}. 
We adopt the same Proximal Policy Optimization~(PPO)~\cite{schulman2017proximal} objective as in CLoSD~\cite{tevet2025closd}, which is composed of the policy loss $\loss_{\pi}$, value loss $\loss_{V}$, and discriminator loss $\loss_{D}$:
\begin{equation}
\begin{split}
    &\loss_{\pi} = \mathbb{E}_{\mathcal{B}_{\text{PPO}}} \left[ \min \left( \text{rat}(\pi) \mathcal{A}^h, \text{clip}(\text{rat}(\pi), 1-\epsilon_{\text{clip}},1+\epsilon_{\text{clip}}) \right) \right], \\
    &\loss_{V} = \mathbb{E}_{\mathcal{B}_{\text{PPO}}} \left[ ( V(\boldsymbol{s^h}) -V_{\text{target}}^h )^2 \right], \\
    &\loss_{D} = \mathbb{E}_{(\boldsymbol{\tau}_{\text{stand}},\boldsymbol{\tau}) \sim\mathcal{B}_{ \text{AMP}}} \left[ -\log D(\boldsymbol{\tau}_{\text{stand}}) -\log (1-D( \boldsymbol{\tau}))  \right],
\end{split}
\end{equation}
where $\text{rat}(\pi)=\frac{\pi(\boldsymbol{a}^h|\boldsymbol{s}^h;\theta_{\pi})}{\pi(\boldsymbol{a}^h|\boldsymbol{s}^h;\theta_{\pi}^{\text{old}})}$ is the policy ratio,
$V_{\text{target}}^h$ is the cumulative reward for a trajectory of length 32 using a discounting factor $\gamma$, and $\mathcal{A}^h = Q(\boldsymbol{s}^h,\boldsymbol{a}^h)-V(\boldsymbol{s}^h)$ is the advantage estimate at timestep $h$.

% The discriminator loss $\loss_{D}$ follows the original CLoSD implementation, where $\boldsymbol{\tau}_{\text{stand}}$ is a 10-frame trajectory representing standing poses.
The discriminator loss $\loss_{D}$ follows the original CLoSD implementation, where $\boldsymbol{\tau}_{\text{stand}}$ is a 10-frame trajectory constructed by replicating a single standing pose state $\boldsymbol{s}_{\text{stand}}$.
The humanoid agent is initialized in a standing pose for all experiments.

As illustrated in \cref{alg:adaptation_policy}, \ours{} differs from CLoSD by incorporating additional regularization losses into the PPO objective.
Specifically, the additional operations introduced by \ours{} are shown in lines 14 to 18 and 24 to 26 of \cref{alg:adaptation_policy}.
% Specifically, the operations introduced by \ours{} appear in lines 14 to 17 and 23 to 25 of \cref{alg:adaptation_policy}.
%
Details of the $P2R$ network used in lines 14–18 are provided in \cref{appendix_subsec:pos_to_rot_network}. 
General PPO hyperparameters are summarized in \cref{appendix_tab:ppo_hparams}.
\input{algorithm/test_time_adapation_alg}
\input{tables/ppo_hyperparameters}
%
\input{algorithm/diffusion_sampling_with_test_time_guidance_alg}
%
% \textbf{Hyper-parameters.}
% PPO와 관련된 하이퍼파라미터는 \red{Table XX}에 나타내있다.
% % 여기서는 모든 task들에 공통적으로 적용되는 GAE와 같은 하이퍼파라미터 값들을 나열한다. 
% % Table XX를 참조하기 바란다.
% 각 task에 대한 세부적인 hyper-paramter는 \red{Appendix XX}를 참고하기 바란다.

\subsection{Test-Time Guidance}
\label{appendix_subsec:test_time_guidance}

This section provides a detailed procedure for the proposed test-time guidance and describes the construction of the guidance loss $\loss_{\text{guide}}$. 
The loss is formulated as a weighted sum of one or more individual terms that promote physical plausibility, goal alignment, and scene awareness.

\textbf{The Proposed Test-time Guidance.}
% The proposed signal-space test-time guidance, which operates without backpropagation through the diffusion model.
%We propose a signal-space test-time guidance method that operates without backpropagation through the diffusion model.
Our signal-space test-time guidance is detailed in \cref{alg:diffusion_sampling_with_test_time_guidance}.
The overall sampling process follows that of \cite{tevet2025closd}, except for lines 5 to 12, where our method injects guidance signals directly into the samples. 
Inspired by \cite{huang2024constrained}, which demonstrated strong performance when applying guidance iteratively, our update rule in \cref{eq_x_0_update} is repeated $J$ times. 
We set $J = 5$ and $\eta = 1$ for the obstacle avoidance and HSI tasks.
We describe the individual loss terms that compose $\loss_{\text{guide}}$, \ie, $\{\mathcal{L}_{\text{guide},n}\}_{n=1}^N$ in~\cref{alg:diffusion_sampling_with_test_time_guidance} as follows.

\textbf{Collision Detection.}
For obstacle avoidance and HSI tasks, we adopt a collision detection loss used in \cite{rempe2023trace,rempe2022generating}. 
This loss computes the minimum distance between the points in the overlapping and non-overlapping regions of the obstacle and the humanoid.
Optimization encourages maximizing this minimum distance.
The obstacle region is represented by a walkable map $\boldsymbol{\mathcal{W}}_{\text{walk}} \in \{0,1\}^{\text{Height} \times \text{Width}}$, where 0 indicates free space and 1 indicates obstacles.
% \todo{Height, Width 정의 설명}
$\text{Width}$ and $\text{Height}$ denote the width and height of the 2D rectangular area within which the agent is allowed to navigate.
Let $\boldsymbol{x}_{\text{plan},k_{\text{pelvis}}}^h$ be the $k_{\text{pelvis}}$-th joint position, and $k_{\text{pelvis}}$ denotes a pelvis joint index.
The humanoid region is modeled by a 10$\times$10 grid of points $\boldsymbol{\mathcal{G}}(\boldsymbol{x}_{\text{plan},k_{\text{pelvis}}}^h) \in \mathbb{R}^{10\times10\times2}$ around the pelvis joint~\cite{rempe2023trace}.
The minimum distance $d_{\text{min}}(\boldsymbol{x}_{\text{plan}}^h)$ is defined as:
\begin{equation} \label{appendix_eq:d_min}
\begin{split}
    &d_{\text{min}}(\boldsymbol{x}_{\text{plan}}^h) = \\
    &\min_{i \in \boldsymbol{\mathcal{W}}_{\text{grid}} \setminus \mathcal{I}_{\text{inter}},  j \in \mathcal{I}_{\text{inter}}} \| \boldsymbol{\mathcal{G}}(\boldsymbol{x}_{\text{plan},k_{\text{pelvis}}}^h)_{i} - \text{sg}( \boldsymbol{\mathcal{G}}(\boldsymbol{x}_{\text{plan},k_{\text{pelvis}}}^h)_{j}) \|_2,
\end{split} 
\end{equation}
where $\boldsymbol{\mathcal{W}}_{\text{grid}}$ is the 2D projected map of $\boldsymbol{\mathcal{G}}(\boldsymbol{x}_{\text{plan},k_{\text{pelvis}}}^h)$, $\mathcal{I}_{\text{inter}}=\boldsymbol{\mathcal{W}}_{\text{grid}} \cap \boldsymbol{\mathcal{W}}_{\text{walk}}$, and $\text{sg}(\cdot)$ denotes the stop-gradient operator.
Then, the corresponding loss is defined as:
\begin{equation} \label{appendix_eq:obj_obstacle_avoidance}
    \loss_{\text{Obst}} = 1 - \left (\frac{d_{\text{min}}(\boldsymbol{x}_{\text{plan}}^h)}{b}\right ),
\end{equation}
where $b$ is the diagonal length of the agent's grid points. 
We compute the total obstacle loss over an $H$-frame motion plan as $\sum_{h=1}^{H} \loss_{\text{Obst}}(\boldsymbol{x}_{\text{plan}}^h)$, which is used during test-time guidance as a collision detection.

\textbf{Inpainting Joint Position.}
To guide a specific joint $k \in \{1,\dots,K\}$ to reach a global target position $\mathcal{P}_{\text{target}}\in\mathbb{R}^3$ at the last frame $H$, we define:
\begin{equation} \label{eq:obj_inpaint_pos_last_frame}
    \loss_{\text{Pos}} = \| \boldsymbol{x}_{\text{plan},k}^{H} - \mathcal{P}_{\text{target}} \|_2^2.
\end{equation}
To ensure a smooth trajectory that gradually approaches the target, we also use frame-weighted $\loss_{\text{Pos}}$ as:
\begin{equation} \label{eq:obj_inpaint_pos_all_frames}
    \loss_{\text{WeightedPos}} = \sum_{h\in \{1, \dots, H\}} \frac{h}{H} \| \boldsymbol{x}_{\text{plan},k}^{h} - \mathcal{P}_{\text{target}} \|_2^2.
\end{equation}

\textbf{Inpainting Heading.}
To guide the humanoid's facing direction, we compute heading vectors as the cross product between the \emph{Z}-up normal vector and a linear combination of vectors from the left to the right hip and shoulder positions~\cite{guo2022generating}.
The heading estimation process is described in \cref{alg:inpaint_heading}.
\input{algorithm/estimation_heading_alg}
Given the estimated heading vector $\Tilde{\boldsymbol{u}}_{\text{head}}^{1:H} \in \mathbb{R}^{H\times2}$ and target heading $\boldsymbol{u}_{\text{head}}$, the loss is defined as:
\begin{equation} \label{eq:obj_heading}
    \loss_{\text{Head}} = \sum_{h \in \{ 1, \dots, H \}} \frac{h}{H} \left[ 1 - u_{\text{head}}^T \Tilde{\boldsymbol{u}}_{\text{head}}^{h} / \left( \| u_{\text{head}} \|_2 \| \Tilde{\boldsymbol{u}}_{\text{head}}^{h} \|_2 \right) \right],
\end{equation}
which promotes consistent orientation across frames toward the desired direction, such as for seated interactions.

\textbf{Smoothness.}
To promote motion continuity, we minimize the inter-frame displacement~\cite{schulman2014motion}:
\begin{equation} \label{eq:obj_smooth}
    \loss_{\text{Smooth}} = \|\boldsymbol{x}_{\text{plan}}^{2:H} - \boldsymbol{x}_{\text{plan}}^{1:H-1}\|_2^2
\end{equation}

\textbf{Jerk.}
Jerk, the third derivative of position, correlates with motion jitter~\cite{zhao2025dart}.
We follow \cite{barquero2024seamless,zhao2025dart} to approximate jerk via finite differences, and define:
\begin{equation*}
    \loss_{\text{Jerk}} =  \max_{k\in \{ 1, \dots, K\} } \|\ddot{\boldsymbol{x}}_{\text{plan},k}^{2:H-2} - \ddot{\boldsymbol{x}}_{\text{plan},k}^{1:H-3} \|_1,
\end{equation*}
where velocity and acceleration are approximated as:
\begin{equation} \label{appendix_eq:finite_difference}
\begin{split}
    &\dot{\boldsymbol{x}}_{\text{plan}}^{1:H-1} = \boldsymbol{x}_{\text{plan}}^{2:H} - \boldsymbol{x}_{\text{plan}}^{1:H-1} \\
    &\ddot{\boldsymbol{x}}_{\text{plan}}^{1:H-2} = \dot{\boldsymbol{x}}_{\text{plan}}^{2:H-1} - \dot{\boldsymbol{x}}_{\text{plan}}^{1:H-2}.
\end{split}
\end{equation}

\textbf{Wrist-to-head constraint.}
The humanoid often fails to sit on armchairs because their narrow width causes the armrests to interfere with the wrists during the sitting motion.
To alleviate this, we encourage lifting the wrists toward the head using:
\begin{equation}
    \loss_{\text{Hand-up}} = \sum_{k\in\{ k_{\text{lw}}, k_{\text{rw}} \}} \| \boldsymbol{x}_{\text{plan},k}^{1:H} - \text{sg}\left( \boldsymbol{x}_{\text{plan},k_{\text{head}}}^{1:H} \right) \|_2^2,
\end{equation}
where $k_{\text{head}}$, $k_{\text{lw}}$, and $k_{\text{rw}}$ denote the head, left wrist, and right wrist joints, respectively.
The stop-gradient operator prevents the head position from being updated during optimization in test-time guidance.
The specific combinations of individual loss functions are detailed in \cref{appendix_sec:task_description} across the different tasks.

\subsection{P2R Network Design}
\label{appendix_subsec:pos_to_rot_network}

% This section describes the dataset construction and training procedure of the $P2R$ network based on inverse kinematics. Inverse kinematics is an ill-posed problem that aims to estimate joint rotations from joint positions. The $P2R$ network employed in this work follows the analytical inverse kinematics framework introduced in~\cite{li2021hybrik}.
This section describes the dataset construction and training procedure for the $P2R$ network, which is based on inverse kinematics. Inverse kinematics is an ill-posed problem that seeks to estimate joint rotations from joint positions. The $P2R$ network used in this work adopts the analytical inverse kinematics framework introduced in~\cite{li2021hybrik}.

\textbf{Background.}
% Let $\mathcal{P}_k$ and $\mathcal{R}_k$ be global position and rotation for joint $k$.
Let $\mathcal{P}_k$ and $\mathcal{R}_k$ denote the global position and rotation of joint $k$, respectively.
% $\mathcal{R}_k = \mathcal{R}_{pa(k)} \mathcal{R}_{pa(k),k}$ is the global rotation of joint $k$, and $pa(k)$ denotes its parent joint. 
% $\mathcal{R}_{pa(k),k}$ represents the relative rotation of the joint $k$ with respect to its parent joint.
The global rotation $\mathcal{R}_k$ can be expressed as $\mathcal{R}_k = \mathcal{R}_{pa(k)} \mathcal{R}_{pa(k),k}$, where $pa(k)$ denotes the parent joint of $k$, and $\mathcal{R}_{pa(k),k}$ represents the relative rotation of joint $k$ with respect to its parent.
Following \cite{li2021hybrik}, we leverage a twist-and-swing decomposition to analytically recover 3-DoF joint rotations from a 1-DoF twist angle, given the SMPL~\cite{loper2015smpl} rest pose. 
% This decomposition factorizes a 3D rotation into a swing and a twist component: the swing rotation aligns two directional vectors, while the twist rotation rotates around a given axis. Formally, the relative rotation between joint $k$ and its parent can be written as
This decomposition factorizes a 3D rotation into two components: a swing rotation, which aligns two directional vectors, and a twist rotation, which rotates around a given axis. Formally, the relative rotation between joint $k$ and its parent can be written as
\begin{equation} \label{appendix_eq:twist_and_swing_rotations}
    \mathcal{R}_{pa(k),k} = \mathcal{D}^{sw}( \overrightarrow{v}_k,\overrightarrow{u}_k ) \mathcal{D}^{tw}(\overrightarrow{u}_k, \phi_k) = \mathcal{R}_k^{sw} \mathcal{R}_k^{tw},
\end{equation}
where $\overrightarrow{u}_k = \mathcal{P}_{\text{rest},k} - \mathcal{P}_{\text{rest},pa(k)}$ is the bone vector in the rest pose, $\overrightarrow{v}_k = \mathcal{P}_k - \mathcal{P}_{pa(k)}$ is the corresponding vector in the posed configuration, and $\phi_k$ is the twist angle around $\overrightarrow{v}_k$. The operators $\mathcal{D}^{sw}$ and $\mathcal{D}^{tw}$ compute the swing and twist rotations, respectively. Since the swing rotation can be derived analytically from the rest pose and current joint positions, estimating only the twist angle is sufficient to reconstruct the full 3-DoF rotation from joint positions.
%

\input{algorithm/p2r_dataset_alg}

\textbf{Dataset Construction.}
% To predict $\phi$ from global positions, we compute AMASS~\cite{mahmood2019amass} rotation data into twist angles to construct a $(\mathcal{P}, \phi)$ dataset.
% The detailed procedure is described in \cref{alg:p2r_dataset}.
To predict $\phi$ from global joint positions, we convert AMASS~\cite{mahmood2019amass} rotation data into twist angles, constructing a dataset of $(\mathcal{P}, \phi)$ pairs.
The detailed procedure is provided in \cref{alg:p2r_dataset}. 

\input{algorithm/p2r_training_alg}

\textbf{Training Position-to-Rotation Network.}
% $P2R$ is formulated as $P2R = \mathcal{D} \circ \Omega$, where a neural network $\Omega$ predicts twist angles from global positions and a function $\mathcal{D}$ converts the predicted twist angles into rotation matrices.
% The training procedure is summarized in \cref{alg:p2r_training}.
We define $P2R$ as a composition $P2R = \mathcal{D} \circ \Omega$, where the neural network $\Omega$ predicts twist angles from global positions, and the function $\mathcal{D}$ converts the predicted twist angles into rotation matrices.
The overall training procedure is summarized in \cref{alg:p2r_training}.
In line 2 of \cref{alg:p2r_training}, the variable $\mathcal{P}_{\text{root-rel}}$ denotes global joint positions transformed into root-relative coordinates. 
% Because global positions $\mathcal{P}$ can take on unbounded values, learning maybe difficult. To mitigate this, we preprocess the data using root-relative coordinates to constrain input magnitude. 
Since global positions $\mathcal{P}$ can take on unbounded values, learning becomes difficult. To mitigate this issue, we preprocess the data by converting it into root-relative coordinates, thereby constraining the input magnitude.
% $\mathcal{T}_{\text{rotmat26d}}$ is a function which maps a rotation matrix into a 6D rotation representation~\cite{zhou2019continuity}.
The function $\mathcal{T}_{\text{rotmat26d}}$ maps a rotation matrix to a 6D rotation representation~\cite{zhou2019continuity}.
We train the $P2R$ model for 200 epochs using the Adam optimizer~\cite{kingma2014adam} with a fixed learning rate of 0.01.

\section{More Details of Long-Term Tasks}
\label{appendix_sec:task_description}

% We provide additional details of setup and evaluation procedure of the goal-reaching, the obstacle avoidance and HSI tasks.
We provide additional details regarding the setup and evaluation procedure for the goal-reaching, obstacle avoidance, and HSI tasks.
% 그에 앞서, 사용한 세팅과 평가 과정에 대해서 설명하기 전에 \ours{}가 어떻게 subtask를 실행하는 지를 설명하고자 한다.
% Text-to-motion을 포함한 모든 long-term task들은 서로 같거나 다른 subtask들로 구성된 sequence를 수행하는 문제이다. 주어진 subtask를 완료하면 바로 next subtask의 condition이 모션 플래너에 입력이되는 구조를 따른다. 이러한 구조는 \ours{}는 물론 CLoSD도 동일하게 사용한다.
We first describe how \ours{} executes subtasks, as shown in \cref{appendix_fig:log_term_task_detail}.
All long-term tasks are formulated as sequences of subtasks, which may vary or remain identical throughout the sequence.
Once the agent completes a subtask, the condition for the next subtask is immediately provided to the motion planner~$G$.
If the agent fails to complete a subtask, it discontinues the current sequence and immediately begins a new one.
This sequential execution framework is applied to both \ours{} and CLoSD.
\input{figs/supp_long_term_task_detail}

% \subsection{Goal-reaching}
\subsection{Goal-Reaching and Obstacle Avoidance}
\label{appendix_subsec:goal_reach_description}

\begingroup
\setcounter{section}{3}   % D장이 section 3이라면
\setcounter{figure}{0}    % D.1부터 시작하고 싶다면
\renewcommand{\thefigure}{D.\arabic{figure}}  % Figure D.1 형식
\input{figs/visualization_goal_reaching}
\endgroup

\textbf{Goal-reaching.}
In the goal-reaching task, both \ours{} and CLoSD utilize a motion planner $G_{\text{text+target}}$ conditioned on both the text prompt and the target point. 
The target point is determined by the following process.
Let $\mathcal{P}_{\text{agent}}$ and $\mathcal{P}_{\text{target}}$ be the current humanoid position and the final destination position, respectively. 
We define $\mathcal{P}_{\text{agent}}$ as the position of the pelvis joint.
$\mathcal{P}_{\text{target}}$ is randomly determined such that:
\begin{equation*}
    \text{scale} \cdot 1 \leq \|\mathcal{P}_{\text{target}}-\mathcal{P}_{\text{agent}}\|_2 \leq \text{scale} \cdot 3,
\end{equation*}
%
% where a scale lies in the range of 1 to 100.
where the scale lies in the range from 1 to 100.
After determining $\mathcal{P}_{\text{target}}$, we follow the reach task procedure of CLoSD~\cite{tevet2025closd} by selecting a point along the straight line connecting $\mathcal{P}_{\text{agent}}$ and $\mathcal{P}_{\text{target}}$ that lies within 1.2 meters from $\mathcal{P}_{\text{agent}}$, and use it as the condition for the goal-reaching task.
We refer to this point as the intermediate target point $\mathcal{P}_{\text{int-target}}$. 
The intermediate target point is repeatedly sampled until $\|\mathcal{P}_{\text{target}}-\mathcal{P}_{\text{agent}}\|_2 \leq 1.2\,\mathrm{m}$.
% This setup is applied consistently to both CLoSD and \ours{}.
Under this setup, the humanoid reaches a specified goal position from a given start position using one of four locomotion styles, \emph{walk}, \emph{walk fast}, \emph{jog slow}, and \emph{jog}.
% The text descriptions for \emph{walk}, \emph{walk fast}, \emph{jog slow}, and \emph{jog} are ``A person is walking.'', ``A person is walking fast.'', ``A person is jogging slowly.'', and ``A person is jogging.'', respectively. 
The corresponding text descriptions for these styles are ``\emph{A person is walking.}'', ``\emph{A person is walking fast.}'', ``\emph{A person is jogging slowly.}'', and ``\emph{A person is jogging.}'', respectively.
The goal-reaching task is considered successful if the agent reaches the target point~$\mathcal{P}_{\text{target}}$ within the designated time limit~\cite{tevet2025closd}. The agent is deemed to have reached the target point when the distance between the pelvis joint~$\mathcal{P}_{\text{agent}}$ and the target point is less than or equal to 0.3 meters.
By default, the time limit for goal-reaching tasks with distances ranging from 1 to 3 meters is set to 500 frames~\cite{tevet2025closd}. This time limit increases proportionally with the scale. For instance, when the scale is set to 100, the corresponding time limit becomes 50,000 frames.

\begingroup
\setcounter{section}{3}   % D장이 section 3이라면
\setcounter{figure}{1}    % D.1부터 시작하고 싶다면
\renewcommand{\thefigure}{D.\arabic{figure}}  % Figure D.1 형식
\input{figs/visualization_obstacle_avoidance}
\endgroup

\textbf{Obstacle Avoidance.}
% \subsection{Obstacle Avoidance}
% \label{appendix_subsec:oa_description}
% \input{figs/visualization_obstacle_avoidance}
% In the obstacle avoidance task, both \ours{} and CLoSD utilize a motion planner $G_{\text{text+target}}$ conditioned on both the text prompt and the target point.
% The setup is similar to the long-term reach task, which is to reach a specified goal position from a given start position using one of four locomotion styles, \emph{walk}, \emph{walk fast}, \emph{jog slow}, and \emph{jog}, except that obstacles are placed along the straight-line path to the goal.
For the obstacle avoidance task, we follow the same setup as the goal-reaching task, except that obstacles are placed along the straight-line path to the goal.

%\textbf{Test-time Guidance.}
Obstacles are randomly placed pillars and walls in the scene, following the protocol of \cite{rempe2023trace, rudin2022learning}.
% We use the walkable map provided by the environment to specify a region within a humanoid can move.
We use the walkable map provided by the environment to specify a region within which the humanoid can move.
The guidance loss $\loss_{\text{guide}}$ only uses $\loss_{\text{Obst}}$ in \cref{appendix_eq:obj_obstacle_avoidance} with its weight $w_{\text{guide}}$ set to 1.
To improve navigating performance in narrow spaces composed of dense obstacles, we use $d_{\text{min}}$ not only for the pelvis joint but also for both wrist joints.

% \textbf{Optimizing condition at test-time.}
%\textbf{Optimization of Test-time Conditions.}
We observed that applying the collision loss directly to an intermediate target point in the condition input yields better results when used alongside test-time guidance. The details of this optimization procedure are provided in \cref{alg:optimization_target_point}. 
The optimization procedure is carried out for 200 steps using the Adam optimizer~\cite{kingma2014adam}, with a fixed step size of 0.01. 
The initialization involves setting $d_{\text{int-target}}$ to 1.
\input{algorithm/optimization_target_using_collision_loss_alg}

\subsection{Human-Scene Interaction}
\label{appendix_subsec:hsi_description}
\begingroup
\setcounter{section}{3}   % D장이 section 3이라면
\setcounter{figure}{2}    % D.1부터 시작하고 싶다면
\renewcommand{\thefigure}{D.\arabic{figure}}  % Figure D.1 형식
\input{figs/hsi_further_qualitative_res}
\endgroup

\textbf{Room Visiting Order and Challenges.}
The humanoid follows a predefined scene plan, where it is initialized outside the nine-room environment as shown in~\cref{appendix_fig:fig_supp_scene_process}(a), and is required to sequentially visit each room, interact with objects inside, and then return to its initial position.

The humanoid is initialized outside \emph{Room3}. 
We manually set the visitation order as follows: \emph{Room1}, \emph{Room2}, \emph{Room3}, \emph{Room4}, \emph{Room5}, \emph{Room6}, \emph{Room7}, \emph{Room8}, and \emph{Room9}, performing interactions in each room. 
Upon completing the \emph{Room9} interactions, it returns by visiting \emph{Room9}, \emph{Room8}, \emph{Room6}, \emph{Room4}, and \emph{Room3} in reverse, ending outside \emph{Room3}.

Several subtasks frequently lead to failure.
In \emph{Room1}, the second \emph{SIT} subtask uses a sofa with a low seat height (0.3 m), often causing failure.
In \emph{Room2}, the second \emph{SIT} subtask uses an armchair with narrow armrests that obstruct the humanoid's arms.
After finishing \emph{Room8}, the \emph{REACH} subtask toward \emph{Room9} often fails due to the doorway obstructing the humanoid's path. This is attributed to low collision scores for the doorframe, resulting in suboptimal A* trajectories. 
Details of the scene plan generation are described in~\cref{appendix_sec:scene_plan}.

\textbf{Walkable Map Construction.}
To construct the walkable map, we projected the surface points of all objects contained in the scene onto the \emph{XY}-plane under \emph{Z}-up coordinate system.
Specifically, we sampled 100,000 points across object surfaces, weighting the face normals of object against the up-axis to prioritize horizontal surfaces. 
Let $\boldsymbol{\mathcal{W}}_{\text{walk}} \in \{0, 1\}^{\text{Width} \times \text{Height}}$ be initialized with all entries set to zero.
The sampled points $\{ \mathcal{P}_{i} \}_{i=1}^{10^5}$ were then discretized into pixels, and the corresponding positions in the walkable map $\boldsymbol{\mathcal{W}}_{\text{walk},i}$ were set to one.
For HSI task, the $\text{Width}$ and $\text{Height}$ of the walkable map are set to 360 and 400, respectively.

\textbf{Path planning using A* algorithm.}
\emph{REACH} actions in HSI are executed in two stages. First, we apply the A* algorithm to compute a path that avoids obstacles in the cluttered indoor environment. Second, we select the point on the planned path approximately one meter ahead of the humanoid as the target point, which is then used as part of the condition for the diffusion model $G$. 
Let $x_{\text{curr}}$, $x_{\text{next}}$, and $x_{\text{goal}}$ denote 2D pixel coordinates on the \emph{XY}-plane, corresponding to the current point, the candidate point under exploration, and the final target point, respectively.
The cost function $\mathcal{C}$ and heuristic function $\mathcal{H}$ for the A* algorithm are defined as:
\begin{equation}
\begin{split}
    &\mathcal{C}(x_{\text{curr}},x_{\text{next}}) = \| x_{\text{curr}} - x_{\text{next}} \|_2 + \loss_{\text{Obst}}(x_{\text{next}}) \\
    &\mathcal{H}(x_{\text{next}},x_{\text{goal}}) = \| x_{\text{next}} - x_{\text{goal}} \|_2 + 10 \loss_{\text{Obst}}(x_{\text{next}}),
\end{split}
\end{equation}
where the weight of 10 in the heuristic term is chosen heuristically to account for unknown future collisions.

\textbf{Details of Subtasks.}
We describe the detailed setup for each subtask in the HSI task. To reflect the unique characteristics and constraints of each subtask, we note that both the type of guidance loss terms ($\loss_{\text{guide}}$) and the associated target joints differ across subtasks.
We set the time limit for all subtasks to 500 frames.

1) \emph{REACH}.
The condition $c$ for the \emph{REACH} subtask consists of a walking-style text prompt and a target position for the pelvis joint. The text is sampled from either ``\emph{A person is walking.}'' or ``\emph{A person is walking fast.}'', corresponding to \emph{walk} and \emph{walk fast}, respectively.
The loss terms used for test-time guidance are $\loss_{\text{Obst}}$, $\loss_{\text{Smooth}}$, and $\loss_{\text{Jerk}}$, with respective weights of 1, 1000, and 1.
The success criterion follows \cite{tevet2025closd}: the distance between the pelvis and the target is less than 0.3 meters.

2) \emph{SIT}. The condition $c$ for the \emph{SIT} subtask includes the text ``\emph{A person is sitting down on a bench.}'' and the pelvis target position. 
The target is placed near the midpoint of the sitting object, adjusted by the link length between the humanoid's knees and hips. 
The loss terms used for test-time guidance are $\loss_{\text{Pos}}$, $\loss_{\text{Head}}$, $\loss_{\text{Smooth}}$, and $\loss_{\text{Jerk}}$, with corresponding weights of 10, 1, 1, and 1. In the case of an armchair, $\loss_{\text{Hand-up}}$ is additionally incorporated with a weight of 100.
The success criterion is based on the pelvis height falling within a predefined range, as in \cite{tevet2025closd}.

3) \emph{GETUP}. The condition $c$ for the \emph{GETUP} subtask includes the text ``\emph{A person is getting up from a bench and standing.}'' and the pelvis target position. 
This task is only performed after successful completion of the \emph{SIT} subtask. The loss terms used for test-time guidance are $\loss_{\text{Pos}}$, $\loss_{\text{Head}}$, $\loss_{\text{Smooth}}$, and $\loss_{\text{Jerk}}$, with respective weights of 1, 1, 10, and 1.
The success criterion matches that of \emph{REACH} subtask.

4) \emph{TOUCH}. The condition $c$ for the \emph{TOUCH} subtask includes the text ``\emph{A person is touching with the right hand}'' and the right wrist target position. 
Only the right hand is used for fair comparison with \cite{xiao2024unified}. Unlike other tasks with fixed targets, the target for \emph{TOUCH} is sampled randomly within a 0.1\,$\mathrm{m}$ radius around a predefined surface point $\mathcal{P}_{\text{touch}}$.
The loss terms used for test-time guidance are $\loss_{\text{Obst}}$, $\loss_{\text{WeightedPos}}$, $\loss_{\text{Head}}$, $\loss_{\text{Smooth}}$, and $\loss_{\text{Jerk}}$, with corresponding weights of 1, 50, 1, 1, and 0.1.
$\loss_{\text{Obst}}$ is computed only for the pelvis joint, consistent with the obstacle avoidance task.
Success is defined as the right wrist being within a 0.1\,$\mathrm{m}$ of the target.
\begingroup
\setcounter{section}{3}   % D장이 section 3이라면
\setcounter{figure}{3}    % D.1부터 시작하고 싶다면
\renewcommand{\thefigure}{D.\arabic{figure}}  % Figure D.1 형식
% \input{figs/hsi_further_qualitative_res}
\input{figs/failure_cases}
\endgroup

\section{More Visualizations}
\label{appendix_sec:visualization}

We provide additional visualizations for various long-terms tasks in~\cref{appendix_fig:visualization_goal_reach,appendix_fig:visualization_oa,appendix_fig:hsi_further_qualitative_res} and for failure cases in~\cref{appendix_fig:failure_cases}, respectively.

\setcounter{section}{4}   % D장이 section 3이라면
\setcounter{figure}{0}    % D.1부터 시작하고 싶다면
\section{Scene Plan Generation}
\label{appendix_sec:scene_plan}
\input{figs/supp_scene_process}

\subsection{Scene Configuration}
\label{appendix_subsec:detail_scene_description}
\cref{appendix_fig:fig_supp_scene_process} shows the overall procedure of the scene plan generation.
We use the largest indoor environment, ``ProcTHOR-Val-615'', from the ProcTHOR-10K~\cite{procthor} dataset, which consists of nine rooms and 202 objects.
Objects in the scene are described in a YAML-style format in~\cref{table_supp_scene_yaml}.
Their common properties are given as \emph{Default Object Configuration}, each specified by a transformation matrix and a mesh path.
% This category includes all physical objects in the scene, each specified by a transformation matrix and a mesh path.
% Our selected scene includes 202 such objects.
% 2) \emph{Interactive Objects}. A subset of 24 objects that are interactable by the humanoid. These are manually selected to support four types of subtasks: \emph{SIT} (e.g., armchair, sofa, toilet, bed), \emph{TOUCH} (e.g., television, table, countertop, fridge, sink), \emph{REACH}, and \emph{GETUP}. The configuration also specifies 3D spatial extents, target points. 
\emph{Interactive Objects Configuration} specifies 24 objects that the humanoid can interact with.
These objects are used for each of the four subtasks, \emph{SIT} (e.g., armchair, sofa, toilet, bed), \emph{TOUCH} (e.g., television, table, countertop, fridge, sink), \emph{REACH}, and \emph{GETUP}.
The configuration also specifies 3D spatial bounding boxes corresponding to these objects and target points for interaction with the humanoid.
Finally, \emph{Room Configuration} defines the spatial layout of the nine rooms, including the coordinate transforms of room boundaries and the door positions.
All original configurations provided by ProdTHOR~\cite{procthor} use a \emph{Y}-up coordinate system, while we adopt a \emph{Z}-up system for physics simulation environment.
Additional transformation steps are applied to align the coordinate systems.
%
\input{tables/scene_interaction_task/supp_scene_yaml}

%
\subsection{Structure of Scene Plan}

The scene plan is a sequence of subtasks, where each subtask is specified by a text prompt describing the intended action and a target position corresponding to the joint, as detailed in \cref{appendix_subsec:hsi_description}.
\input{tables/scene_interaction_task/supp_scene_plan_yaml}

The scene plan generation consists of two steps.
First, we generate a room-wise scene plan by specifying a starting point for the agent and guiding it to circulate the room while interacting with nearby objects. Then, the room-wise plans are concatenated into a single long-term subtask sequence.
We use a predefined room visiting order.
We used ChatGPT-4o~\cite{achiam2023gpt} as an LLM to generate individual room plans.

\textbf{Room-wise Scene Plan Generation.}
To generate each room-wise scene plan, we provide the LLM with a text prompt, the room's configuration, and a top-view image capturing the full layout of the room.
We use a prompt similar to that in~\cite{xiao2024unified}.
As detailed in \cref{appendix_subsec:detail_scene_description}, the configuration includes transformation parameters for all interactive objects in the room.
The top-view image offers spatial context regarding object placements, allowing the LLM to generate more realistic plans. 
We repeat the process of the scene plan generation using the LLM for each of the room-wise configurations according to the predefined room visiting plan.

\textbf{Input Prompt to LLM.}
\cref{fig_supp_llm_input} depicts a text prompt as input to the LLM, encouraging the LLM to generate output text of three sections: scene description, scenario, and room-wise scene plan, as shown in~\cref{table_supp_prompt_stages}.
%The output of each stage is generated in a single step, and examples of the results at each stage are shown in \cref{table_supp_prompt_stages}.
\input{tables/scene_interaction_task/supp_llm_stages}
The scene description section enables the LLM to be aware of the spatial layout of the room using the given room configuration and top-view image as input.
As a result, LLM generates the relative spatial relationships among the interactive objects in the room.
The scenario section encourage the LLM to leverage the previously generated scene description so that the scenario can be grounded in spatial information.
Finally, the room-wise scene plan is used to generate a room-wise scene plan, which is expressed as a sequence of subtasks such as \emph{REACH}, \emph{SIT}, \emph{GETUP}, and \emph{TOUCH} to be used for input to the diffusion motion planner of \ours.
\input{figs/supp_llm_input}
%An example of the prompt can be found in \cref{fig_supp_llm_input}.

\textbf{Long-term Motion Sequence Generation.}
\cref{table_supp_scene_plan_yaml} is an example of a YAML-style LLM output, which consists of three sections \emph{Room Visiting Plan}, \emph{Sequence of Object Interaction}, and \emph{Object-wise Subtask Sequence}.
A single long-term sequence is simply generated by combining each of entries in these sections, resulting in a sequence of the four subtasks.

\section{Limitations}
\label{subsec_limitation}

\ours{} performs frame-wise local optimization during test-time guidance, which may limit its ability to plan over long-horizon requiring global path awareness.
If the distribution of subtasks in a long sequence is imbalanced, the adaptation procedure may cause the policy to overfit to frequent subtasks, reducing generalization to rare ones.
In addition, simulation failures early in the sequence, especially on difficult subtasks, can compound and degrade performance in later stages due to the autoregressive nature of execution.

%% file: tables/bric_task_specific_hyperparameters.tex
\begin{table}[t]
    \centering
    \resizebox{1.0\linewidth}{!}{%
        \begin{tabular}{l cc cc} \hline \toprule
            Task & Text-to-Motion & Goal-Reaching & Obstacle Avoidance & HSI
            \\ \hline
            
            Epochs & 2000 & 4000 & 2000 & 4000
            \\
            
            $\lambda_{\text{CF}}$ & 1 & 1 & 1 & 1
            \\
            
            $\lambda_{\text{Robust}}$ & $1\times 10^{-4}$ & $1\times 10^{-4}$ & $1\times 10^{-5}$ & $1\times 10^{-5}$ 
            \\
            
            $\alpha$ & 0.999 & 0.999 & 0.999 & 0.999
            \\ \bottomrule \hline
        \end{tabular}
    }
    \caption{
        \textbf{Task-specific hyperparameters for \ours{}.} 
    }
    \label{appendix_tab:bric_hparams}
\end{table}

%% file: algorithm/test_time_adapation_alg.tex
\begin{algorithm}[t!]
\caption{Adapting policy via \ours{}}\label{alg:adaptation_policy}
% \hspace*{\algorithmicindent} \textbf{Input}: Motion planner $G$, condition $c$, RL-based policy $\pi$, physics simulation $\Lambda$, position-to-rotation network $P2R$  \\
\hspace*{\algorithmicindent} \textbf{Input}: Motion planner $G$ \\
\hspace*{\algorithmicindent} \quad \quad \quad Condition $c$ \\
\hspace*{\algorithmicindent} \quad \quad \quad RL-based policy $\pi$ \\
\hspace*{\algorithmicindent} \quad \quad \quad Physics simulation $\Lambda$ \\
\hspace*{\algorithmicindent} \quad \quad \quad Position-to-rotation network $P2R$ \\
\hspace*{\algorithmicindent} \textbf{Output}: Adapted policy $\pi$
\begin{algorithmic}[1]
    %\State $\boldsymbol{x}_{\text{sim}}=\mathcal{T}_{\text{kine-to-sim}}(\boldsymbol{x}_{\text{kine}})$\ \Comment{$\boldsymbol{x}_{\text{kine}} \gets \hat{\boldsymbol{x}}_0$}
    %\State $\boldsymbol{x}_{\text{sim}}=M2N(\boldsymbol{x}_{\text{sim}})$ \Comment{$\boldsymbol{x}_{\text{sim}} \in \mathbb{R}^{K \times d_{\text{sim}}}$} 

    % \State $\boldsymbol{x}_{\text{sim}}=\mathcal{T}_{\text{kine2sim}}(\boldsymbol{x}_{\text{kine}})$ \Comment{$\boldsymbol{x}_{\text{sim}} \in \mathbb{R}^{K \times d_{\text{sim}}}$} 
    %\State $\boldsymbol{x}_{\text{sim}}=\mathcal{T}_{\text{kine-to-sim}}(\boldsymbol{x}_{\text{kine}})$
    % \Comment*[r]{inverse kinematics by HML3D}

    \State $h_{\text{global}} = 0$
    \For{epoch=1, $\dots$, Epochs}

    % \State Sample $\boldsymbol{x}_{0}^{1:H}$ conditioned on $c$. \Comment{See \cref{alg:diffusion_sampling_with_test_time_guidance}}
    % \State Transform motion to global position, $\boldsymbol{x}_{\text{plan}}^{1:H}=\Psi(\boldsymbol{x}_0^{1:H})$
    % \State $\boldsymbol{x}_{\text{plan}}^{1:H}=\Psi(\boldsymbol{x}_0^{1:H})$
    % \State \textbf{경험 버퍼 구축}
    \State $\mathcal{B}_{\text{PPO}} = \{ \emptyset \}$, $\mathcal{B}_{\text{AMP}} = \{ \emptyset \}$
    % \For{$h =1, \dots,H$}
    % \State $//$ 경험 버퍼 구축
    \Statex \hspace*{\algorithmicindent} // 32 is a horizon of PPO.
    \For{$i = 1, \dots,32$} 
        % \Comment{32 is a horizon of PPO}
        % \State $//$ 경험 버퍼 구축
        \If{$h_{\text{global}} \bmod H = 0$}
            \Statex \hspace*{\algorithmicindent} \hspace*{2em} // See \cref{alg:diffusion_sampling_with_test_time_guidance}.
            \State Sample $\boldsymbol{x}_{0}^{1:H}$ conditioned on $c$
            % ~(See \cref{alg:diffusion_sampling_with_test_time_guidance}). 
            % \Comment{See \cref{alg:diffusion_sampling_with_test_time_guidance}}
            \State $\boldsymbol{x}_{\text{plan}}^{1:H}=\Psi(\boldsymbol{x}_0^{1:H})$
        \EndIf
        \State $h=h_{\text{global}} \bmod H + 1$
        \State $\boldsymbol{s}_{\text{g}}^{h-1}= \boldsymbol{x}_{\text{plan}}^{h} \ominus \boldsymbol{s}_{\text{p}}^{h_{\text{global}}} $
        % \State $\boldsymbol{s}^h=(\boldsymbol{s}_{\text{p}}^{h-1}, \boldsymbol{s}_{\text{g}}^{h-1})$
        \Statex \hspace*{\algorithmicindent} \hspace*{1em} // $\boldsymbol{s}^{h-1}=(\boldsymbol{s}_{\text{p}}^{h_{\text{global}}}, \boldsymbol{s}_{\text{g}}^{h-1})$.
        \State $\boldsymbol{a}^{h-1} \sim \pi(\boldsymbol{s}^{h-1})$ 
        % \Comment{$\boldsymbol{s}^{h-1}=(\boldsymbol{s}_{\text{p}}^{h_{\text{global}}}, \boldsymbol{s}_{\text{g}}^{h-1})$}
        \Statex \hspace*{\algorithmicindent} \hspace*{1em} // Simulation.
        \State $\boldsymbol{s}_{\text{p}}^{h_{\text{global}}+1}=\Lambda (\boldsymbol{s}_{\text{p}}^{h_{\text{global}}}, \boldsymbol{a}^{h-1})$ 
        % \Comment{Simulation}
        \State $r^{h-1}= 0.5r_{\text{imitation}}^{h-1} + 0.5r_{\text{amp}}^{h-1} +r_{\text{energy}}^{h-1}$
        % \State $//$ 모션 계획을 정책에 대한 상태로 변환
         % \todo{주석기호 추가}\Statex \hspace*{3em} \textbf{--- Start Construction Planned Policy State ---}
        \Statex \hspace*{\algorithmicindent} \hspace*{1em} // Construction planned policy state.
        \State $\mathcal{P}_{\text{plan}}^h = \boldsymbol{x}_{\text{plan}}^{h}$
        \State $\mathcal{R}_{\text{plan}}^h=P2R(\mathcal{P}_{\text{plan}}^h)$ 
        % \Comment{$\mathcal{P}_{\text{plan}}^h = \boldsymbol{x}_{\text{plan}}^{h}$}
        \State $\boldsymbol{s}_{\text{plan-p}}^{h-1} = (\mathcal{P}_{\text{plan}}^{h-1},\mathcal{R}_{\text{plan}}^{h-1},\dot{\mathcal{P}}_{\text{plan}}^{h-1},\dot{\mathcal{R}}_{\text{plan}}^{h-1})$
        \State $\boldsymbol{s}_{\text{plan-g}}^{h-1}=(\mathcal{P}_{\text{plan}}^{h}-\mathcal{P}^{h-1}, \dot{\mathcal{P}}_{\text{plan}}^h - \dot{\mathcal{P}}^{h-1},\mathcal{P}_{\text{plan}}^{h} )$
        \State $\boldsymbol{s}_{\text{plan}}^{h-1} = (\boldsymbol{s}_{\text{plan-p}}^{h-1},\boldsymbol{s}_{\text{plan-g}}^{h-1})$
         % \todo{주석기호 추가}\Statex \hspace*{3em} \textbf{--- End Construction Planned Policy State ---}
        % \State $//$ 경험 버퍼 업데이트
        \State $\mathcal{B}_{\text{PPO}} \leftarrow \mathcal{B}_{\text{PPO}} \cup (\boldsymbol{s}^{h-1}, \boldsymbol{a}^{h-1}, r^{h-1}, \boldsymbol{s}^{h}, \boldsymbol{s}_{\text{plan}}^{h-1})$
        % \If{some condition is true}
        \Statex \hspace*{\algorithmicindent} \hspace*{1em} // $\boldsymbol{\tau}_{\text{stand}} = \{\boldsymbol{s}_{\text{stand}}\}^{10}$
        \Statex \hspace*{\algorithmicindent} \hspace*{1em} // $\boldsymbol{\tau} = [\boldsymbol{s}_{\text{amp}}^{h_{\text{global}}-9}, \dots, \boldsymbol{s}_{\text{amp}}^{h_{\text{global}}}]^T$
        \State $\mathcal{B}_{\text{AMP}} \leftarrow \mathcal{B}_{\text{AMP}} \cup (\boldsymbol{\tau}_{\text{stand}}, \boldsymbol{\tau})$
        % \State $\mathcal{B}_{\text{AMP}} \leftarrow \mathcal{B}_{\text{AMP}} \cup [\boldsymbol{s}_{\text{amp}}^{h_{\text{global}}-9}, \dots, \boldsymbol{s}_{\text{amp}}^{h_{\text{global}}}]^T$
        \State $h_{\text{global}} \leftarrow h_{\text{global}}+1$
        % \EndIf
    \EndFor
    
    % \State $\mathcal{M}_{\text{plan}}=[\boldsymbol{s}_{}^{1}, \dots, \boldsymbol{s}^{H}]$ 
    % \State $\mathcal{M}_{\text{exec}}=[\boldsymbol{s}^{1}, \dots, \boldsymbol{s}^{H}]$ 

    % \State $\theta_V \leftarrow \arg\min_{\theta_V} \sum_{h=1}^H (V(\boldsymbol{s}_p^h ;\theta_V)-r^h)^2 + \lambda_{\text{CF}}(V(\boldsymbol{s}_p^h;\theta_V)-V(\boldsymbol{s}_p^h;\theta_V^{'}))^2$ \Comment{$(V(\boldsymbol{s}_p^h;\theta_V)-V(\boldsymbol{s}_p^h;\theta_V^{'}))^2$ is a component of $\loss_{\text{CF}}$}

    % \State $\theta_{\pi} \leftarrow \arg\min_{\theta_{\pi}} \sum_{h=1}^{H} \min \left(\frac{\pi_{\text{PHC}}(\boldsymbol{a}^h|\boldsymbol{s}^h;\theta_{\pi}^{\text{old}})}{\pi_{\text{PHC}}(\boldsymbol{a}^h|\boldsymbol{s}^h;\theta_{\pi})} \mathcal{A}^h, \text{clip}(\frac{\pi_{\text{PHC}}(\boldsymbol{a}^h|\boldsymbol{s}^h;\theta_{\pi}^{\text{old}})}{\pi_{\text{PHC}}(\boldsymbol{a}^h|\boldsymbol{s}^h;\theta_{\pi})}, 1-\epsilon_{\text{clip}},1+\epsilon_{\text{clip}}) \right) + \lambda_{\text{CF}} \| \mu(\boldsymbol{s}^h;\theta_{\pi}) - \mu(\boldsymbol{s}^h;\theta_{\pi}^{'}) \|_2^2$

    % \State $\theta_{D} \leftarrow \arg\min_{\theta_D} -\log D(\tau_{\text{standing}}) + \mathbb{E}_{\mathcal{M}_{\text{plan}}} \left[ -\log (1-D(\tau)) + \lambda_{\text{CF}}(D(\tau ;\theta_D)-D(\tau;\theta_D^{'}))^2 \right]$

    % \State $\theta_{V}^{'} \leftarrow \alpha \theta_{V}^{'} + (1-\alpha)\theta_{V}$
    % \State $\theta_{\pi}^{'} \leftarrow \alpha \theta_{\pi}^{'} + (1-\alpha)\theta_{\pi}$
    % \State $\theta_{V}^{'} \leftarrow \alpha \theta_{D}^{'} + (1-\alpha)\theta_{D}$

    \Statex \hspace*{\algorithmicindent} // $\mathcal{B}_{\text{PPO}}$ and $\mathcal{B}_{\text{AMP}}$ are used.
    % \todo{주석표시 방법을 통일할 것}
    \State $\loss_{\text{PPO}} = \loss_{\pi} + \loss_{V} + \loss_{D}$
    \State $\loss_{\text{PPO-TTA}} = \loss_{\text{PPO}} + \lambda_{\text{CF}} \loss_{\text{CF}} + \lambda_{\text{Robust}} \loss_{\text{Robust}}$
    \State Update $\theta_{\pi},\theta_{V},\theta_{D}$ to minimize $\loss_{\text{PPO-TTA}}$
    % \State Update $\theta_{\pi},\theta_{V},\theta_{D}$ to minimize $\mathbb{E}_{\mathcal{B}_{\text{PPO}},\mathcal{B}_{\text{PPO}}} \left[\loss_{\text{PPO-TTA}} \right]$
    \State Update $\theta_{\pi}', \theta_{V}', \theta_{D}'$ using EMA.
    \EndFor
\end{algorithmic}
\end{algorithm}

%% file: tables/ppo_hyperparameters.tex
\begin{table}[t]
    \centering
    \resizebox{0.9\linewidth}{!}{%
        % \begin{tabular}{ccc ccc} \hline \toprule
        %     & Batch Size & Learning Rate & $\sigma_{\pi}$ & $\gamma$ & $\epsilon_{\text{clip}}$
        %     \\ \hline
            
        %     Value & \red{XX.XX} & $2.5 \times 10^{-5}$ & \red{XX.XX} & \red{XX.XX} & \red{XX.XX}
        %     \\ \hline
            
        %     & $w_{\text{jp}}$ & $w_{\text{jr}}$ & $w_{\text{jv}}$ & $w_{\text{j$\omega$}}$ & 
        %     \\ \hline
            
        %     Value & \red{XX.XX} & \red{XX.XX} & \red{XX.XX} & \red{XX.XX} & \red{XX.XX}
        %     \\ \bottomrule \hline
        % \end{tabular}
        % \begin{tabular}{l | c} \hline \toprule
        %     Hyperparameter & Value \\ \hline
        %     Batch size & 32768 \\
        %     Learning rate & $2.5 \times 10^{-5}$ \\
        %     $\gamma$ & 0.99 \\
        %     $\epsilon_{\text{clip}}$ & 0.2
        %     \\ \bottomrule \hline
        % \end{tabular}
        \begin{tabular}{c | cccc} \hline \toprule
            Hyperparameter & Batch size & Learning rate & $\gamma$ & $\epsilon_{\text{clip}}$ \\ \hline
            Value & 32768 & $2.5 \times 10^{-5}$ & 0.99 & 0.2
            \\ \bottomrule \hline
        \end{tabular}
    }
    \caption{
        \textbf{Hyperparameters for PPO.} 
    }
    \label{appendix_tab:ppo_hparams}
\end{table}

%% file: algorithm/diffusion_sampling_with_test_time_guidance_alg.tex
%-------------------------------------------------------------------------
% \RestyleAlgo{ruled}
%% This is needed if you want to add comments in
%% your algorithm with \Comment
% \SetKwComment{Comment}{/* }{ */}
\begin{algorithm}
\caption{Diffusion sampling with test-time guidance}\label{alg:diffusion_sampling_with_test_time_guidance}
% \hspace*{\algorithmicindent} \textbf{Input}: Diffusion model $G$, condition $c$, loss function set $\{\mathcal{L}_{\text{guide},n}\}_{n=1}^N \subset C^{1}(\mathbb{R}^{H\times 24\times3}, \mathbb{R})$, weights $\{ w_{\text{guide},n} \}_{n=1}^N$ \\
\hspace*{\algorithmicindent} \textbf{Input}: Diffusion model $G$\\
\hspace*{\algorithmicindent} \quad \quad \quad Condition $c$ \\
\hspace*{\algorithmicindent} \quad \quad \quad Loss function set $\{\mathcal{L}_{\text{guide},n}\}_{n=1}^N$ \\
\hspace*{\algorithmicindent} \quad \quad \quad Weight set $\{ w_{\text{guide},n} \}_{n=1}^N$ \\
\hspace*{\algorithmicindent} \textbf{Output}: Motion plan $\boldsymbol{x}_{\text{plan}}^{1:H}$
\begin{algorithmic}[1]
    % \State \textbf{Motion Planner Adaptation}
    \Statex // $N$ is the number of the objectives for guidance.
    \State $\boldsymbol{x}_T^{1:H} \sim \mathcal{N}(\boldsymbol{0},\boldsymbol{\mathrm{I}})$
    \State $\boldsymbol{\epsilon} \sim \mathcal{N}(\boldsymbol{0},\boldsymbol{\mathrm{I}})$
    \For{$t =T, \dots,1$}\
        % \State $\boldsymbol{x}_0^{1:H}=G(\boldsymbol{x}_t^{1:H},\boldsymbol{x}_t^{\text{past}},t,\emptyset)-s\{ G(\boldsymbol{x}_t^{1:H},\boldsymbol{x}_t^{\text{past}},t,c) -G(\boldsymbol{x}_t^{1:H},\boldsymbol{x}_t^{\text{past}},t,\emptyset) \}$
        \Statex \hspace*{\algorithmicindent} // $G$ and $c$ are used.
        \State $\boldsymbol{x}_{0}^{1:H}=g(\boldsymbol{x}_{t}^{1:H})$ 
        % \Comment{$c$ is used.}
        \For{$j =1, \dots,J$}
            % \State $\boldsymbol{x}_{\text{plan}}^{1:H}=\Psi(\boldsymbol{x}_{0}^{1:H})$
            \State $\loss_{\text{guide}} = 0$
            \For{$n = 1, \dots,N$}
                \State $\loss_{\text{guide}}=\loss_{\text{guide}}+w_{\text{guide},n} \loss_{\text{guide},n}(\Psi(\boldsymbol{x}_0^{1:H}))$
            \EndFor
            \Statex \hspace*{\algorithmicindent} \hspace*{1em} // \cref{eq_x_0_update} in the main paper.
            \State $\boldsymbol{x}_0^{1:H} \leftarrow \boldsymbol{x}_0^{1:H} - \eta \nabla_{\boldsymbol{x}_{0}^{1:H}} \loss_{\text{guide}}$ 
            % \Comment{\cref{eq_x_0_update}}
            % \State Update $\boldsymbol{x}_0^{1:H}$ using \cref{eq_x_0_update}. \Comment{$\mathcal{L_{\text{disc}}}(\boldsymbol{x}_{\text{plan}}^{1:H})$ is used.}
        \EndFor
        \Statex \hspace*{\algorithmicindent} // Forward process for updated $\boldsymbol{x}_{0}^{1:H}$.
        \State $\boldsymbol{x}_t^{1:H} = \sqrt{\Bar{\alpha}_t} \boldsymbol{x}_0^{1:H} + (1-\Bar{\alpha}_t)\boldsymbol{\epsilon}$ 
        % \Comment{Forward process} 
        % \Comment{$\boldsymbol{x}_t^{1:H} \sim q(\boldsymbol{x}_t^{1:H} | \boldsymbol{x}_0^{1:H})$}
        \Statex \hspace*{\algorithmicindent} // Reverse process.
        \State $\boldsymbol{x}_{t-1}^{1:H} = \frac{\beta_t \sqrt{\bar{\alpha}_{t-1}}}{1-\bar{\alpha}_{t}} \boldsymbol{x}_0^{1:H} + \frac{(1-\bar{\alpha}_{t-1})\sqrt{\alpha_t}}{1-\bar{\alpha}_t} \boldsymbol{x}_t^{1:H} + \sigma_t \boldsymbol{\epsilon}$ 
        % \Statex \Comment{$\boldsymbol{x}_{t-1} \sim q(\boldsymbol{x}_{t-1} | \boldsymbol{x}_0, \boldsymbol{x}_t)$}
    \EndFor

    \State $\boldsymbol{x}_{\text{plan}}^{1:H}=\Psi(\boldsymbol{x}_0^{1:H})$
\end{algorithmic}
\end{algorithm}

%% file: algorithm/estimation_heading_alg.tex
\begin{algorithm}[t]
\caption{Estimating heading vector of motion}
\label{alg:inpaint_heading}
\hspace*{\algorithmicindent} \textbf{Input}: Generated motion plan $\boldsymbol{x}_{\text{plan}}^{1:H}$ \\
% \textbf{Parameter}: Optional list of parameters\\
\hspace*{\algorithmicindent} \textbf{Output}: Estimated heading vector $\Tilde{\boldsymbol{u}}_{\text{head}}$
\begin{algorithmic}[1] %[1] enables line numbers
    \State $\boldsymbol{v}_1 = \boldsymbol{x}_{\text{plan},k_{\text{right-hip}}}^{1:H} - \boldsymbol{x}_{\text{plan},k_{\text{left-hip}}}^{1:H}$
    \State $\boldsymbol{v}_2 = \boldsymbol{x}_{\text{plan},k_{\text{right-shoulder}}}^{1:H} - \boldsymbol{x}_{\text{plan},k_{\text{left-shoulder}}}^{1:H}$
    \State $\boldsymbol{v} = \boldsymbol{v}_1 +\boldsymbol{v}_2$.
    \State Normalize $\boldsymbol{v}$ as $\Bar{\boldsymbol{v}} = \boldsymbol{v} / \|\boldsymbol{v}\|_2$
    \Statex // $\times$ is the cross product.
    \State $\Tilde{\boldsymbol{u}}_{\text{head}} = [0,0,1]^T \times \Bar{\boldsymbol{v}}$ 
    % \Comment{$\times$ is the cross product}
    % \State $\text{cossim} = \boldsymbol{u}_{\text{head}}^T \Tilde{\boldsymbol{u}}_{\text{head}} / \left( \| \boldsymbol{u}_{\text{head}} \|_2 \| \Tilde{\boldsymbol{u}}_{\text{head}} \|_2 \right)$.
    % \State 
\end{algorithmic}
\end{algorithm}

%% file: algorithm/p2r_dataset_alg.tex
% \begin{algorithm}[tb]
% \caption{Dataset construction for position–twist angle pairs}
%     \label{alg:p2r_dataset}
%     \textbf{Input}: Motion dataset $\mathcal{M}$\\
%     % \textbf{Parameter}: Optional list of parameters\\
%     \textbf{Output}: Position-to-rotation dataset $\mathcal{M}_{P2R}$ 
%     \begin{algorithmic}[1] %[1] enables line numbers
%     \State $\{ \mathcal{P}, \mathcal{R} \}_{n=1}^N \sim \mathcal{M}$.
%     \State Derive $\{ \boldsymbol{R} \}_{n=1}^N$ from $\{ \mathcal{R} \}_{n=1}^N$, $\boldsymbol{R}=\{ \mathcal{R}_{pa(k),k} \}_{k=1}^K$
%     \State Calculate $\{ \boldsymbol{R}^{sw} \}_{n=1}^N$ from $\{ \mathcal{P} \}_{n=1}^N$, $\boldsymbol{R}^{sw}=\{ \mathcal{R}^{sw}_{k} \}_{k=1}^K$
%     \State Calculate $\{ \boldsymbol{R}^{tw} \}_{n=1}^N$ from $\{ \boldsymbol{R} \}_{n=1}^N$ and $\{ \boldsymbol{R}^{sw} \}_{n=1}^N$, $\boldsymbol{R}^{tw}=\{ \mathcal{R}^{tw} \}_{k=1}^K$ and $\mathcal{R}^{tw}=(\mathcal{R}^{sw})^{-1} \mathcal{R}_{pa(k),k}$
%     \State $(\overrightarrow{\mathcal{P}}_{\text{temp},k}, \phi)=$ \text{rotmat2axis-angle}($\mathcal{R}^{tw}$)
%     \State $\mathcal{B}=\{ \mathcal{P}, \boldsymbol{\Phi} \}_{n=1}^N$, $\boldsymbol{\Phi} = \{ \phi_k \}_{k=1}^K$
%     \State Repeat, $\mathcal{M}_{P2R}=\{ \mathcal{B} \}_{i=1}^{| \mathcal{M} |}$
%     \end{algorithmic}
% \end{algorithm}
\begin{algorithm}[tb]
\caption{Dataset construction for position–twist angle pairs}
\label{alg:p2r_dataset}
% \textbf{Input}: Motion dataset $\mathcal{M}$\\
\hspace*{\algorithmicindent} \textbf{Input}: Motion dataset $\mathcal{M}$ \\
% \textbf{Parameter}: Optional list of parameters\\
% \textbf{Output}: Position-to-rotation dataset $\mathcal{M}_{P2R}$ 
\hspace*{\algorithmicindent} \textbf{Output}: Position-to-rotation dataset $\mathcal{M}_{P2R}$ 
\begin{algorithmic}[1] %[1] enables line numbers
    \State $\mathcal{M}_{P2R}=\{ \emptyset \}$
    \Statex // $\boldsymbol{\mathcal{P}} = \{ \mathcal{P}_k \}_{k=1}^K$, $\boldsymbol{\mathcal{R}} = \{ \mathcal{R}_k \}_{k=1}^K$.
    \State $\{ \boldsymbol{\mathcal{P}}, \boldsymbol{\mathcal{R}} \}_{n=1}^N \sim \mathcal{M}$
    % \Comment{$\boldsymbol{\mathcal{P}} = \{ \mathcal{P}_k \}_{k=1}^K$, $\boldsymbol{\mathcal{R}} = \{ \mathcal{R}_k \}_{k=1}^K$}
    \Statex // $\boldsymbol{\mathcal{R}}_{\text{Rel}}=\{ \mathcal{R}_{pa(k),k} \}_{k=1}^K$.
    \State Calculate $\{ \boldsymbol{\mathcal{R}}_{\text{Rel}} \}$ 
    % \Comment{$\boldsymbol{\mathcal{R}}_{\text{Rel}}=\{ \mathcal{R}_{pa(k),k} \}_{k=1}^K$}
    % \State Derive $\{ \boldsymbol{R} \}_{n=1}^N$ from $\{ \mathcal{R} \}_{n=1}^N$, $\boldsymbol{R}=\{ \mathcal{R}_{pa(k),k} \}_{k=1}^K$
    % \State Calculate $\{ \boldsymbol{R}^{sw} \}_{n=1}^N$ from $\{ \mathcal{P} \}_{n=1}^N$, $\boldsymbol{R}^{sw}=\{ \mathcal{R}^{sw}_{k} \}_{k=1}^K$
    \Statex // $\boldsymbol{\mathcal{R}}^{sw} = \{ \mathcal{R}^{sw}_k \}_{k=1}^K$.
    \State Calculate $\{ \boldsymbol{\mathcal{R}}^{sw} \}_{n=1}^N$ 
    % \Comment{$\boldsymbol{\mathcal{R}}^{sw} = \{ \mathcal{R}^{sw}_k \}_{k=1}^K$}
    \Statex // $\boldsymbol{\mathcal{R}}^{tw} = \{ (\mathcal{R}^{sw}_k)^{-1} \mathcal{R}_{pa(k),k} \}_{k=1}^K$.
    \State Calculate $\{ \boldsymbol{\mathcal{R}}^{tw} \}_{n=1}^N$ 
    % \Comment{$\boldsymbol{\mathcal{R}}^{tw} = \{ (\mathcal{R}^{sw}_k)^{-1} \mathcal{R}_{pa(k),k} \}_{k=1}^K$}
    % \State Calculate $\{ \boldsymbol{R}^{tw} \}_{n=1}^N$ from $\{ \boldsymbol{R} \}_{n=1}^N$ and $\{ \boldsymbol{R}^{sw} \}_{n=1}^N$, $\boldsymbol{R}^{tw}=\{ \mathcal{R}^{tw} \}_{k=1}^K$ and $\mathcal{R}^{tw}=(\mathcal{R}^{sw})^{-1} \mathcal{R}_{pa(k),k}$
    \Statex // $\boldsymbol{\Phi} = \{ \phi_k \}_{k=1}^K$, $(\overrightarrow{\mathcal{P}}_{\text{rest},k}, \phi_k)=\mathcal{T}_\text{rotmat2axis-angle}(\mathcal{R}^{tw}_k)$.
    \State Calculate $\{ \boldsymbol{\Phi}\}_{n=1}^N$ 
    % \Comment{$\boldsymbol{\Phi} = \{ \phi_k \}_{k=1}^K$, $(\overrightarrow{\mathcal{P}}_{\text{rest},k}, \phi_k)=\mathcal{T}_\text{rotmat2axis-angle}(\mathcal{R}^{tw}_k)$}
    % $(\overrightarrow{\mathcal{P}}_{\text{temp},k}, \phi_k)=$ \text{Rotmat2Axis-angle}($\mathcal{R}^{tw}_k$)
    % \State $(\overrightarrow{\mathcal{P}}_{\text{temp},k}, \phi)=$ \text{Rotmat2Axis-angle}($\mathcal{R}^{tw}$)
    \State $\mathcal{M}_{P2R} \leftarrow \mathcal{M}_{P2R} \cup \{ \boldsymbol{\mathcal{P}}, \boldsymbol{\Phi} \}_{n=1}^N$
    % \State Repeat, $\mathcal{M}_{P2R}=\{ \mathcal{B} \}_{i=1}^{| \mathcal{M} |}$
\end{algorithmic}
\end{algorithm}

%% file: algorithm/p2r_training_alg.tex
% \begin{algorithm}[tb]
% \caption{Dataset construction for position–twist angle pairs}
%     \label{alg:p2r_training}
%     \textbf{Input}: Motion dataset $\mathcal{M}$, position-to-twist angle network $\Omega$, $\mathcal{T}_{\text{rotmat26d}}$ \\
%     % \textbf{Parameter}: Optional list of parameters\\
%     \textbf{Output}: Trained position-to-twist angle network $\Omega$
%     \begin{algorithmic}[1] %[1] enables line numbers
%     \State $\{ \mathcal{P}, \boldsymbol{\Phi} \} \sim \mathcal{M}_{P2R}$
%     \State $\mathcal{P}_{\text{relative}} = \{ \mathcal{P}_k-\mathcal{P}_1 \}_{k=1}^K$
%     \State $\Tilde{\boldsymbol{\Phi}} = \Omega(\mathcal{P}_{\text{relative}})$
%     \State Calculate $\Tilde{\mathcal{R}}^{tw}$ from $\Tilde{\boldsymbol{\Phi}}$, $\Tilde{\mathcal{R}}^{tw} = \mathcal{D}^{tw}(\mathcal{P}_{\text{temp},k} - \mathcal{P}_{\text{temp},pa(k)}, \Tilde{\boldsymbol{\Phi}})$
%     \State $\Tilde{\mathcal{R}}_{pa(k),k} = \mathcal{R}^{sw} \Tilde{\mathcal{R}}^{tw}$
%     \State Minimize $\| \boldsymbol{\Phi} - \Tilde{\boldsymbol{\Phi}} \|_2^2 + \|\mathcal{T}_{\text{rotmat26d}}(\mathcal{R}_{pa(k),k}) - \mathcal{T}_{\text{rotmat26d}}(\Tilde{\mathcal{R}}_{pa(k),k})\|_2^2$  
%     \end{algorithmic}
% \end{algorithm}
\begin{algorithm}[tb]
\caption{Training $P2R$ over $\mathcal{M}_{P2R}$}
\label{alg:p2r_training}
% \textbf{Input}: Position-to-rotation dataset $\mathcal{M}_{P2R}$, position-to-twist angle network $\Omega$, $\mathcal{T}_{\text{rotmat26d}}$ \\
\hspace*{\algorithmicindent} \textbf{Input}: Position-to-rotation dataset $\mathcal{M}_{P2R}$\\
\hspace*{\algorithmicindent} \quad \quad \quad Position-to-twist angle network $\Omega$ \\
% \textbf{Parameter}: Optional list of parameters\\
% \textbf{Output}: Trained position-to-twist angle network $\Omega$
\hspace*{\algorithmicindent} \textbf{Output}: Trained position-to-twist angle network $\Omega$
\begin{algorithmic}[1] %[1] enables line numbers
    \State $\{ \boldsymbol{\mathcal{P}}, \boldsymbol{\Phi} \} \sim \mathcal{M}_{P2R}$
    \State $\boldsymbol{\mathcal{P}}_{\text{root-rel}} = \{ \mathcal{P}_k-\mathcal{P}_1 \}_{k=1}^K$
    \Statex // $\Tilde{\boldsymbol{\Phi}} = \{ \Tilde{\phi}_k \}_{k=1}^K $, $\theta_{\Omega}$ is a paramter of $\Omega$.
    \State $\Tilde{\boldsymbol{\Phi}} = \Omega( \boldsymbol{\mathcal{P}}_{\text{root-rel}} ;\theta_{\Omega})$ 
    % \Comment{$\Tilde{\boldsymbol{\Phi}} = \{ \Tilde{\phi}_k \}_{k=1}^K $, $\theta_{\Omega}$ is a paramter of $\Omega$}
    \Statex // $ \Tilde{\boldsymbol{\mathcal{R}}}^{tw}=\{ \Tilde{\mathcal{R}}^{tw}_{k} \}_{k=1}^K$.
    \Statex // $\Tilde{\mathcal{R}}^{tw}_{k} = \mathcal{D}^{tw}(\mathcal{P}_{\text{rest},k} - \mathcal{P}_{\text{rest},pa(k)}, \Tilde{\phi}_k)$.
    \State Calculate $\Tilde{\boldsymbol{\mathcal{R}}}^{tw}$ from $\Tilde{\boldsymbol{\Phi}}$ 
    % \Comment{$ \Tilde{\boldsymbol{\mathcal{R}}}^{tw}=\{ \Tilde{\mathcal{R}}^{tw}_{k} \}_{k=1}^K$, $\Tilde{\mathcal{R}}^{tw}_{k} = \mathcal{D}^{tw}(\mathcal{P}_{\text{temp},k} - \mathcal{P}_{\text{temp},pa(k)}, \Tilde{\phi}_k)$} 
    \Statex // $\Tilde{\boldsymbol{\mathcal{R}}}_{\text{Rel}} = \{ \mathcal{R}^{sw}_{k} \Tilde{\mathcal{R}}^{tw}_{k} \}_{k=1}^K$.
    \State Calculate $\Tilde{\boldsymbol{\mathcal{R}}}_{\text{Rel}}$
    % \Comment{$\Tilde{\boldsymbol{\mathcal{R}}}_{\text{Rel}} = \{ \mathcal{R}^{sw}_{k} \Tilde{\mathcal{R}}^{tw}_{k} \}_{k=1}^K$}
    % \State $\Tilde{\mathcal{R}}_{pa(k),k} = \mathcal{R}^{sw} \Tilde{\mathcal{R}}^{tw}$
    % \Statex $\mathcal{T}_{\text{rotmat26d}}$ is a function maps rotation matrix into 6d rotation representation. 
    \State Update $\theta_{\Omega}$ to minimize $\| \boldsymbol{\Phi} - \Tilde{\boldsymbol{\Phi}} \|_2^2 + \|\mathcal{T}_{\text{rotmat26d}}(\boldsymbol{\mathcal{R}}_{\text{Rel}}) - \mathcal{T}_{\text{rotmat26d}}(\Tilde{\boldsymbol{\mathcal{R}}}_{\text{Rel}})\|_2^2$
    % \State Minimize $\| \boldsymbol{\Phi} - \Tilde{\boldsymbol{\Phi}} \|_2^2 + \|\mathcal{T}_{\text{rotmat26d}}(\mathcal{R}_{pa(k),k}) - \mathcal{T}_{\text{rotmat26d}}(\Tilde{\mathcal{R}}_{pa(k),k})\|_2^2$  
    \end{algorithmic}
\end{algorithm}

%% file: figs/supp_long_term_task_detail.tex
% \begin{figure*}[t!]
% % \captionsetup[subfigure]{justification=centering}
%     \centering
%     \includegraphics[width=1.0\textwidth]{figs/supplementary/supp_long_term_task_description_v1.pdf}
%     \caption{
%         {\bf More details of Long-Term Tasks.} 
%     }
%     \label{appendix_fig:log_term_task_detail}
% \end{figure*}

\begin{figure}[t!]
% \captionsetup[subfigure]{justification=centering}
    \centering
    \includegraphics[width=\linewidth]{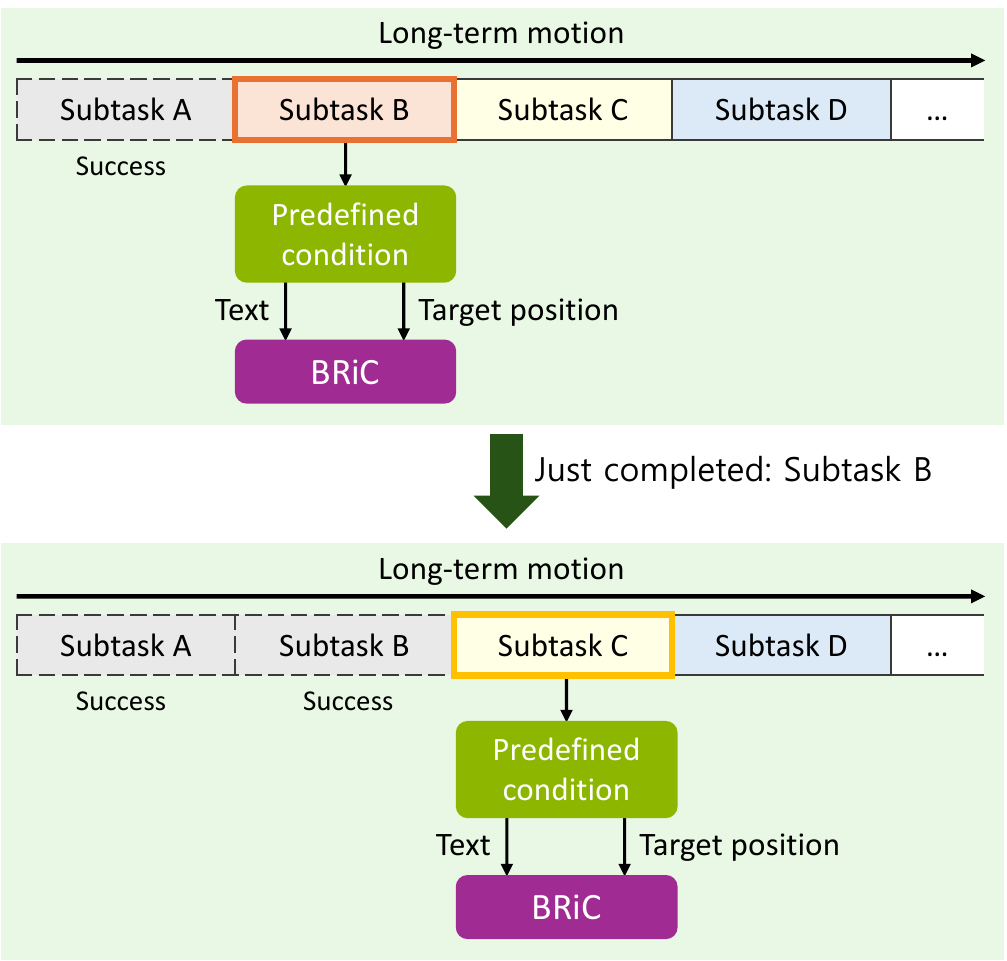}
    \caption{
        % {\bf More details of Long-Term Tasks.} 
        {\bf The process of long-term motion execution.} 
        % 이 그림은 subtask들로 구성된 long-term motion을 \our{}가 실행하는 과정을 나타낸다. 현재 subtask에 대해서 성공이 발생할 경우, \our{}에 자동적으로 다음 subtask의 condition이 입력된다.
        %This figure illustrates how \ours{} executes long-term motion composed of subtasks. 
        Upon successful completion of the current subtask, the condition for the next subtask is automatically fed into \ours{}.
    }
    \label{appendix_fig:log_term_task_detail}
\end{figure}

%% file: figs/visualization_goal_reaching.tex
% \begin{figure*}[t]
% \captionsetup[subfigure]{justification=centering}
%     \centering
%     \begin{subfigure}[t]{0.25\linewidth}
%         \centering
%         \includegraphics[width=1.0\textwidth]{figs/failure_cases/oa_failure_case.png}
%         \caption{\red{XX}th frame}
%     \end{subfigure}\hfil%
%     \begin{subfigure}[t]{0.25\linewidth}
%         \centering
%         \includegraphics[width=1.0\textwidth]{figs/failure_cases/oa_failure_case.png}
%         \caption{\red{XX}th frame}
%     \end{subfigure}\hfil%
%     \begin{subfigure}[t]{0.25\linewidth}
%         \centering
%         \includegraphics[width=1.0\textwidth]{figs/failure_cases/oa_failure_case.png}
%         \caption{\red{XX}th frame}
%     \end{subfigure}\hfil
%     \begin{subfigure}[t]{0.25\linewidth}
%         \centering
%         \includegraphics[width=1.0\textwidth]{figs/failure_cases/oa_failure_case.png}
%         \caption{\red{XX}th frame}
%     \end{subfigure}\hfil
%     \caption{
%         {\bf Visualization of goal-reaching.}\\
%         text\\
%         text\\
%     }
%     \label{appendix_fig:visualization_goal_reach}
% \end{figure*}
\begin{figure*}[!t]
% \captionsetup[subfigure]{justification=centering}
    \centering
    \includegraphics[width=1.0\textwidth]{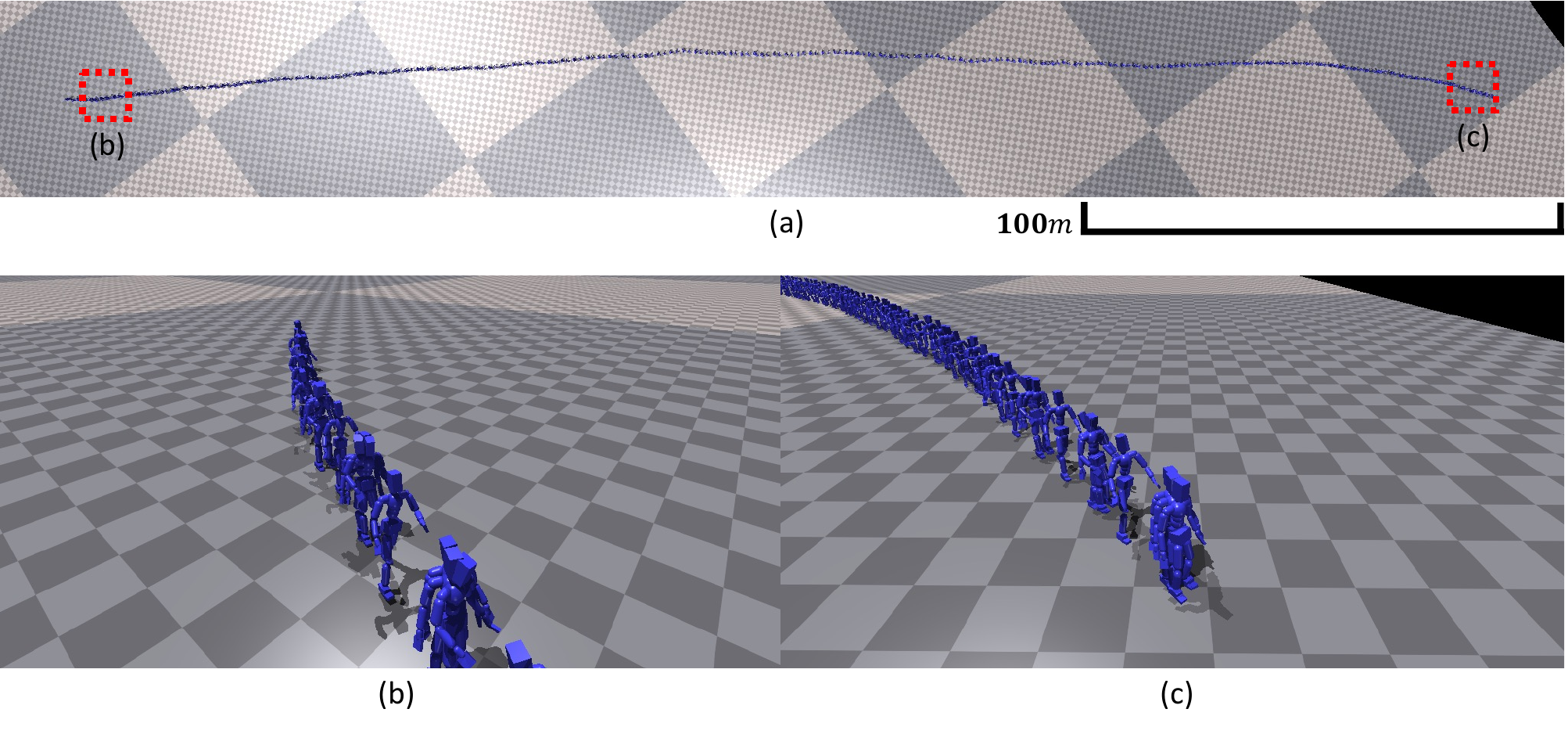}
    \caption{
        {\bf Qualitative results of goal-reaching.} 
        % 이 그림은 시작 지점에서 마지막 지점까지의 거리가 300m인 goal-reaching task에서 jog 스타일로 \ours{}이 locomotion을 수행한 모션에 대한 시각화이다. (a)는 실행된 모션의 진행의 초반부와 (b)는 후반부를 나타낸다.
        % This figure visualizes the motion performed by \ours{} in a goal-reaching task with a total distance of 300 meters, executed in a \emph{jog} style. (b) shows the early phase of the trajectory, and (c) depicts the later phase.
        (a) A 300-meter goal-reaching task in a \emph{jog} style, with the corresponding trajectory shown at (b) the start and (c) the goal location.
    }
    \label{appendix_fig:visualization_goal_reach}
\end{figure*}

%% file: figs/visualization_obstacle_avoidance.tex
% \begin{figure*}[t]
% \captionsetup[subfigure]{justification=centering}
%     \centering
%     \begin{subfigure}[t]{0.25\linewidth}
%         \centering
%         \includegraphics[width=1.0\textwidth]{figs/failure_cases/oa_failure_case.png}
%         \caption{\red{XX}th frame}
%     \end{subfigure}\hfil%
%     \begin{subfigure}[t]{0.25\linewidth}
%         \centering
%         \includegraphics[width=1.0\textwidth]{figs/failure_cases/oa_failure_case.png}
%         \caption{\red{XX}th frame}
%     \end{subfigure}\hfil%
%     \begin{subfigure}[t]{0.25\linewidth}
%         \centering
%         \includegraphics[width=1.0\textwidth]{figs/failure_cases/oa_failure_case.png}
%         \caption{\red{XX}th frame}
%     \end{subfigure}\hfil
%     \begin{subfigure}[t]{0.25\linewidth}
%         \centering
%         \includegraphics[width=1.0\textwidth]{figs/failure_cases/oa_failure_case.png}
%         \caption{\red{XX}th frame}
%     \end{subfigure}\hfil
%     \caption{
%         {\bf Visualization of obstacle avoidance.}\\
%         text\\
%         text\\
%     }
%     \label{appendix_fig:visualization_oa}
% \end{figure*}
\begin{figure*}[t!]
\captionsetup[subfigure]{justification=centering}
    \centering
    \includegraphics[width=1.0\textwidth]{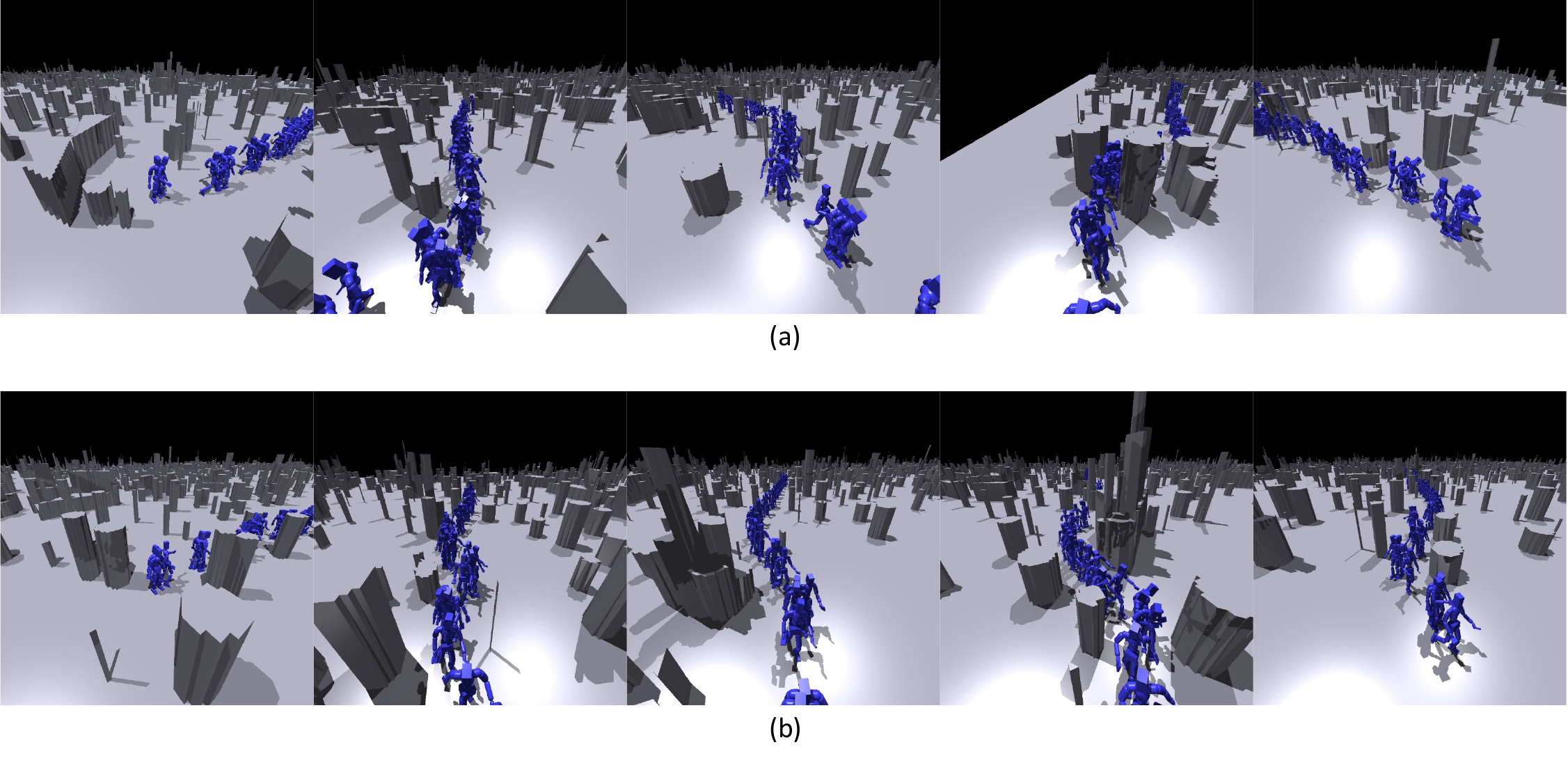}
    \caption{
        {\bf Visualization of obstacle avoidance.} 
        % (a)와 (b)는 각각 다르게 배치된 장애물들을 극복한 \ours{}을 보여준다. ``(1)''부터 ``(5)''까지는 시작부터 최종 지점까지 \ours{}가 jog 스타일로 locomotion을 하는 동시에 장애물을 극복하는 모습을 나타낸다.
        (a) and (b) navigating through differently arranged obstacles.  
        % Frames (1) to (5) illustrate the agent performing \emph{jog}-style locomotion while simultaneously navigating obstacles from the start to the final goal.
        From left to right, the agent performs \emph{jog}-style locomotion while simultaneously avoiding obstacles. 
        The leftmost and rightmost frames correspond to the agent's start and goal positions, respectively.
    }
    \label{appendix_fig:visualization_oa}
\end{figure*}

%% file: algorithm/optimization_target_using_collision_loss_alg.tex
\begin{algorithm}[tb]
\caption{Optimizing intermediate target points}
\label{alg:optimization_target_point}
% \textbf{Input}: Humanoid point $\mathcal{P}_{\text{agent}}$, intermediate target point $\mathcal{P}_{\text{int-target}}$, target point $\mathcal{P}_{\text{target}}$, intermediate target distance $d_{\text{int-target}}$ \\
% \textbf{Parameter}: Optional list of parameters\\
\hspace*{\algorithmicindent} \textbf{Input}: Humanoid point $\mathcal{P}_{\text{agent}}$\\
\hspace*{\algorithmicindent} \quad \quad \quad Target point $\mathcal{P}_{\text{target}}$ \\
\hspace*{\algorithmicindent} \quad \quad \quad Intermediate target point $\mathcal{P}_{\text{int-target}}$ \\
\hspace*{\algorithmicindent} \quad \quad \quad Intermediate target distance $d_{\text{int-target}}$ \\
\hspace*{\algorithmicindent} \textbf{Output}: Optimized $\hat{\mathcal{P}}_{\text{int-target}}^{XY}$ 
\begin{algorithmic}[1] %[1] enables line numbers
    \State $\mathcal{P}_{\text{agent}}^{XY},\mathcal{P}_{\text{int-target}}^{XY} \leftarrow \text{projection } \mathcal{P}_{\text{agent}},\mathcal{P}_{\text{int-target}}$ onto XY-plane
    \Statex $\mathcal{P}_{\text{target}}^{XY} \leftarrow \text{projection } \mathcal{P}_{\text{target}}$ onto XY-plane
    % \State $\mathcal{P}_{\text{target}}^{XY} \leftarrow \text{projection } \mathcal{P}_{\text{target}}$ on XY-plane.
    \State Compute \emph{yaw} angle from $[0,1]^T$ to $\mathcal{P}_{\text{int-target}}^{XY} - \mathcal{P}_{\text{agent}}^{XY}$ 
    % \State $\mathcal{R}_{yaw} = [[\cos \emph{yaw}, \sin \emph{yaw}]^T,[\cos \emph{yaw}, -\sin \emph{yaw}]^T]$
    \State $\mathcal{R}_{yaw} = 
        \begin{bmatrix}
            \cos \emph{yaw} & -\sin \emph{yaw} \\
            \sin \emph{yaw} & \cos \emph{yaw}
        \end{bmatrix}$
    \State $\hat{\mathcal{P}}_{\text{int-target}}^{XY}=\mathcal{R}_{yaw} (d_{\text{int-target}} [0, 1]^T) + \mathcal{P}_{\text{agent}}^{XY}$
    \State Update \emph{yaw}, $d_{\text{int-target}}$ to minimize $\loss_{\text{Obst}}(\hat{\mathcal{P}}_{\text{int-target}}^{XY}) + 0.1\| \hat{\mathcal{P}}_{\text{int-target}}^{XY}-\mathcal{P}_{\text{target}}^{XY} \|_2$ subject to $0.5 \leq d_{\text{int-target}} \leq 1$
    \State Repeat lines 3--5 until convergence
\end{algorithmic}
\end{algorithm}

%% file: figs/hsi_further_qualitative_res.tex
\begin{figure*}[h]
\captionsetup[subfigure]{justification=centering}
    \centering
    \begin{subfigure}[t]{0.25\linewidth}
        \centering
        \includegraphics[width=1.0\textwidth]{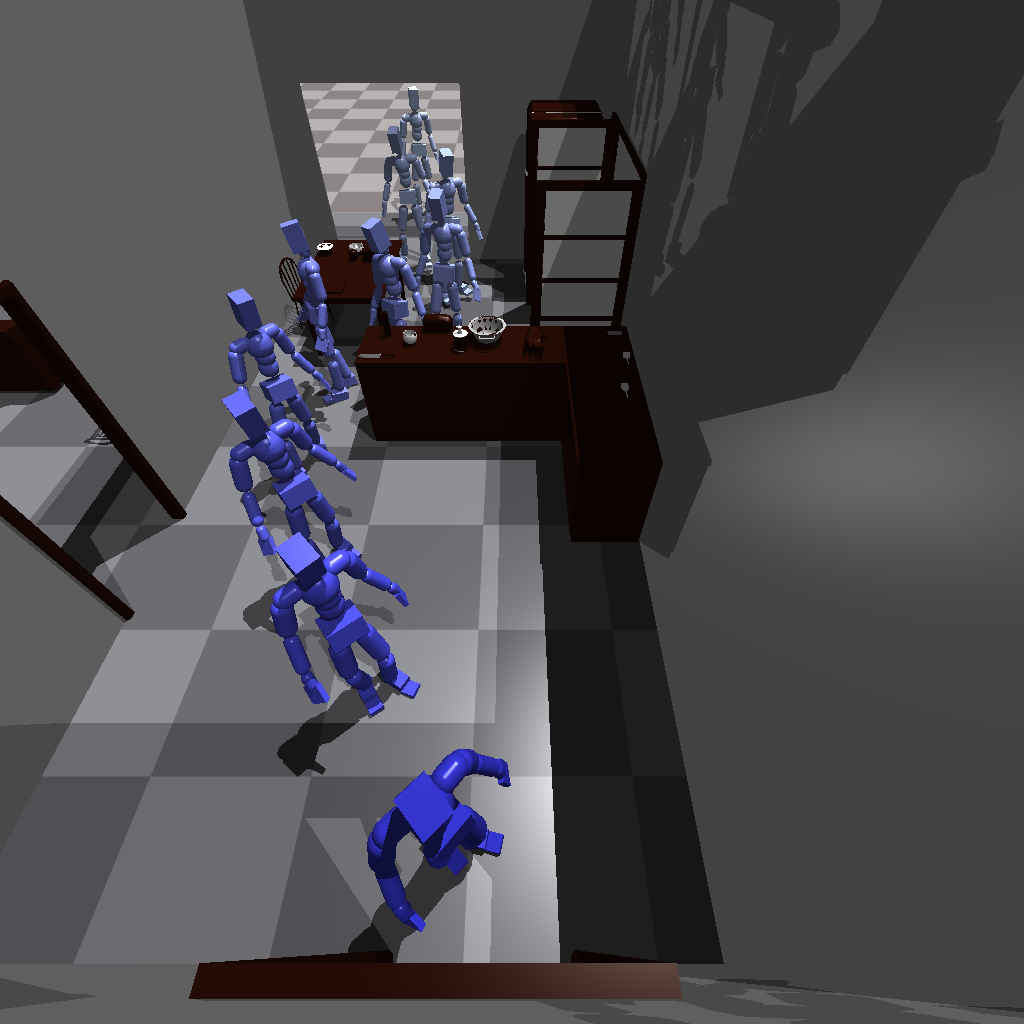}
        \caption{\emph{Enter}~(\emph{Room3})}
    \end{subfigure}\hfil%
    \begin{subfigure}[t]{0.25\linewidth}
        \centering
        \includegraphics[width=1.0\textwidth]{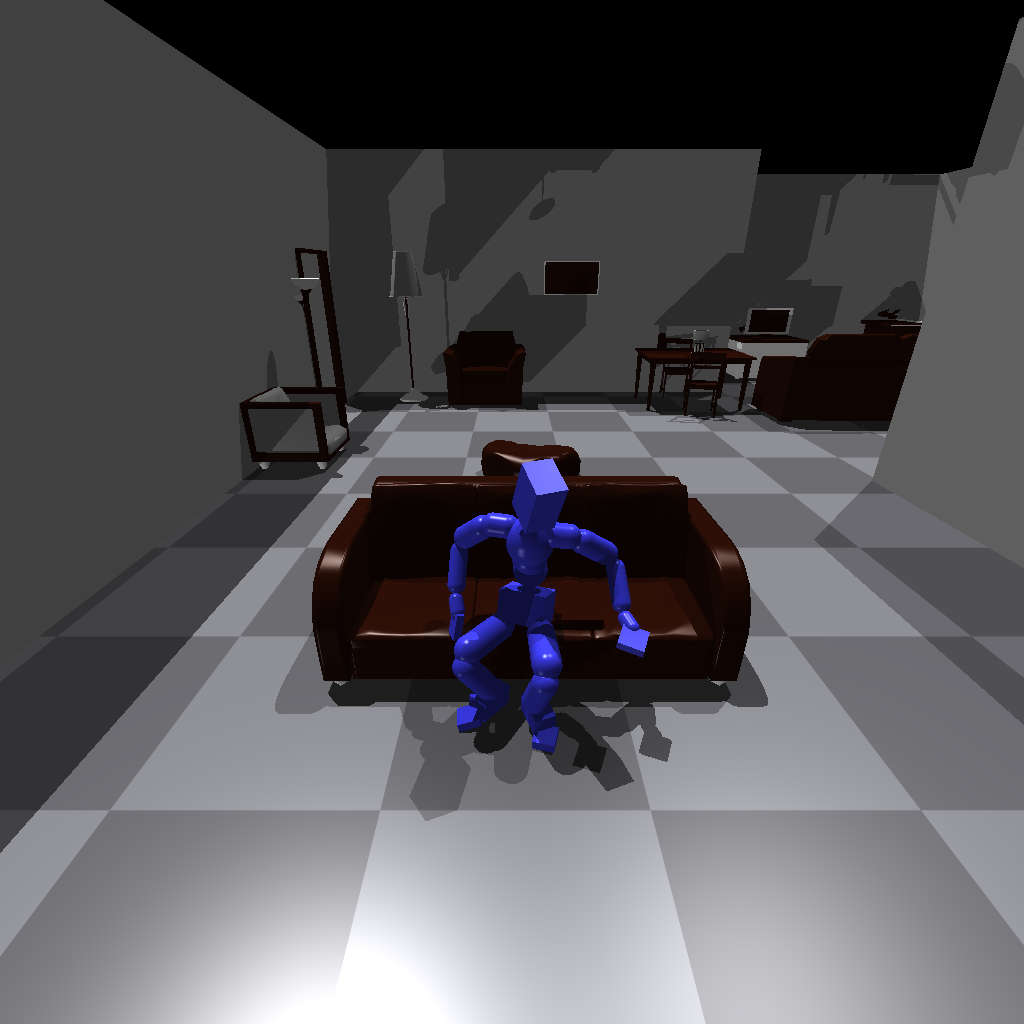}
        \caption{\emph{SIT}~(\emph{Room1})}
    \end{subfigure}\hfil%
    \begin{subfigure}[t]{0.25\linewidth}
        \centering
        \includegraphics[width=1.0\textwidth]{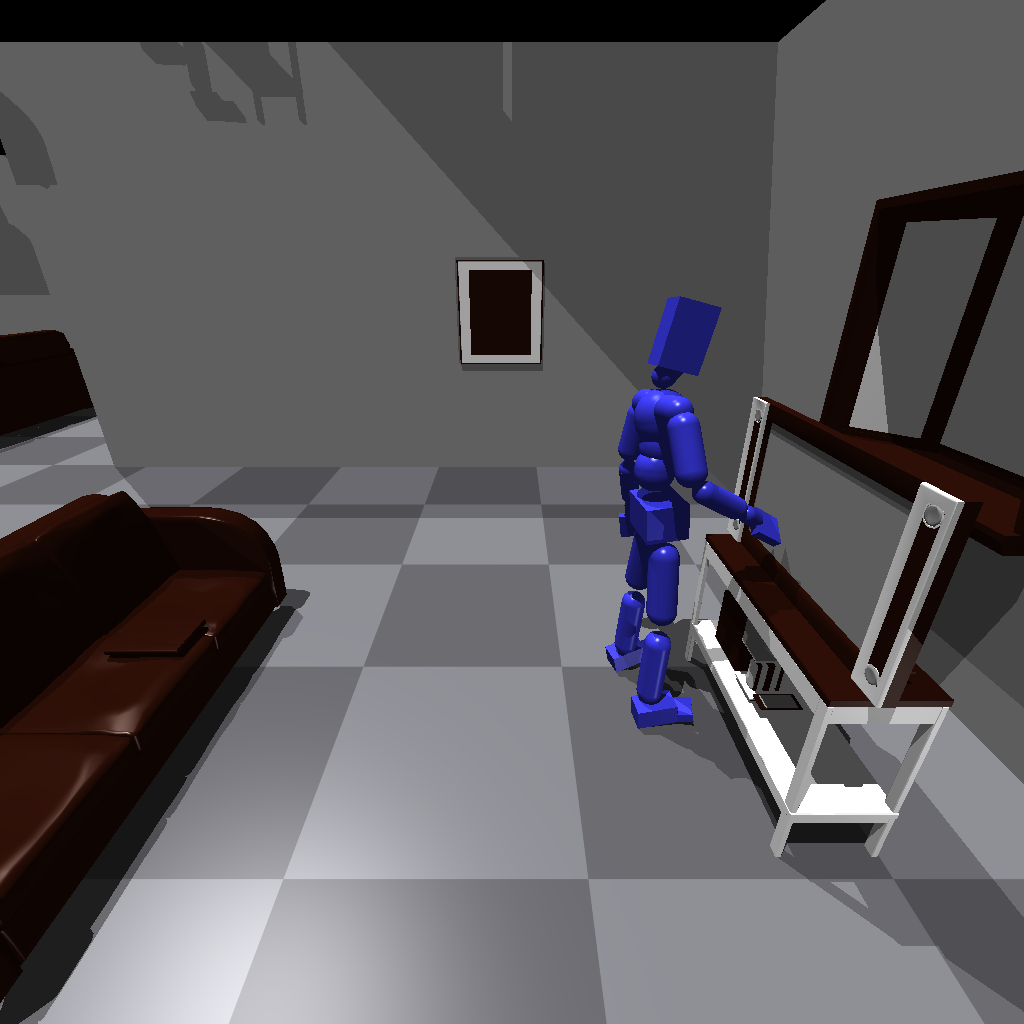}
        \caption{\emph{TOUCH}~(\emph{Room1})}
    \end{subfigure}\hfil%
    \begin{subfigure}[t]{0.25\linewidth}
        \centering
        \includegraphics[width=1.0\textwidth]{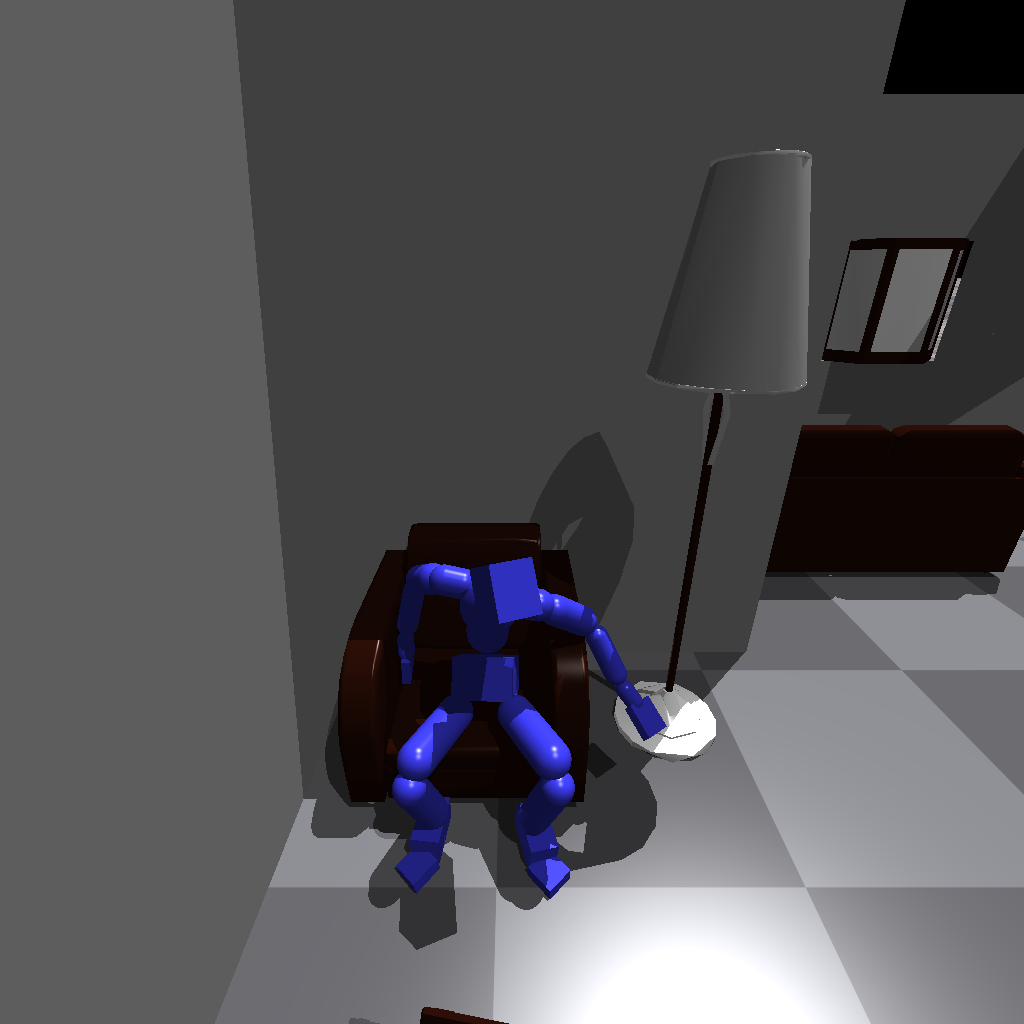}
        \caption{\emph{SIT}~(\emph{Room2})}
    \end{subfigure}\hfil
    \begin{subfigure}[t]{0.25\linewidth}
        \centering
        \includegraphics[width=1.0\textwidth]{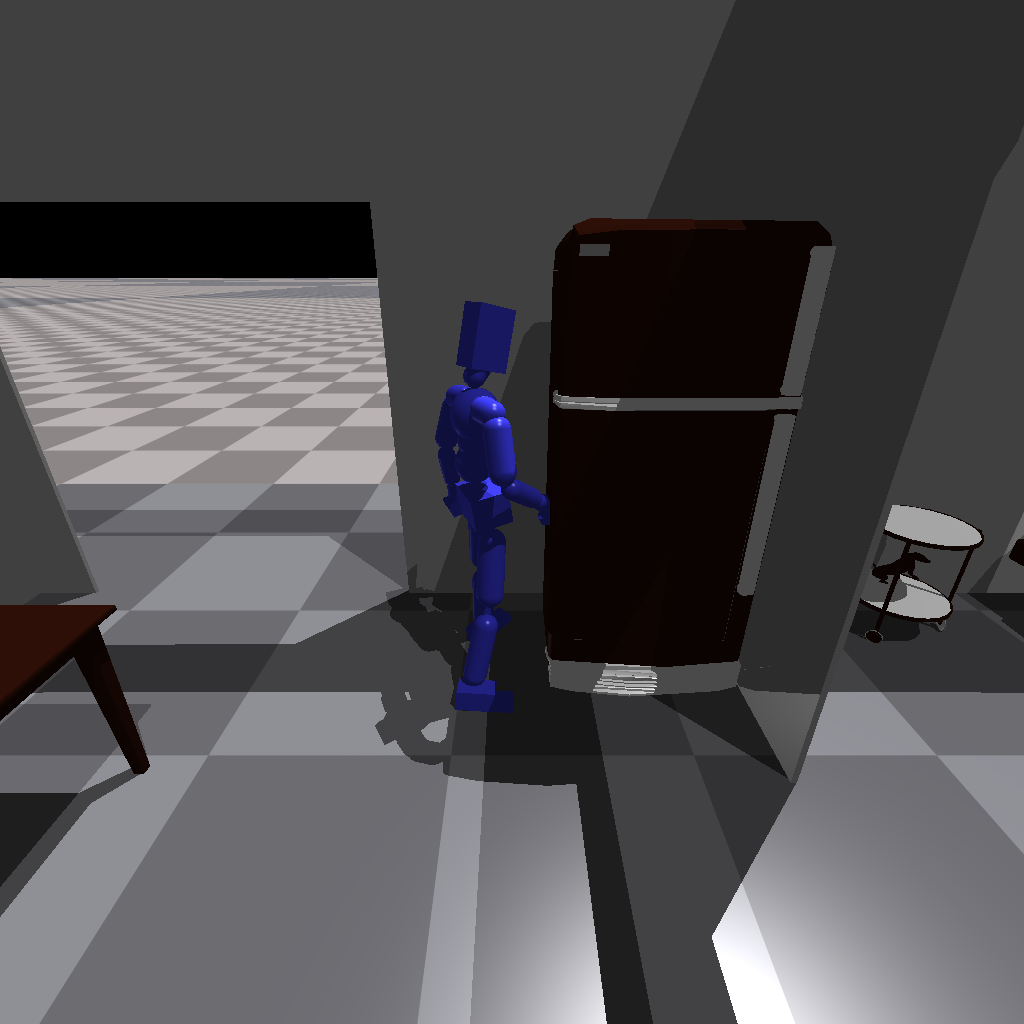}
        \caption{\emph{TOUCH}~(\emph{Room3})}
    \end{subfigure}\hfil%
    \begin{subfigure}[t]{0.25\linewidth}
        \centering
        \includegraphics[width=1.0\textwidth]{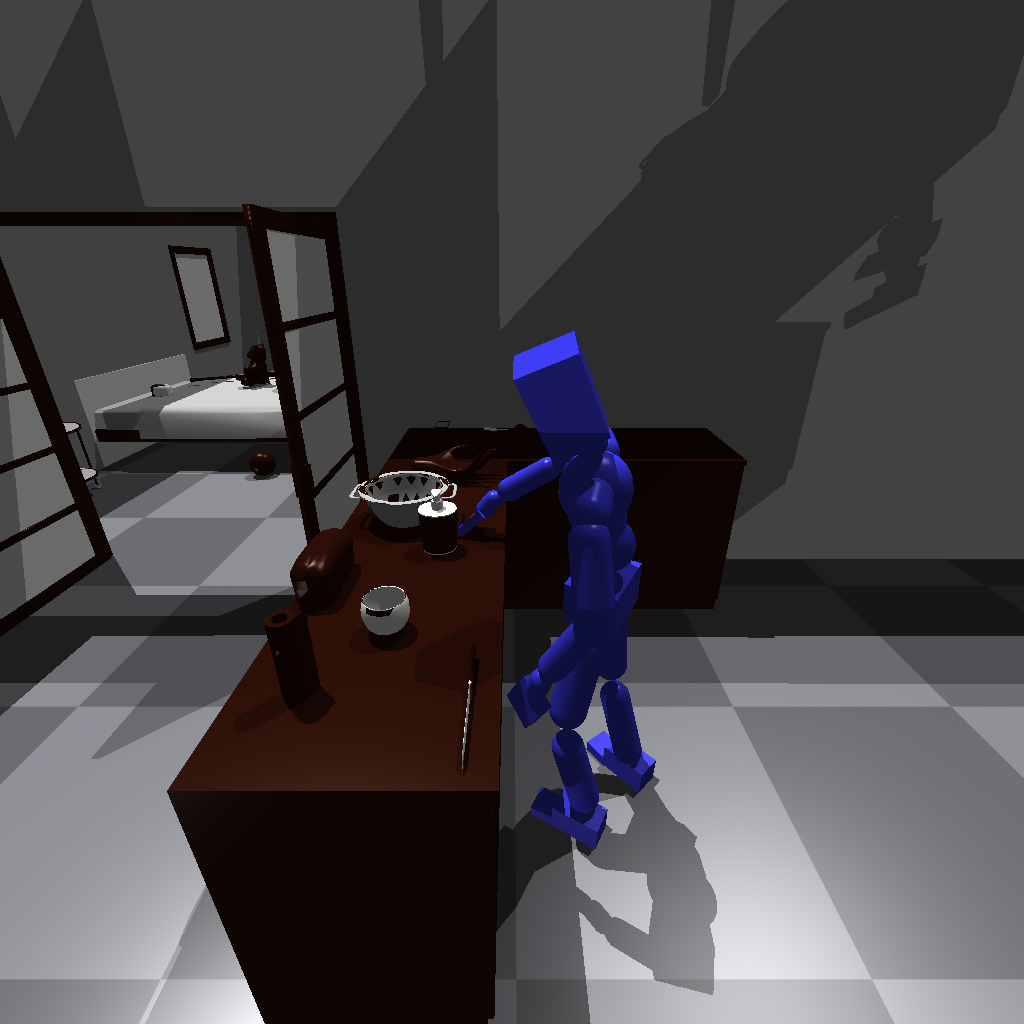}
        \caption{\emph{TOUCH}~(\emph{Room3})}
    \end{subfigure}\hfil%
    \begin{subfigure}[t]{0.25\linewidth}
        \centering
        \includegraphics[width=1.0\textwidth]{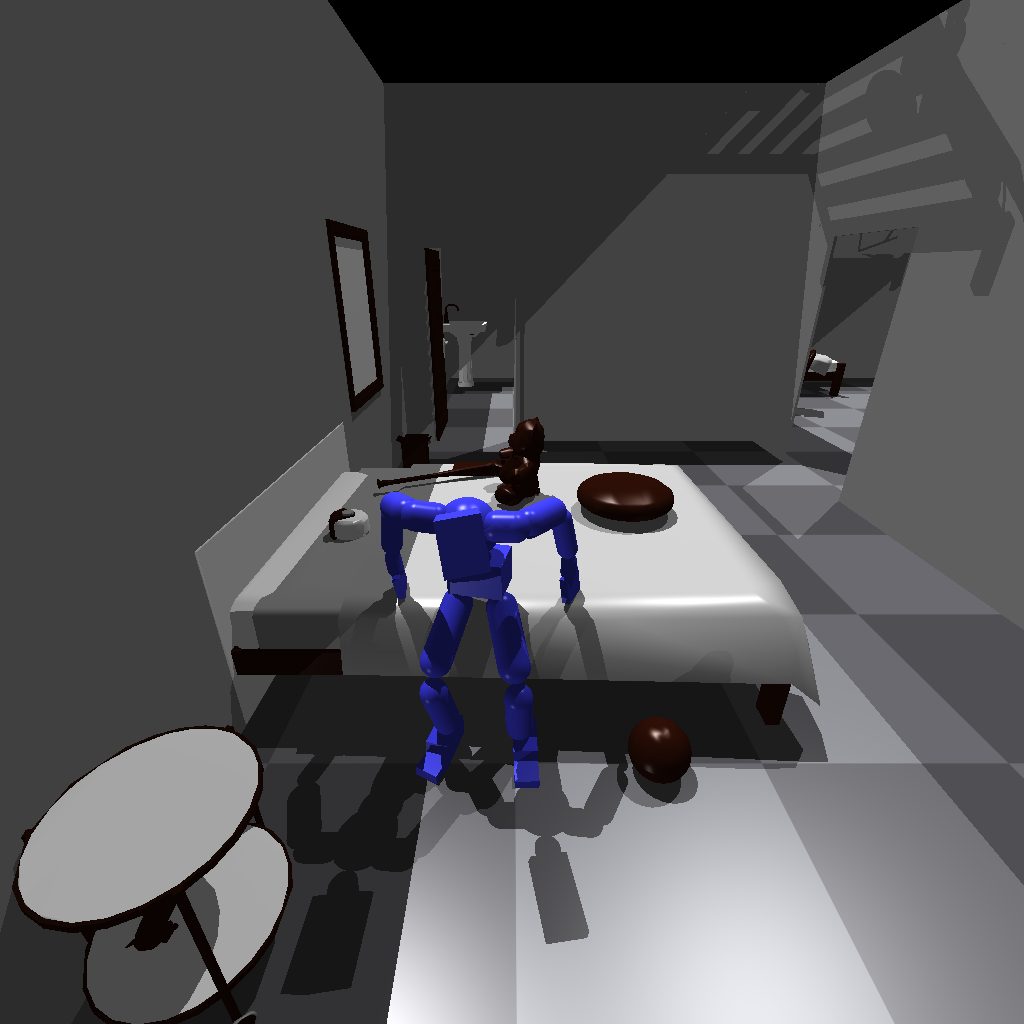}
        \caption{\emph{SIT}~(\emph{Room4})}
    \end{subfigure}\hfil
    \begin{subfigure}[t]{0.25\linewidth}
        \centering
        \includegraphics[width=1.0\textwidth]{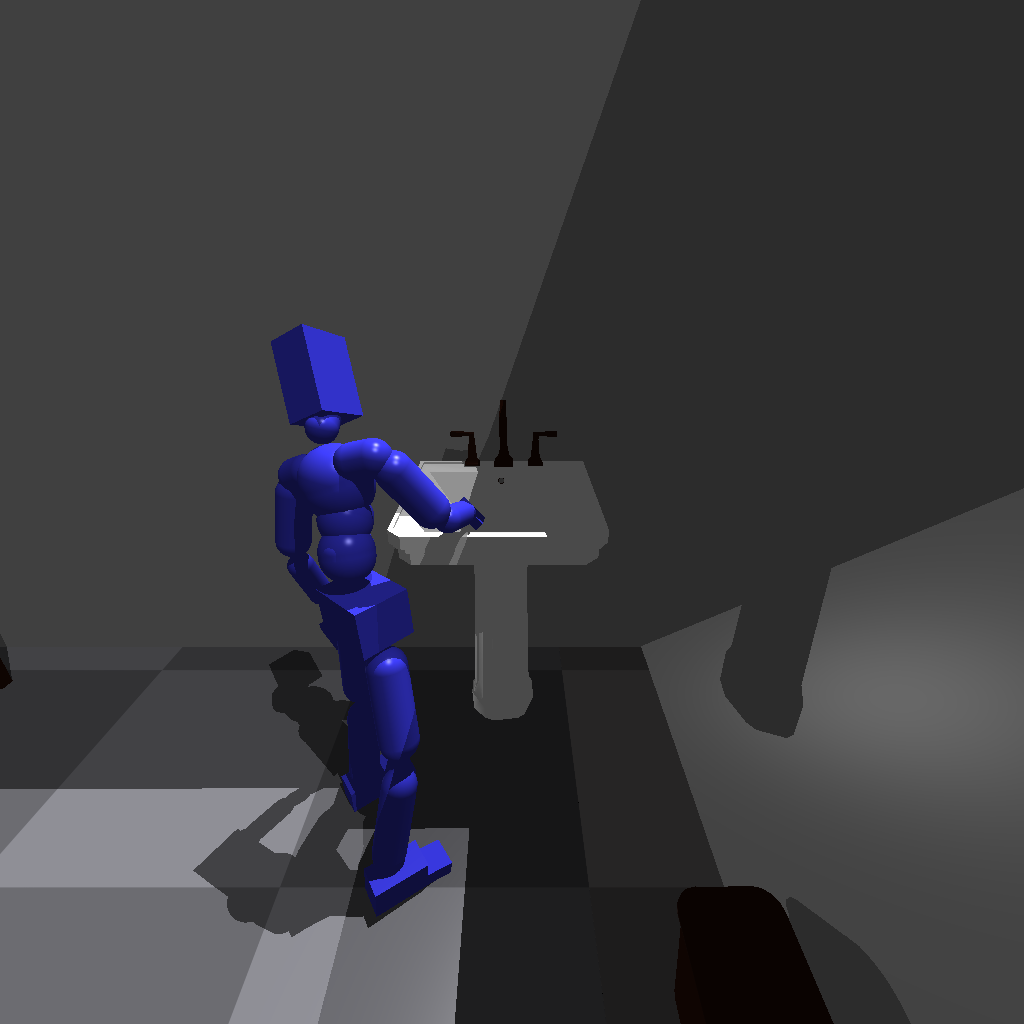}
        \caption{\emph{TOUCH}~(\emph{Room5})}
    \end{subfigure}\hfil
    \begin{subfigure}[t]{0.25\linewidth}
        \centering
        \includegraphics[width=1.0\textwidth]{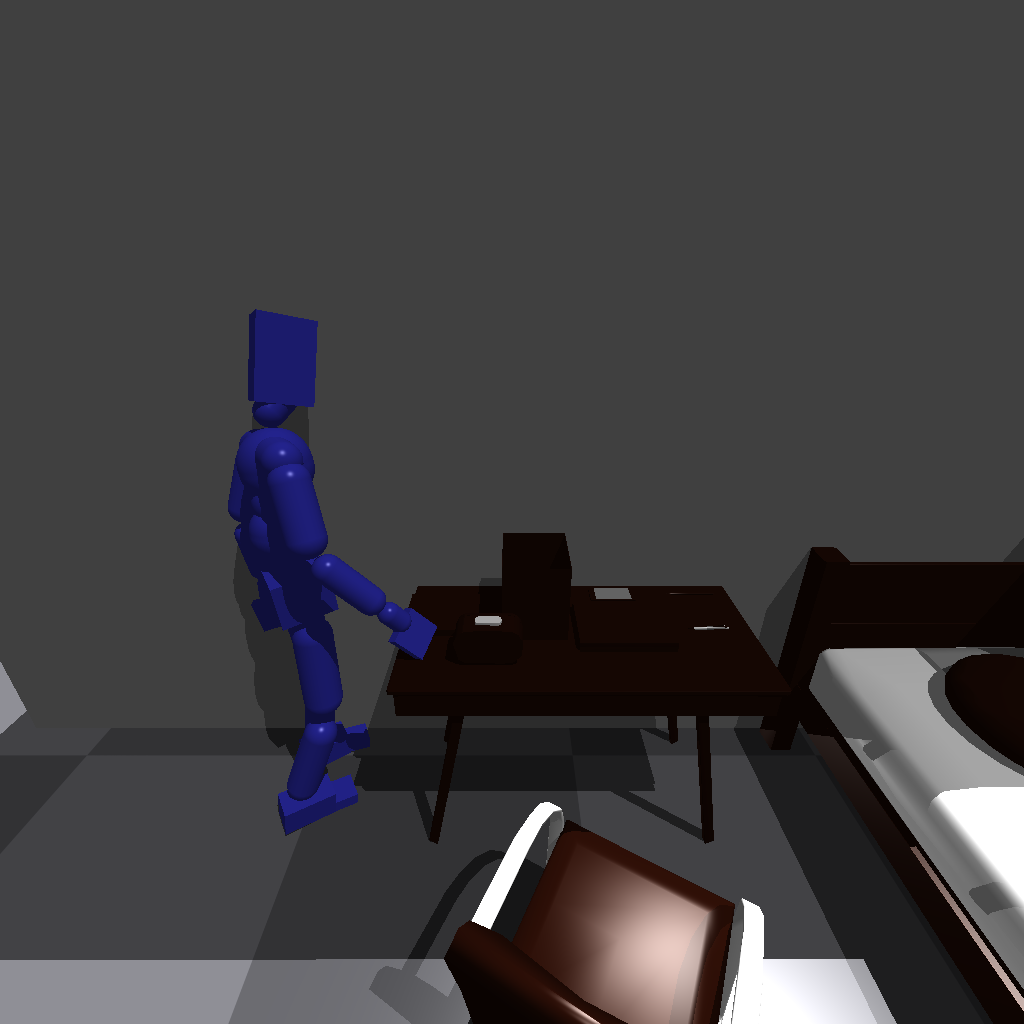}
        \caption{\emph{TOUCH}~(\emph{Room6})}
    \end{subfigure}\hfil%
    \begin{subfigure}[t]{0.25\linewidth}
        \centering
        \includegraphics[width=1.0\textwidth]{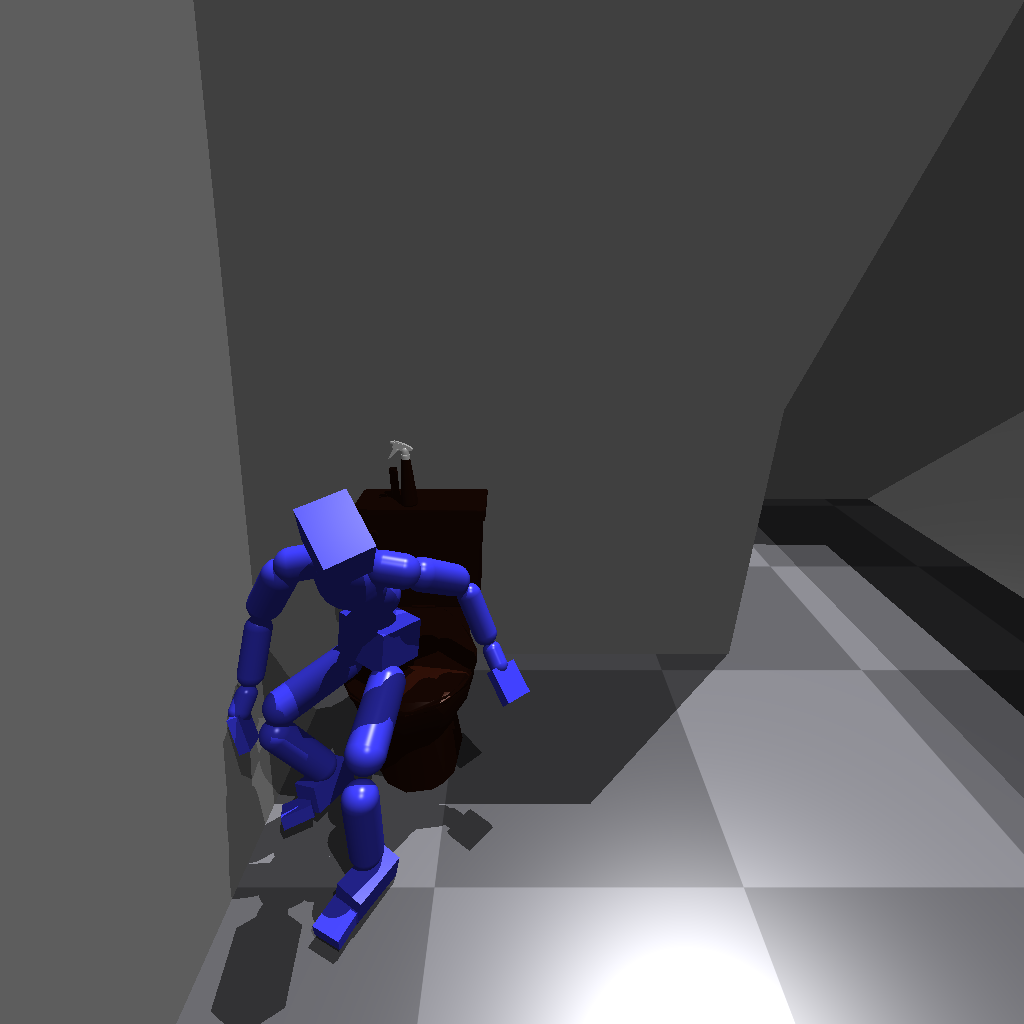}
        \caption{\emph{SIT}~(\emph{Room7})}
    \end{subfigure}\hfil%
    \begin{subfigure}[t]{0.25\linewidth}
        \centering
        \includegraphics[width=1.0\textwidth]{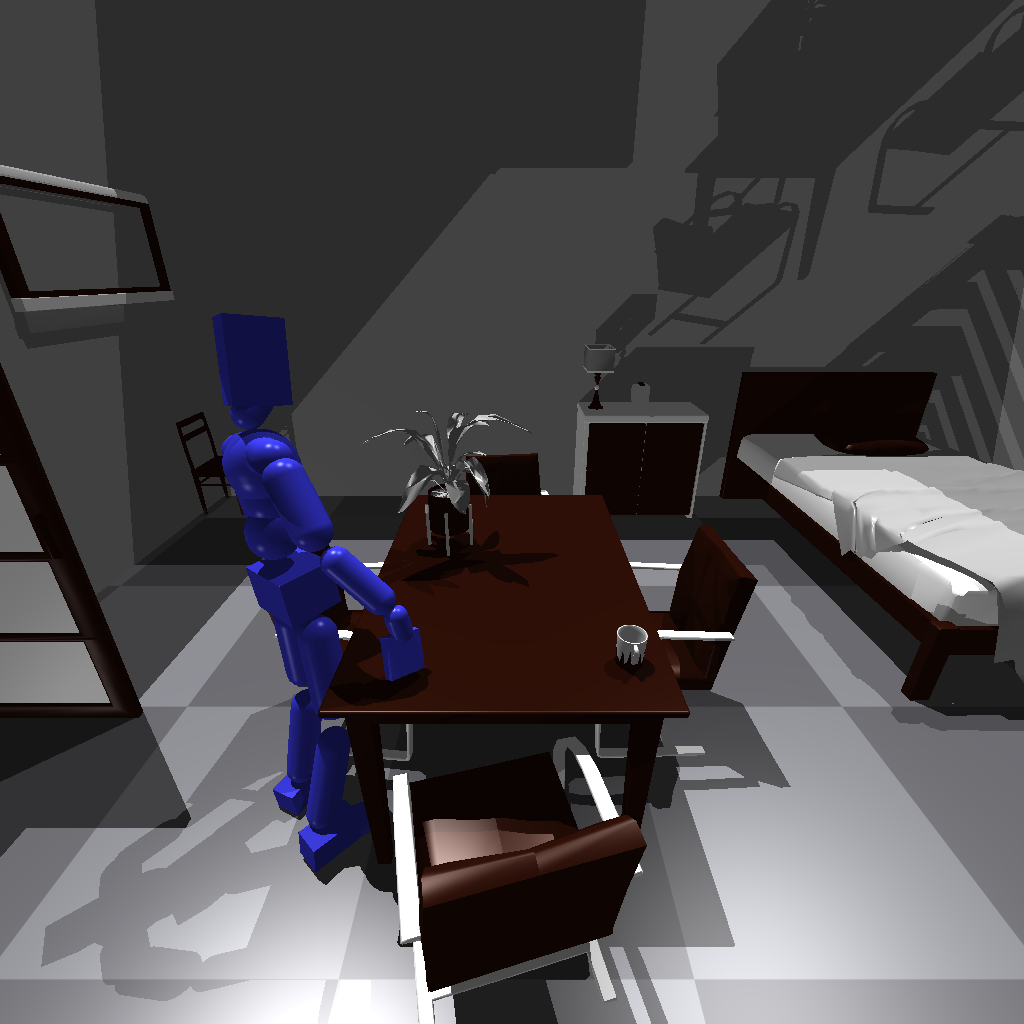}
        \caption{\emph{TOUCH}~(\emph{Room8})}
    \end{subfigure}\hfil
    \begin{subfigure}[t]{0.25\linewidth}
        \centering
        \includegraphics[width=1.0\textwidth]{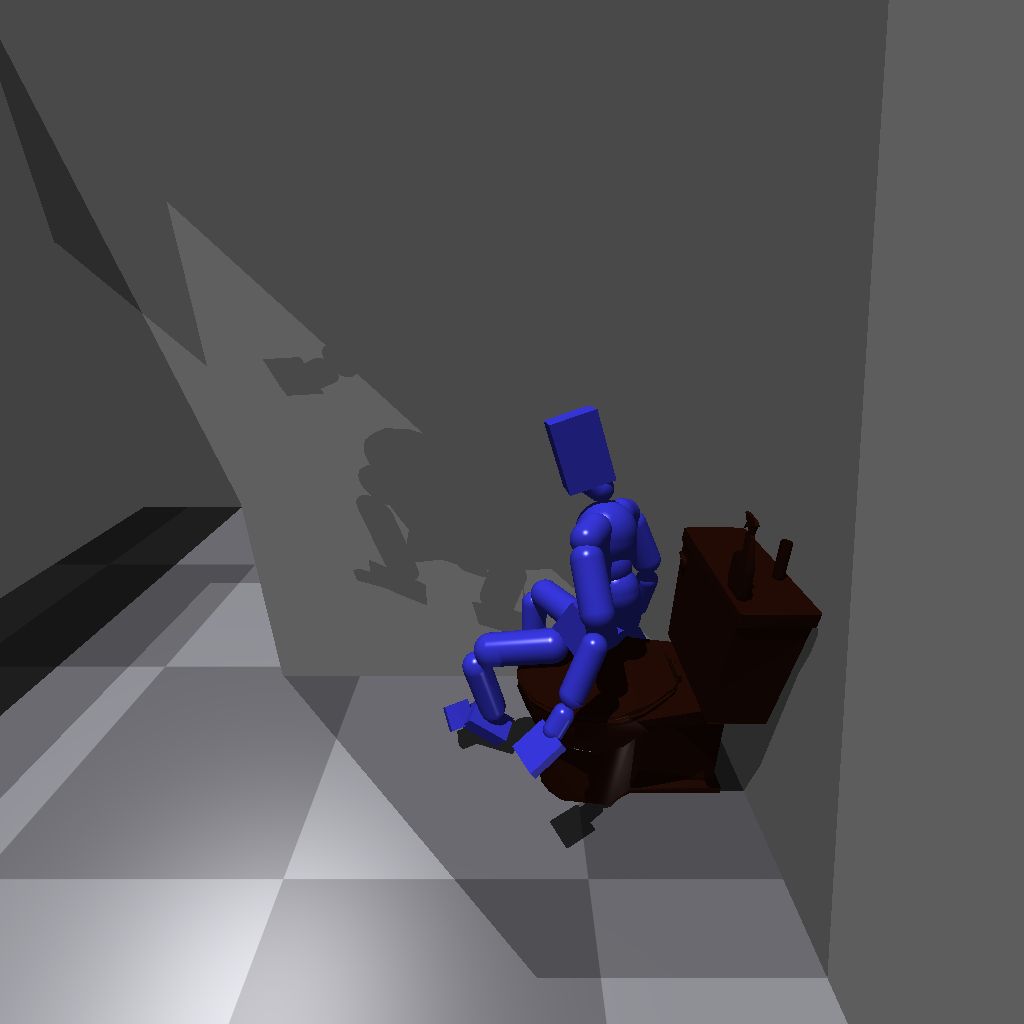}
        \caption{\emph{SIT}~(\emph{Room9})}
    \end{subfigure}\hfil%
    \begin{subfigure}[t]{0.25\linewidth}
        \centering
        \includegraphics[width=1.0\textwidth]{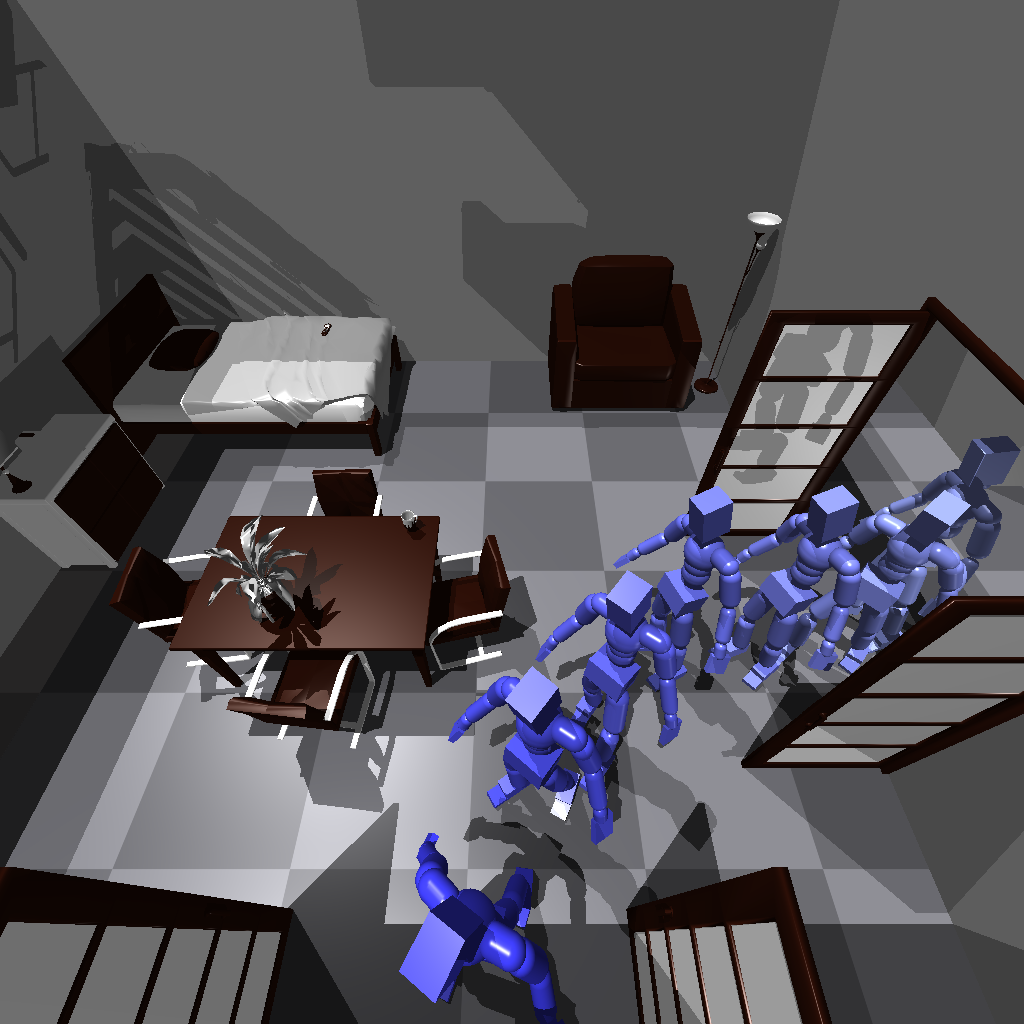}
        \caption{\emph{Exit}~(\emph{Room8})}
    \end{subfigure}\hfil
    \begin{subfigure}[t]{0.25\linewidth}
        \centering
        \includegraphics[width=1.0\textwidth]{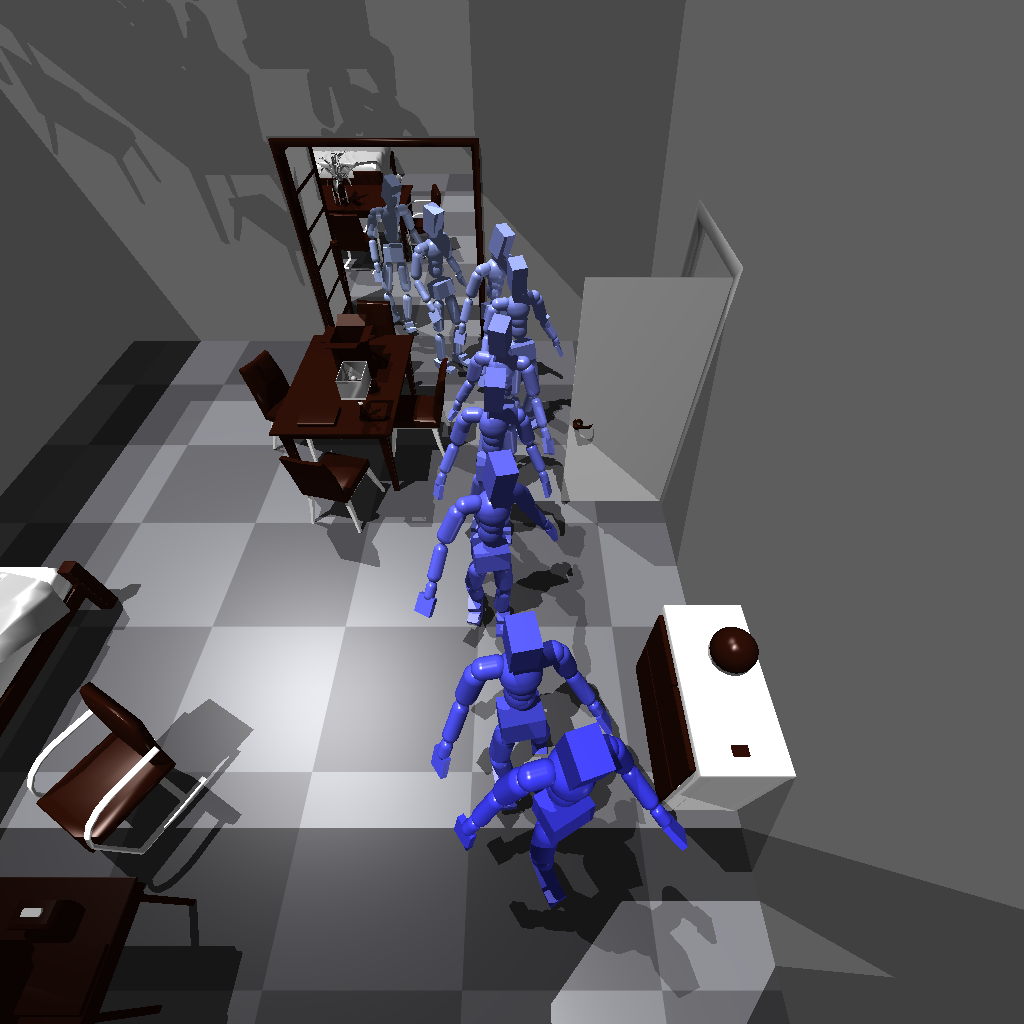}
        \caption{\emph{Exit}~(\emph{Room6})}
    \end{subfigure}\hfil
    \begin{subfigure}[t]{0.25\linewidth}
        \centering
        \includegraphics[width=1.0\textwidth]{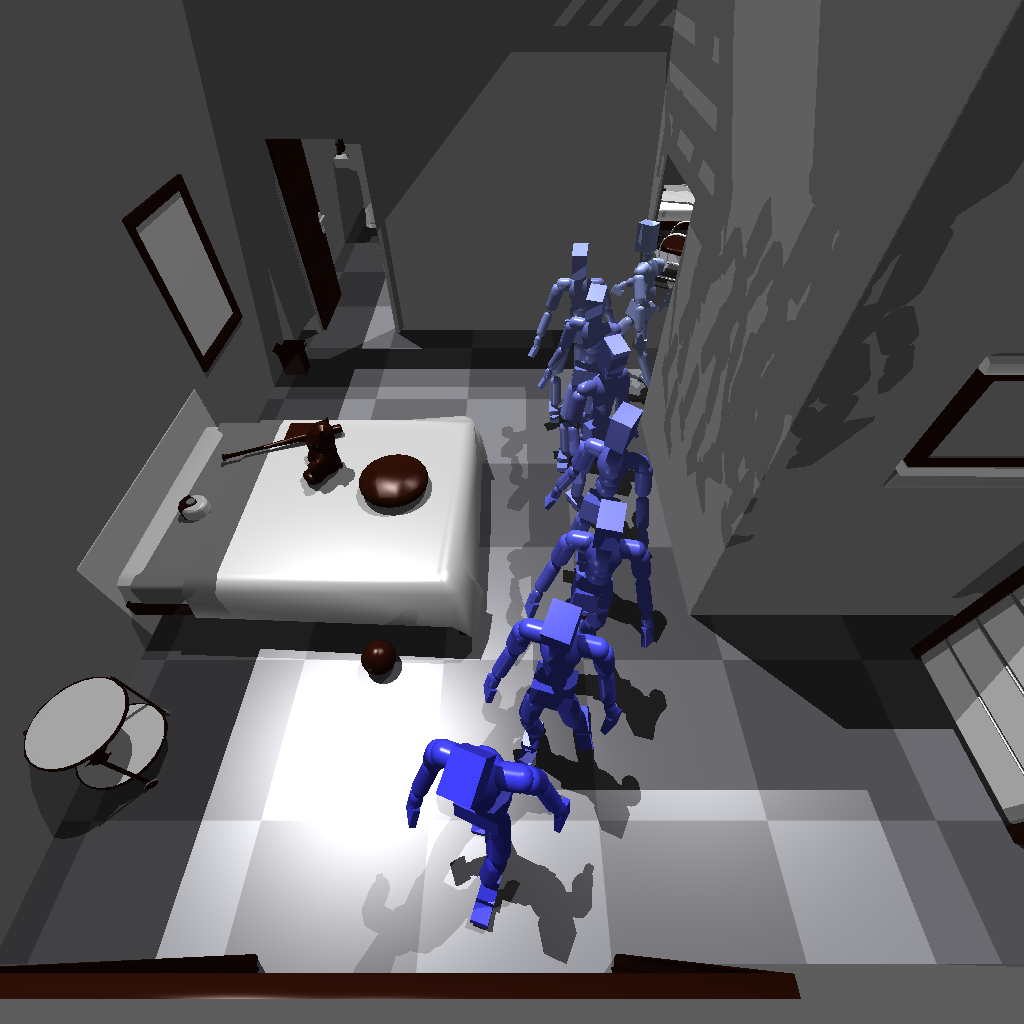}
        \caption{\emph{Exit}~(\emph{Room4})}
    \end{subfigure}\hfil%
    \begin{subfigure}[t]{0.25\linewidth}
        \centering
        \includegraphics[width=1.0\textwidth]{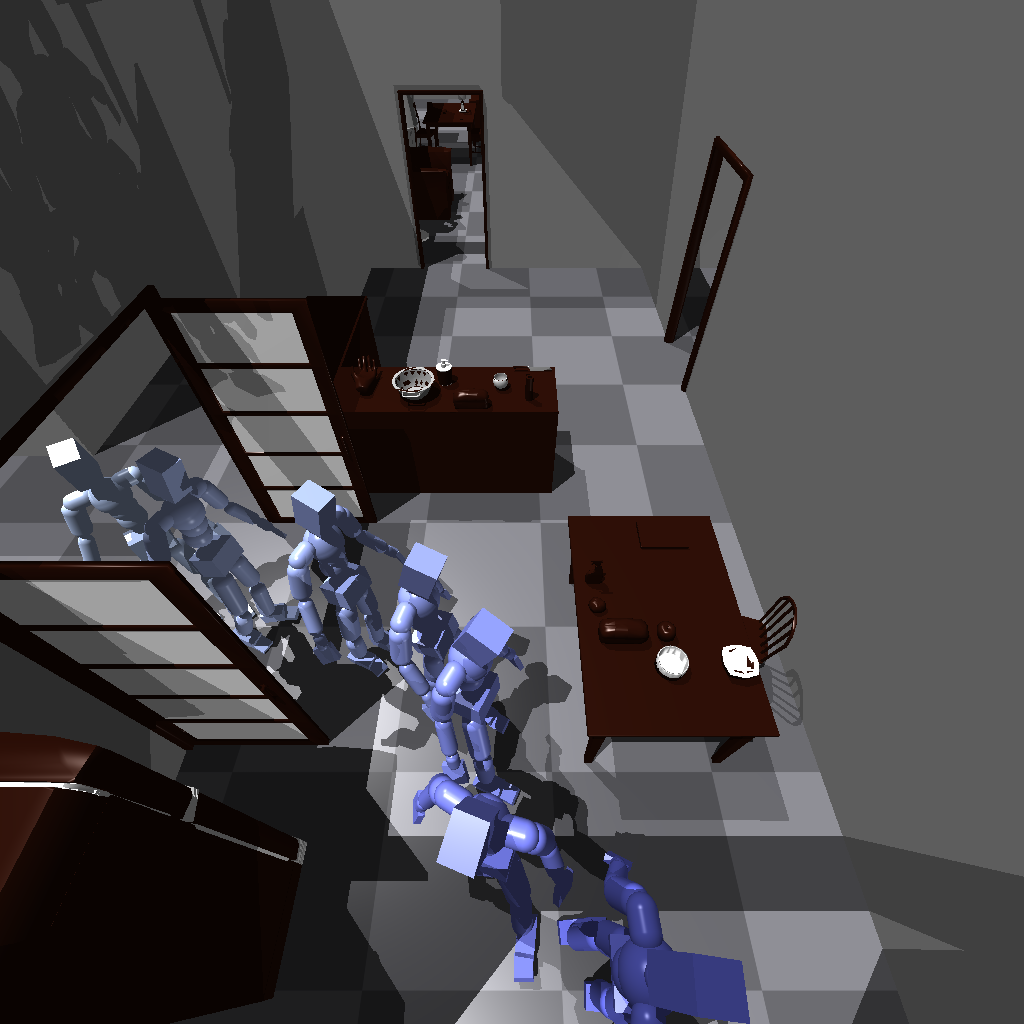}
        \caption{\emph{Exit}~(\emph{Room3})}
    \end{subfigure}\hfil%

    \caption{
        {\bf Visualization of human scene interaction.}
        (a) The \emph{Enter} task, (b)–(l) various tasks performed in each room, and (m)–(p) the \emph{Exit} task. For the \emph{Enter} and \emph{Exit} tasks, the humanoid is rendered in increasingly darker shades to indicate the passage of time.
    }
    \label{appendix_fig:hsi_further_qualitative_res}
\end{figure*}

%% file: figs/failure_cases.tex
% \begin{figure}[htbp]
% \captionsetup[subfigure]{justification=centering}
%     \centering
%     \begin{subfigure}[t]{0.5\linewidth}
%         \centering
%         \includegraphics[width=1.0\textwidth]{figs/failure_cases/oa_failure_case.png}
%         \caption{Obstacle avoidance task}
%     \end{subfigure}\hfil%
%     \begin{subfigure}[t]{0.5\linewidth}
%         \centering
%         \includegraphics[width=1.0\textwidth]{figs/failure_cases/oa_failure_case.png}
%         \caption{Obstacle avoidance task}
%     \end{subfigure}\hfil%
%     \begin{subfigure}[t]{0.5\linewidth}
%         \centering
%         \includegraphics[width=1.0\textwidth]{figs/failure_cases/oa_failure_case.png}
%         \caption{Obstacle avoidance task}
%     \end{subfigure}\hfil%
%     \begin{subfigure}[t]{0.5\linewidth}
%         \centering
%         \includegraphics[width=1.0\textwidth]{figs/failure_cases/oa_failure_case.png}
%         \caption{Obstacle avoidance task}
%     \end{subfigure}\hfil%\\
%     \caption{
%         {\bf Failure cases for human scene interaction task.}
%     }
%     \label{appendix_fig:failure_cases_hsi}
% \end{figure}
\begin{figure*}[t!]
% \captionsetup[subfigure]{justification=centering}
    \centering
    \includegraphics[width=1.0\textwidth]{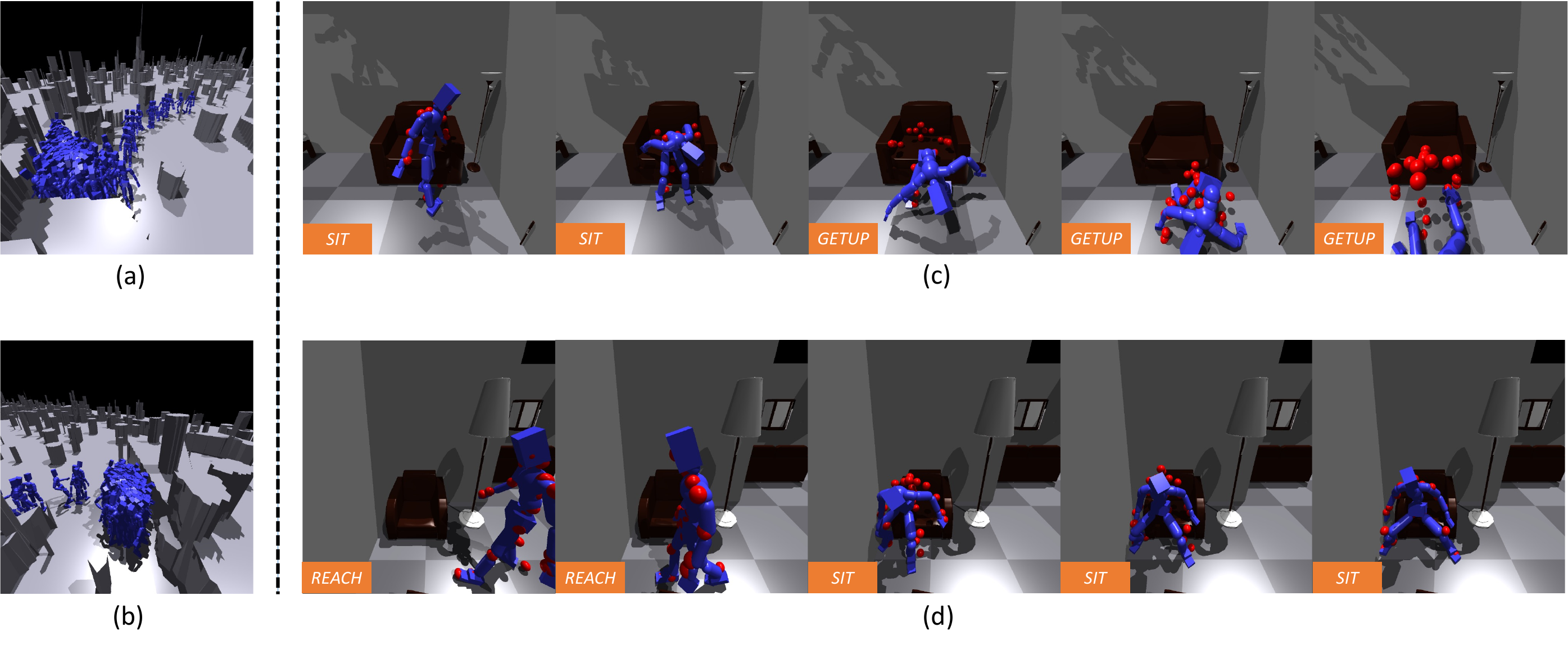}
    \caption{
        {\bf Visualization of failure cases.} 
        % (a)와 (b)는 각각 \emph{Room8}과 \emph{Room7}에 속한 subtask에 대한 failure case를 보여준다. (1)부터 (5)까지 시간 순으로 진행되며, (5)는 모션 실행 시 넘어져서 실패한 humanoid를 나타낸다. 이러한 실패는 모션 계획이 물체와의 상호작용하도록 잘 만들어지지 않았기 때문으로 보인다.
        % (a)와 (b)는 obstacle avoidance task에 대한 실패 사례를 나타내는데, humanoid가 모여 군중처럼 보이는 것은 벽처럼 넓게 놓인 장애물에 가로막혀 생긴 현상이다. (C)와 (d)는 각각 HSI task의 \emph{Room8}과 \emph{Room7}에서 급격히 성공률 하락을 이끄는 subtas에 대한 실패 사례이다. (c)는 자세가 불안정한 상태로 앉았다가 일어나면서 실패하고, (d)는 암체어에 대한 heading 방향이 이상한 상태로 앉다가 시간을 초과해서 실패한다.
        (a) and (b) illustrate failure cases in the obstacle avoidance task.
        Motion plans are repeatedly adjusted to avoid the obstacle but ultimately fail to bypass the wide, wall-like structure, leading to accumulation near the barrier.
        % The humanoids appear crowded, resembling a group of agents, due to being blocked by a wide, wall-like arrangement of obstacles.
        (c) and (d) show subtasks with significantly reduced success rates in the HSI task, specifically in \emph{Room8} and \emph{Room2}, respectively.
        The agent in (c) fails while attempting to stand up from an unstable sitting posture.
        The agent in (d) sits facing the armchair with an incorrect heading and fails due to simulation timeout.
    }
    \label{appendix_fig:failure_cases}
\end{figure*}

%% file: figs/supp_scene_process.tex
\begin{figure*}[t!]
    \centering
    \includegraphics[width=0.85\textwidth]{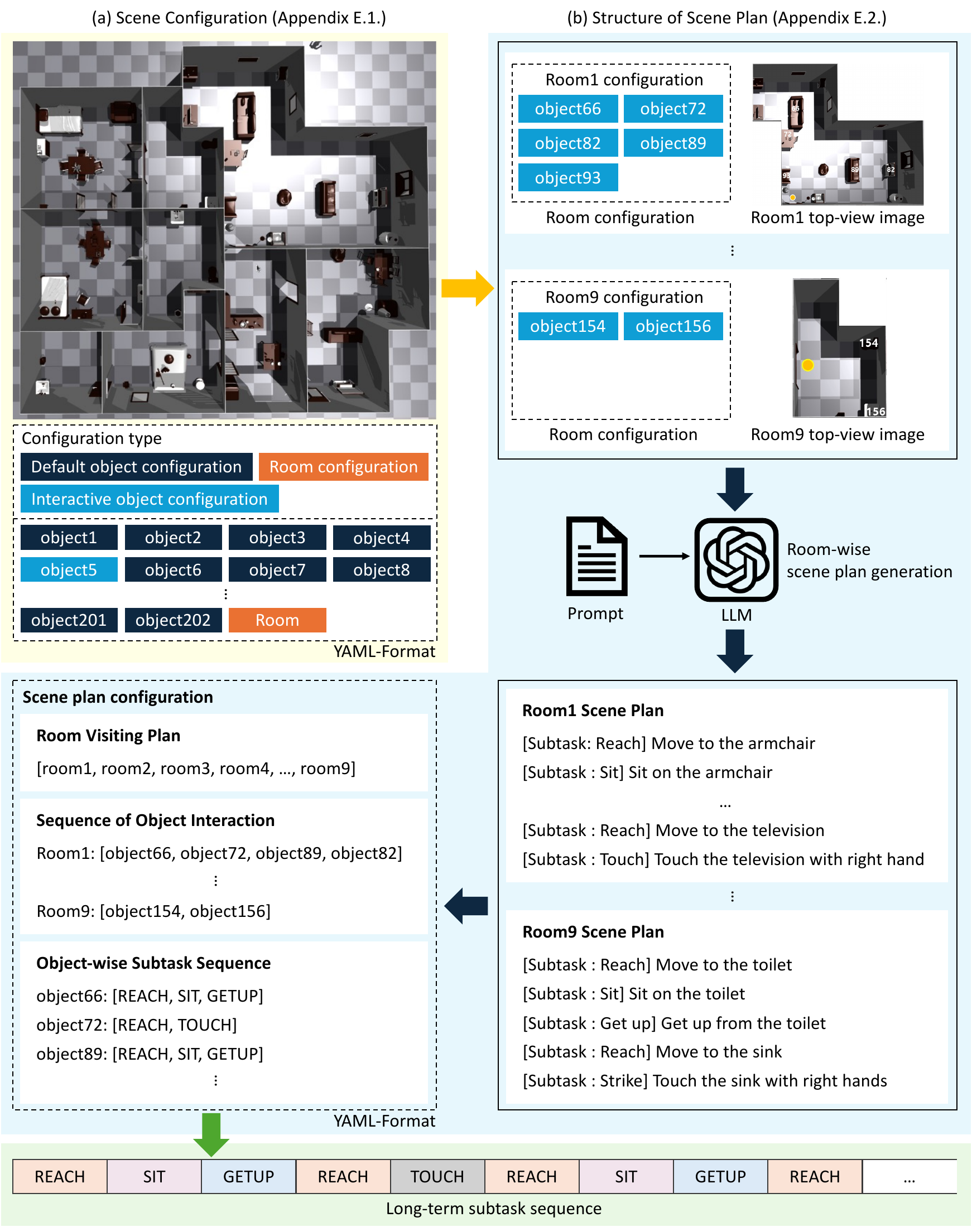}
    \caption{
        {\bf The overall process of constructing the scene and scene plan.} (a) The scene is represented in a YAML-format configuration, where objects are categorized into default objects, interactive objects, and rooms. (b) The LLM takes as input a YAML-format room configuration containing only the interactive objects in each room, a top-view image, and a prompt, and generates a room-wise scene plan.
        The room-wise scene plan consists of a sequence of subtasks. The generated room-wise scene plan is then reorganized into a scene plan configuration that includes object interaction sequences and object-wise subtask sequences. These are used to generate a long-term motion sequence by combining them with the predefined room visiting plan.
    }
    \label{appendix_fig:fig_supp_scene_process}
\end{figure*}

% \begin{strip}\centering
% \includegraphics[width=1.0\textwidth]{figs/supplementary/supp_scene_process.png}
% \captionof{figure}{
%     \textbf{The overall procedure of constructing scene and scene plan}
% }
% \label{fig_supp_scene_process}}
% \end{strip}

% \begin{center}
%     \includegraphics[height=1.0\textheight]{figs/supplementary/supp_scene_process.png}
%     \captionof{figure}{
%         {\bf The overall procedure of constructing scene and scene plan}\\
%     }
%     \label{fig_supp_scene_process}
% \end{center}

% (a) 주어진 scene은 default object, interactive object, room으로 구분되어 YAML-Format의 configuration으로 표현된다. (b) LLM은 각 방의 interactive object만을 포함한 YAML-Format의 room configuration, top-view 이미지 그리고 prompt를 입력으로 받아 subtask seqeunce로 이루어진 room-wise scene plan을 생성한다. 이때 각 subtask는 자신을 설명하는 description과 함께 생성되며, 생성된 description은 모델에서 사용되지 않는다. 생성된 room-wise scene plan은 sequence of obejct interaction과 object-wise subtask sequence로 분리되어 scene plan configuration을 구성하는데 사용된다. 이때, scene plan configuration에는 manual하게 정의된 room visit plan이 함께 포함된다. 

%% file: tables/scene_interaction_task/supp_scene_yaml.tex
% \begin{table*}[t]
%     \centering
%     \hline
%     \begin{lstlisting}[basicstyle=\ttfamily\small]
    % [object0]:
    %   assetRoot: [assetRoot]
    %   assetFileName: Tomato_1.urdf
    %   y_up: false
    %   fixBase: true
    %   rotation:
    %     x: 0.0
    %     y: -1.0
    %     z: -0.0
    %     w: 0.0
    %   translation:
    %     x: -9.625164
    %     y: 0.314864
    %     z: 17.745814
%     \end{lstlisting}
%     \\
%     \hline

%     \caption{
%         \textbf{Examplication of the scene configuration defined in YAML format.}
%     }
%     \label{table_supp_scene_yaml}
% \end{table*}

\begin{table*}[t!]
    \centering
    \begin{minipage}{\textwidth}%
        \fontsize{9}{9.5}\selectfont % 글꼴 크기 직접 설정
        \begin{tabular}{p{0.31\textwidth} p{0.31\textwidth} p{0.31\textwidth}}
            \hline \toprule
            \textbf{Default Object Configuration} & \textbf{Interactable Object Configuration} & \textbf{Room Configuration} \\
            \hline \midrule

            % 1 column
            object0: \par
            \hspace{2em}assetRoot: [assetRoot] \par
            \hspace{2em}assetFileName: [assetFileName] \par
            \hspace{2em}y\_up: true \par
            \hspace{2em}fixBase: true \par
            \hspace{2em}rotation: \par
            \hspace{4em}x: 0.0 \par
            \hspace{4em}y: -1.0 \par
            \hspace{4em}z: -0.0 \par
            \hspace{4em}w: 0.0 \par
            \hspace{2em}translation: \par
            \hspace{4em}x: -9.625164 \par
            \hspace{4em}y: 0.314864 \par
            \hspace{4em}z: 17.745814 
             
             & 

            object5: \par
            \hspace{2em}assetRoot: [assetRoot] \par
            \hspace{2em}assetFileName: [assetFileName] \par
            \hspace{2em}y\_up: true \par
            \hspace{2em}fixBase: true \par
            \hspace{2em}rotation: \par
            \hspace{4em}x: 0.0 \par
            \hspace{4em}y: 0.0 \par
            \hspace{4em}z: -1.0 \par
            \hspace{4em}w: 0.0 \par
            \hspace{2em}translation: \par
            \hspace{4em}x: -9.625703 \par
            \hspace{4em}y: 0.008309 \par
            \hspace{4em}z: 17.795589 \par
            \hspace{2em}stand\_point: \par
            \hspace{4em}x: -0.758882 \par
            \hspace{4em}y: 0.0 \par
            \hspace{4em}z: 0.0 \par
            \hspace{2em}touch\_point: \par
            \hspace{4em}x: -0.458883 \par
            \hspace{4em}y: 0.931318 \par
            \hspace{4em}z: 0.0 \par
            \hspace{2em}obj\_dims: \par
            \hspace{4em}x: -9.625703 \par
            \hspace{4em}y: 0.008309 \par
            \hspace{4em}z: 17.795589 \par             
             & 

            roomContour: \par
            \hspace{2em}assetRoot: [assetRoot] \par
            \hspace{2em}assetFileName: [assetFileName] \par
            \hspace{2em}fixBase: true \par
            \hspace{2em}rotation: \par
            \hspace{4em}x: 0.7071067 \par
            \hspace{4em}y: 0.0 \par
            \hspace{4em}z: 0.0 \par
            \hspace{4em}w: 0.7071067 \par
            \hspace{2em}translation: \par
            \hspace{4em}x: 0.0 \par
            \hspace{4em}y: 0.0 \par
            \hspace{4em}z: 0.0 \par
            rooms: \par
            \hspace{2em} room1: \par
            \hspace{4em} start\_point: \par
            \hspace{6em}x: 0.0 \par
            \hspace{6em}y: 0.0 \par
            \hspace{6em}z: 0.0 \par
            objList: [object0, object1, ...] \par
            interactionObj: [object5, object19, ...] \\

            \bottomrule \hline
        \end{tabular}
    \end{minipage}
    \caption{
        \textbf{Scene structure configuration in YAML format.} The scene configuration consists of three parts: default object configuration, interactive object configuration, and room configuration. All objects in the scene are initially defined using the default object configuration, which includes the mesh path and transformation matrix. Interactive objects are further specified with an interactive object configuration, which includes 3D spatial bounding box and target point information. Objects such as room contours are defined using a room configuration, which includes the room's door positions and two lists that specify the objects in the room and the objects available for interaction.
    }
    \label{table_supp_scene_yaml}
\end{table*}

% Scene configuration은 default object configuration, interacable object configuration, room configuration으로 구성된다. scene에 존재하는 모든 object는 default object configuration 형태로 정의되며, object asset file 정보와 transformation matrix 정보가 포함된다. 모든 default object 중 상호작용이 가능한 object는 3D spatial bounding box 정보와 interaction을 위한 target point 정보가 추가된 interactive object configuration 형태로 정의된다. Scene을 구성하는 object 중에는 room contour와 같은 object가 존재한다. 이러한 object는 room의 시작 위치 정보, room에 존재하는 object와 상호작용이 가능한 object의 리스트가 추가된 room configuration 형태로 정의된다.

%% file: tables/scene_interaction_task/supp_scene_plan_yaml.tex
\begin{table*}[t!]
    \centering
    \begin{minipage}{\textwidth}%
        \fontsize{9}{9.5}\selectfont % 글꼴 크기 직접 설정
        \begin{tabular}{p{0.30\textwidth} p{0.33\textwidth} p{0.30\textwidth}}
        
            \hline \toprule
            \textbf{Room Visiting Plan} & \textbf{Sequence of Object Interaction} & \textbf{Object-wise Subtask Sequence}\\
            \hline \midrule

            [room1, room2, room3, room4, room5, room6, room7, room8, room9] 

            & 
            
            room1: [object66, object72, object89, object82] \par  
            room2: [object37, object29, object46, object52] \par  
            room3: [object27, object5, object19] \par  
            room4: [object164] \par  
            room5: [object179] \par  
            room6: [object107, object101, object124] \par  
            room7: [object128, object131] \par  
            room8: [object143, object136, object148] \par  
            room9: [object154, object156] \par

            & 
            
            object66: [REACH, SIT, GETUP] \par
            object72: [REACH, TOUCH] \par
            object82: [REACH, TOUCH] \par
            object89: [REACH, SIT, GETUP] \par
            object93: [REACH, SIT, GETUP] \par
            object29: [REACH, TOUCH] \par
            object37: [REACH, SIT, GETUP] \par
            object46: [REACH, TOUCH] \par
            object52: [REACH, SIT, GETUP] \par
        
            object5: [REACH, TOUCH] \par
            object19: [REACH, TOUCH] \par
            object27: [REACH, TOUCH] \par
        
            object164: [REACH, SIT, GETUP] \par
        
            object179: [REACH, TOUCH] \par
        
            object101: [REACH, SIT, GETUP] \par
            object107: [REACH, TOUCH] \par
            object124: [REACH, TOUCH] \par
        
            object128: [REACH, SIT, GETUP] \par
            object131: [REACH, TOUCH] \par
        
            object136: [REACH, SIT, GETUP] \par
            object143: [REACH, TOUCH] \par
            object148: [REACH, SIT, GETUP] \par
        
            object154: [REACH, SIT, GETUP] \par
            object156: [REACH, TOUCH]\\
            
            \bottomrule \hline
        \end{tabular}
    \end{minipage}
    \caption{
        \textbf{Scene plan configuration in YAML format.}
        The scene plan configuration includes the room visiting plan, the sequence of object interaction, and the object-wise subtask sequence. The object-wise subtask sequence and the sequence of object interactions are derived from the room-wise scene plan generated by the LLM.
        %The room visit plan is manually defined.
    }
    \label{table_supp_scene_plan_yaml}
\end{table*}

%% file: tables/scene_interaction_task/supp_llm_stages.tex
\begin{table*}[t!]
    \centering
    \begin{minipage}{\textwidth}%
        \fontsize{9}{1.0}\selectfont % 글꼴 크기 직접 설정
        \begin{tabular}{p{0.98\textwidth}}
        
            \hline \toprule
            \textbf{Example of an LLM Output}\\
            \hline \midrule

            \textbf{(1) Scene Description:} \par
            object66 (Sofa\_201\_4.urdf): A large sofa positioned centrally near the northern wall, oriented towards the southern area of the room.\par
            
            object72 (Dining\_Table\_203\_1.urdf): A rectangular dining table placed in the northwest area, situated to the northwest of the large sofa (object66).\par
            
            object82 (Television\_16.urdf): A television located on the far eastern wall, directly facing the small sofa (object89).\par
            
            object89 (Sofa\_207\_4.urdf): A small sofa centrally positioned in the eastern area of the room, facing east towards the television (object82).\par
            
            object93 (ArmChair\_216\_1.urdf): An armchair placed in the southwest part of the room, near the starting location of the humanoid and directly south of the dining table (object72).\par

            \\[1em]
            
            \textbf{(2) Scenario:}\par
            The humanoid starts at the southwest corner (yellow circle). It first approaches the nearby armchair (object93) to sit briefly. Next, it stands up and walks towards the dining table (object72), touching the table to examine it. The humanoid then moves toward the large sofa (object66), sits down to rest, gets up afterward, and continues eastward toward the small sofa (object89). After sitting briefly on the small sofa to view the TV (object82), the humanoid gets up, walks to the television, and touches it to turn it off, completing interactions with all objects in the room.\par

            \\[1em]
            
            \textbf{(3) Scene Plan:}\par
            [Subtask: Reach] Move to the armchair (object93: ArmChair\_216\_1.urdf)\par
            
            [Subtask: Sit] Sit on the armchair (object93: ArmChair\_216\_1.urdf) [BODY: Pelvis]\par
            
            [Subtask: Get up] Get up from the armchair (object93: ArmChair\_216\_1.urdf)\par
            
            [Subtask: Reach] Move to the dining table (object72: Dining\_Table\_203\_1.urdf)\par
            
            [Subtask: Touch] Touch the dining table (object72: Dining\_Table\_203\_1.urdf) with right hand [BODY: Right hand]\par
            
            [Subtask: Reach] Move to the large sofa (object66: Sofa\_201\_4.urdf)\par
            
            [Subtask: Sit] Sit on the large sofa (object66: Sofa\_201\_4.urdf) [BODY: Pelvis]\par
            
            [Subtask: Get up] Get up from the large sofa (object66: Sofa\_201\_4.urdf)\par
            
            [Subtask: Reach] Move to the small sofa (object89: Sofa\_207\_4.urdf)\par
            
            [Subtask: Sit] Sit on the small sofa (object89: Sofa\_207\_4.urdf) [BODY: Pelvis]\par
            
            [Subtask: Get up] Get up from the small sofa (object89: Sofa\_207\_4.urdf)\par
            
            [Subtask: Reach] Move to the television (object82: Television\_16.urdf)\par
            
            [Subtask: Touch] Touch the television (object82: Television\_16.urdf) with right hand [BODY: Right hand] \\

            \bottomrule \hline
        \end{tabular}
    \end{minipage}
    \caption{
        \textbf{Example of an LLM Output.} The LLM is prompted to generate three sections: a scene description, a scenario, and a room-wise scene plan.
    }
    \label{table_supp_prompt_stages}
\end{table*}

%% file: figs/supp_llm_input.tex
\begin{figure*}[t!]
    \centering
    \includegraphics[width=0.90\textwidth]{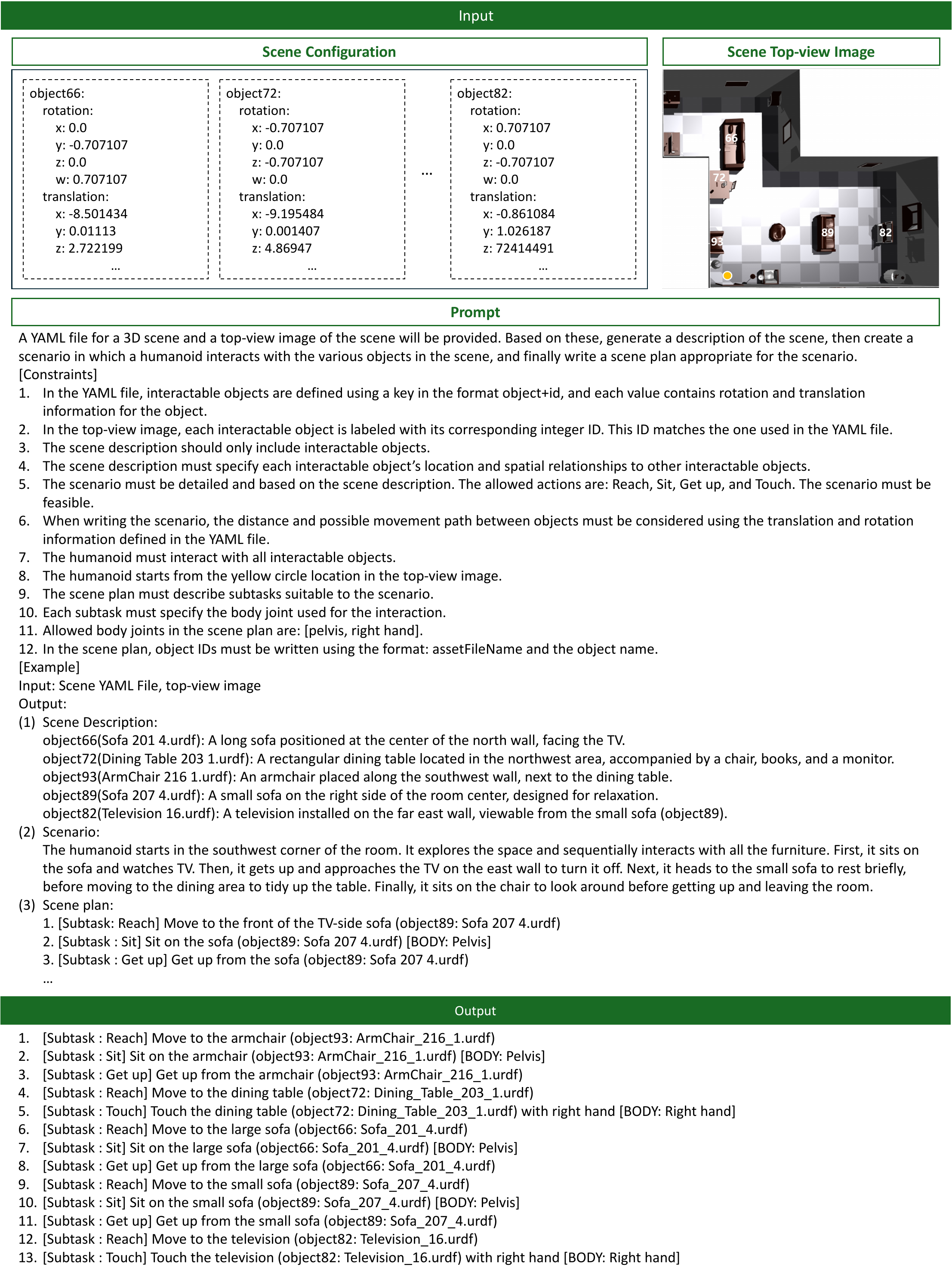}
    \caption{
        {\bf Example of input to an LLM and its output.}
    }
    \label{fig_supp_llm_input}
\end{figure*}